\documentclass[11pt,letterpaper]{article}

\usepackage[T1]{fontenc}
\usepackage[utf8]{inputenc}
\usepackage{amsmath}
\usepackage{amsthm}
\usepackage{newtxtext,newtxmath}
\usepackage[letterpaper,margin=0.82in,headheight=15pt,headsep=18pt,footskip=28pt]{geometry}
\usepackage{microtype}
\usepackage{graphicx}
\usepackage{booktabs}
\usepackage{array}
\usepackage{tabularx}
\usepackage{longtable}
\usepackage{multirow}
\usepackage{makecell}
\usepackage{adjustbox}
\usepackage{float}
\usepackage{wrapfig}
\usepackage{nicefrac}
\usepackage{siunitx}
\usepackage{mathtools}
\usepackage{bm}
\usepackage{cases}
\usepackage{xspace}
\usepackage{enumitem}
\usepackage{needspace}
\usepackage{placeins}
\usepackage{pifont}
\usepackage[table]{xcolor}
\usepackage{colortbl}
\usepackage[ruled,vlined]{algorithm2e}
\usepackage{cite}
\let\citep\cite
\usepackage{url}
\usepackage{tcolorbox}
\tcbuselibrary{theorems,skins,breakable}
\usepackage{etoolbox}
\usepackage{titlesec}
\usepackage{caption}
\usepackage{fancyhdr}
\usepackage[unicode,colorlinks=true,linkcolor=arxivblue,citecolor=arxivblue,urlcolor=arxivblue]{hyperref}

\graphicspath{{Figures/}}
\definecolor{arxivblue}{RGB}{32,82,122}
\definecolor{arxivred}{RGB}{142,27,31}
\definecolor{arxivlight}{RGB}{244,247,250}
\definecolor{myblue}{RGB}{82,140,182}
\definecolor{maincolor}{rgb}{0.2,0.5,0.7}
\definecolor{forestgreen}{RGB}{34,139,34}
\definecolor{darkgreen}{RGB}{0,0,0}
\definecolor{titlebgcolor}{RGB}{82,140,182}
\definecolor{textbgcolor}{RGB}{219,232,241}
\definecolor{mygrey}{RGB}{128,128,128}
\definecolor{verylightgray}{RGB}{240,240,240}
\definecolor{journalblue}{RGB}{0,0,0}
\definecolor{revisionred}{RGB}{0,0,0}

\titleformat{\section}{\Large\bfseries\color{arxivblue}}{\thesection}{0.65em}{}
\titleformat{\subsection}{\large\bfseries\color{arxivblue!88!black}}{\thesubsection}{0.60em}{}
\titleformat{\subsubsection}{\normalsize\bfseries}{\thesubsubsection}{0.55em}{}
\titlespacing*{\section}{0pt}{2.4ex plus .6ex minus .2ex}{1.0ex}
\titlespacing*{\subsection}{0pt}{2.0ex plus .4ex minus .2ex}{0.7ex}

\renewenvironment{abstract}{\begin{tcolorbox}[enhanced,breakable,colback=arxivlight,colframe=arxivblue!45,
    boxrule=0.55pt,arc=1.5pt,left=8pt,right=8pt,top=7pt,bottom=7pt]
  \small\noindent\textbf{Abstract.}\hspace{0.45em}}{\end{tcolorbox}}

\newcommand{\journaltext}[1]{{\color{journalblue}#1}}
\newenvironment{journalnew}{\par\begingroup\color{journalblue}}{\par\endgroup}
\newcommand{\revisiontext}[1]{{\color{revisionred}#1}}
\newenvironment{revisionnew}{\par\begingroup\color{revisionred}}{\par\endgroup}

\makeatletter
\newcommand{\appendixsection}[1]{\clearpage
  \section{#1}\suppressfloats[t]\addcontentsline{apc}{section}{\protect\numberline{\thesection}#1}}
\newcommand{\appendixsubsection}[1]{\subsection{#1}\addcontentsline{apc}{subsection}{\protect\numberline{\thesubsection}#1}}
\newcommand{\appendixcontents}{\begingroup
  \fontsize{9.5pt}{11.45pt}\selectfont
  \setcounter{tocdepth}{2}\setlength{\parskip}{0pt}\renewcommand*\l@section[2]{\vspace{0.24em}\@dottedtocline{1}{0em}{2.6em}{\bfseries\color{arxivblue}##1}{\bfseries ##2}}\renewcommand*\l@subsection[2]{\vspace{0.06em}\@dottedtocline{2}{1.4em}{3.7em}{##1}{##2}}\@starttoc{apc}\endgroup
}
\makeatother

\newtheorem{definition}{Definition}

\newtheorem{theorem}{Theorem}
\newtheorem{proposition}{Proposition}
\newtheorem{corollary}{Corollary}
\newtheorem{lemma}{Lemma}
\newtheorem{remark}{Remark}

\newtcolorbox{infobox}[1][]{
  colframe=titlebgcolor!60,colback=textbgcolor!29,boxrule=0.6pt,arc=1pt,
  left=4pt,right=4pt,top=4pt,bottom=4pt,#1
}
\newtcolorbox{thmbox}[1][]{
  enhanced,colback=textbgcolor!29,frame hidden,boxrule=0pt,
  borderline north={1.0pt}{0pt}{titlebgcolor!70},
  borderline south={0.6pt}{0pt}{titlebgcolor!70},
  left=6pt,right=6pt,top=5pt,bottom=5pt,arc=0mm,#1
}

\def\equationautorefname~#1\null{Equation~(#1)\null}
\def\sectionautorefname~#1\null{Section~#1\null}
\def\subsectionautorefname~#1\null{Section~#1\null}
\def\figureautorefname~#1\null{Fig.~#1\null}
\def\tableautorefname~#1\null{Table~#1\null}
\def\algorithmautorefname~#1\null{Algorithm~#1\null}
\def\theoremautorefname~#1\null{Theorem~#1\null}
\def\propositionautorefname~#1\null{Proposition~#1\null}
\def\corollaryautorefname~#1\null{Corollary~#1\null}
\def\lemmaautorefname~#1\null{Lemma~#1\null}
\def\definitionautorefname~#1\null{Definition~#1\null}
\def\remarkautorefname~#1\null{Remark~#1\null}
\def\assumptionautorefname~#1\null{Assumption~#1\null}

\title{\vspace{-1.5em}\textbf{Hyperbolic Multimodal Continual Learning:}\\[-0.05em]
\textbf{A Closest-Admissible Solution}}

\author{Jiahong Liu\textsuperscript{1}, Ming Shen\textsuperscript{1},
Xiaohao Liu\textsuperscript{2}, Rex Ying\textsuperscript{3},\\[-0.15em]
Menglin Yang\textsuperscript{4}, Tat-Seng Chua\textsuperscript{2},
and Irwin King\textsuperscript{1}\\[0.55em]
\small\textsuperscript{1}The Chinese University of Hong Kong
\quad \textsuperscript{2}National University of Singapore\\[-0.05em]
\small\textsuperscript{3}Yale University
\quad \textsuperscript{4}The Hong Kong University of Science and Technology (Guangzhou)\\[0.30em]
\small Email: \texttt{jiahong.liu21@gmail.com}, \texttt{king@cse.cuhk.edu.hk}
}
\date{}

\hypersetup{
  pdftitle={Hyperbolic Multimodal Continual Learning: A Closest-Admissible Solution},
  pdfauthor={Jiahong Liu, Ming Shen, Xiaohao Liu, Rex Ying, Menglin Yang, Tat-Seng Chua, Irwin King}
}

\begin{document}
\maketitle
\thispagestyle{fancy}

\begin{abstract}
Existing continual-learning methods protect parameters, replayed examples, or Euclidean feature subspaces. When applied to hyperbolic multimodal models, they do not explicitly preserve the Lorentz geometry that jointly encodes within-modality similarity, cross-modal correspondence, and semantic hierarchy; sequential updates can therefore retain task scores while still distorting previously learned relations. We address this gap with Hyperbolic Multimodal Continual Learning (HMCL). We show that preserving the old multimodal geometry amounts to restricting all modalities to one shared hyperbolic isometry, which induces a family of admissible first-order parameter changes. We formulate a joint closest-admissible (CA) correction that retains the shared rotation best matching the candidate modal updates; its minimal-rotation (MR) special case fixes this rotation to zero. Both variants correct the displacement realized by AdamW, and task anchoring bounds within-task accumulation while preserving learning freedom. Across a unified 16-task classification--retrieval stream with three hyperbolic backbones, HMCL improves final performance and backward transfer over sequential fine-tuning and four continual-learning baselines; HMCL-CA gives the highest Overall score on every backbone. A modality-extended stream confirms the retrieval gains. Representation analyses find 81.2--95.5\% less radial, angular, cross-modal, and paired-distance drift; ImageNet--WordNet results show better semantic ancestry and radial hierarchy. Code and data download links are in the \href{https://github.com/HUBERILT/HMCL_ICML/tree/extension}{code repository}.
\end{abstract}

\begin{center}
\small\textbf{Keywords:} continual learning; hyperbolic geometry; multimodal learning; vision--language representation learning
\end{center}
\vspace{0.4em}

\section{Introduction}

\begin{wrapfigure}{r}{0.52\textwidth}
\vspace{-0.75\baselineskip}
\centering
\includegraphics[width=0.84\linewidth]{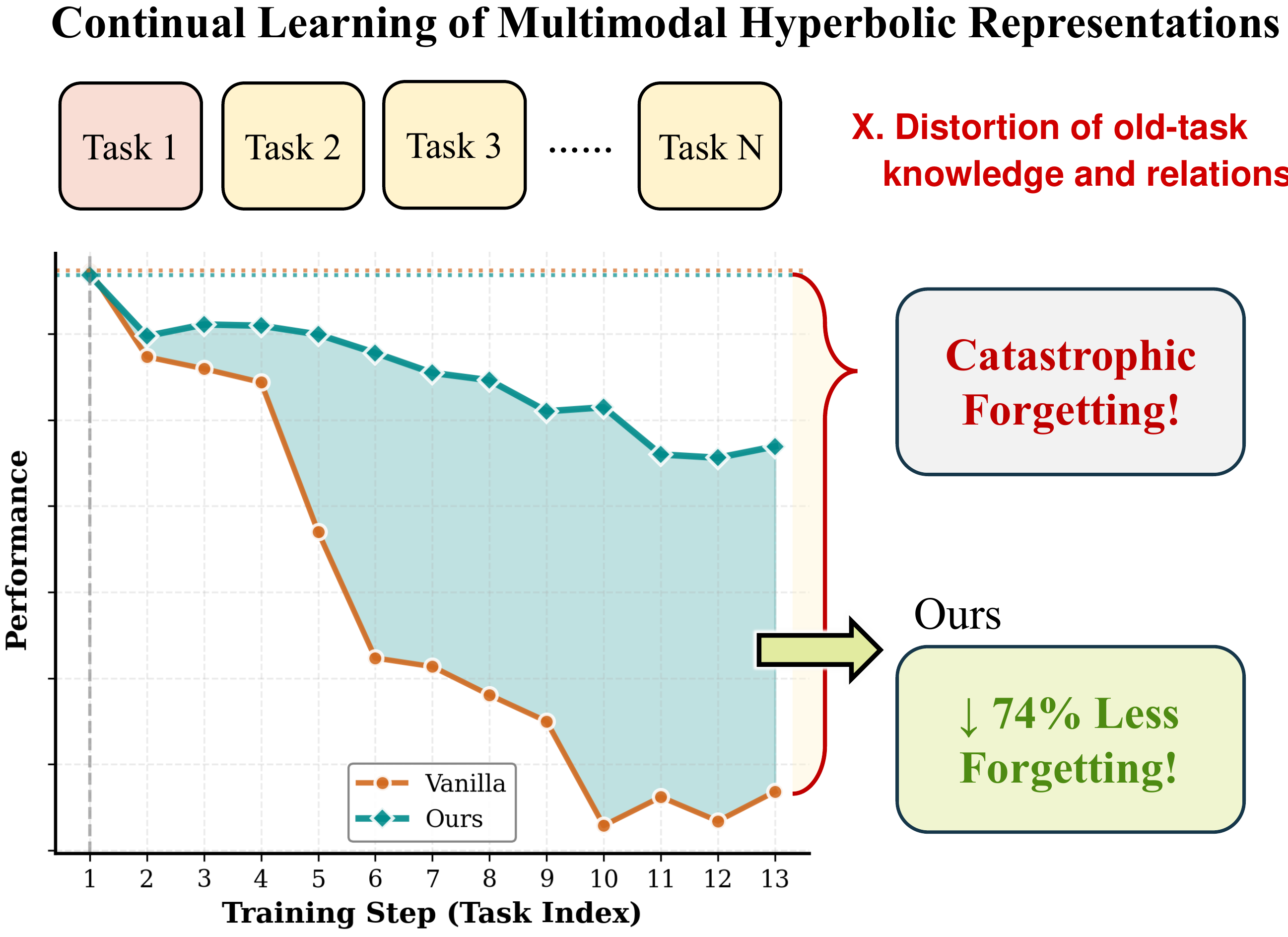}
\captionsetup{font=footnotesize}
\caption{\textbf{Continual adaptation of hyperbolic representations.}
Hyperbolic multimodal embeddings jointly encode task semantics and their hierarchy. Unconstrained sequential optimization may gradually distort this geometry and weaken structures learned from earlier tasks. Geometry-aware updates can reduce this form of forgetting.}
\label{fig:overview}
\vspace{-0.5\baselineskip}
\end{wrapfigure}

\textbf{M}ultimodal contrastive learning (MCL) has become a standard approach for constructing representations shared across data types~\cite{radford2021learning,he2020momentum,yu2024recent}. Trained with large paired corpora and contrastive objectives, these models align modality-specific observations in a semantic space that supports effective zero-shot transfer~\cite{han2023imagebind,chen2023vast,liu2025principled}. Most existing systems place this shared space in a high-dimensional Euclidean domain and therefore treat the organization of semantic concepts as flat.

Semantic organization across modalities is rarely flat, however. A broad concept generally covers numerous specialized instances, creating asymmetric, partially ordered relations between different levels of abstraction. Uniform Euclidean neighborhoods do not explicitly distinguish such general-to-specific structure, making these relations difficult to represent without distortion.

\begin{revisionnew}
\textbf{Hyperbolic Multimodal Learning.}
Negative-curvature geometry offers a useful alternative for hierarchical semantics. MERU~\citep{desai2023hyperbolic} and subsequent compositional-entailment learning~\citep{palcompositional} show that hyperbolic geometry can support effective multimodal representations. Geodesic proximity reflects semantic similarity, whereas radial position and entailment relations encode specificity. Generic concepts tend to lie nearer the origin, and more specific concepts tend to lie farther away. Hyperbolic embeddings can therefore represent alignment and abstraction in the same space.
\end{revisionnew}

\begin{revisionnew}
Despite these advances, existing hyperbolic multimodal models have been developed mainly for static training distributions~\citep{desai2023hyperbolic,palcompositional,mandica2024hyperbolic,kim2024hype}. Real systems instead encounter new tasks, concepts, and domains over time, yet standard continual-learning mechanisms do not explicitly protect the Lorentz relations that organize multimodal similarity and hierarchy. How to adapt such models sequentially without deforming their learned geometry therefore remains open.
\end{revisionnew}

\textbf{Continual Learning as Geometric Preservation.}
For a hyperbolic model, forgetting involves more than reduced accuracy on an earlier task. Parameter changes may deform within-modality neighborhoods, cross-modal correspondences, and the radial ordering that records semantic abstraction. Once these quantities drift, both similarity and hierarchy inherited from old tasks are altered. This observation motivates our central question: \emph{which parameter changes preserve the previously learned hyperbolic multimodal geometry while leaving sufficient freedom to acquire a new task?}

\textbf{Our Approach.}
We view hyperbolic multimodal continual learning\footnote{A preliminary version of this work appeared at ICML 2026.} as optimization over updates that satisfy geometric preservation constraints, which we call \emph{admissible updates}. Under the assumptions in Section~\ref{sec:equivalent analysis}, old-task geometry is retained when all modalities change through the same hyperbolic isometry. A first-order form of this condition describes the corresponding parameter updates. We then pose a joint closest-admissible problem over the modal parameter changes and the generator of their shared rotation. Its unique solution is obtained from one Lyapunov equation and defines \textsc{HMCL-CA}; constraining the generator to zero yields the projection used by \textsc{HMCL-MR}. \journaltext{For the AdamW implementation evaluated in this work, HMCL corrects the parameter step produced after moment estimation and coordinate-wise scaling, and task anchoring limits its accumulation within a task.}

\textbf{Contributions.}
\revisiontext{To the best of our knowledge, this work is the first to study continual hyperbolic multimodal representations explicitly through their geometry.} Our main contributions are:
\begin{itemize}
    \item We relate retention of old multimodal knowledge to a shared hyperbolic isometry and derive the associated set of first-order parameter updates.
    \item \revisiontext{We formulate a joint closest-admissible problem whose shared-rotation solution is obtained from one Lyapunov equation. $\mathrm{HMCL}_{\mathrm{CA}}$ permits the common rotation selected by this solution, while $\mathrm{HMCL}_{\mathrm{MR}}$ restricts the same admissible family to zero rotation. Task anchoring preserves admissibility and bounds within-task accumulation.}
    \item \revisiontext{Across three backbones, both HMCL variants improve final classification and retrieval scores, overall performance, and BWT over all external baselines, with CA attaining the best Overall score in every backbone block. Representation-space analyses further show that these task-level gains coincide with 81.2--95.5\% less radial, angular, cross-modal, and paired-distance drift and with improvements in every reported ImageNet--WordNet hierarchy metric. The placement study identifies post-AdamW correction with task anchoring as the strongest stability--plasticity configuration.}
\end{itemize}

\Needspace{3\baselineskip}
\begin{journalnew}
\begin{revisionnew}
\textbf{Extension beyond the conference version.}
The ICML paper introduced the geometric preservation conditions and the zero-rotation MR update on a 15-task stream~\citep{liu2026hmcl}. This article generalizes that foundation with the rotation-permitting CA update, optimizer-consistent analysis, and broader geometric evaluation:
\begin{enumerate}[label=(\roman*),leftmargin=1.45em,topsep=2pt,itemsep=2pt,parsep=0pt]
    \item \textit{Closest-admissible CA solution.}
    We express preservation as a joint constrained least-squares problem over every modal parameter change and one shared skew-symmetric generator. Its Lyapunov-equation solution gives the rotation-permitting $\mathrm{HMCL}_{\mathrm{CA}}$ update; fixing $\boldsymbol\Omega=\mathbf0$ recovers the conference-version $\mathrm{HMCL}_{\mathrm{MR}}$ update and places both in one admissible family (Section~\ref{subsec: admissible updates}).
    \item \textit{Finite-step preservation and controlled accumulation.}
    Following the backbones' AdamW training paradigm, we correct the realized parameter displacement after moment estimation and preconditioning. We show that task-anchored CA and MR updates remain in the shared admissible family and that their accumulation is uniformly bounded, and we derive residual-drift and complexity results for practical low-rank protection (Section~\ref{subsec:optimizer_consistent_hmcl} and Appendix~\ref{apdx:optimizer_consistent}).
    \item \textit{Hierarchy-aware large-scale evaluation.}
    We add ImageNet-1K to form a unified 16-task continual stream across MERU-L, MERU-B, and HyCoCLIP-B. Its WordNet structure directly verifies that the predictive-retention gains coincide with better preservation of semantic ancestry and abstraction order, rather than only higher aggregate scores (Section~\ref{subsec:journal_extended_results}).
    \item \textit{Controlled component and trajectory analysis.}
    The component study shows that correcting AdamW's realized displacement is stronger than correcting its raw gradient and that task anchoring provides a further stability gain; the complete post-AdamW update therefore gives the best stability--plasticity balance (Section~\ref{subsec:hmcl_component_analysis}). Representation-drift and hierarchy analyses further verify that the gain accompanies less radial and relational distortion (Sections~\ref{sec:exp_drift}--\ref{sec:exp_case_study}).
\end{enumerate}
\end{revisionnew}
\end{journalnew}

\section{Related Work}

\subsection{Hyperbolic Multimodal Learning}
The volume of hyperbolic space grows exponentially with radius. This property makes hyperbolic space useful for hierarchical or scale-free data and has motivated a broad literature on non-Euclidean representation learning~\citep{krioukov2010hyperbolic,nickel2017poincare,nickel2018learning,ganea2018hyperbolic,shimizu2020hyperbolic,chami2019hyperbolic,yang2022hgnnreview,yang2024hypformer,mettes2024survey,liu2026hyperbolictutorial}, including recent calls to extend non-Euclidean geometry to foundation models~\citep{he2025position}. Hyperbolic contrastive learning has been developed for node embeddings and graph-level anomaly detection, together with explicit analyses of dimensional collapse~\citep{liu2022hgcl,zhang2025dimcollapse,fu2026hcglad}. Universal cone constructions further recover implicit hierarchies without prespecified parent--child relations~\citep{yang2024uhcone}. In recommendation and retrieval, hyperbolic geometry captures long-tailed interactions and hierarchical neighborhoods~\citep{yang2022hrcf,yang2022hicf,qiu2024hihpq}. In computer vision, it has also been studied for image representation~\citep{khrulkov2020hyperbolic}, hierarchy-aware zero-shot recognition~\citep{liu2020hyperbolic}, and contrastive modeling of scene--object structure~\citep{ge2023hyperbolic}; its benefits can depend on the task and radial geometry~\citep{moreira2024hyperbolic}. The same idea has recently been applied to multimodal learning, where images and language often have clear relations of abstraction and specificity.

MERU~\citep{desai2023hyperbolic} introduced large-scale hyperbolic vision--language pretraining by mapping CLIP-like representations to a Lorentz hyperboloid. Later studies developed the setting in several directions. Pal et al.~\citep{palcompositional} modeled compositional entailment within each modality, relating complex scenes to visual primitives and compound expressions to simpler concepts. Ibrahimi et al.~\citep{ibrahimi2024intriguing} analyzed the geometry and uncertainty encoded by hyperbolic vision--language embeddings, whereas Ramasinghe et al.~\citep{ramasinghe2024accept} replaced proximity-based cross-modal alignment with an angular objective that accommodates the modality gap. Other work scales hyperbolic learning to multimodal large language models~\citep{mandica2024hyperbolic}, filters underspecified image--text pairs using entailment-cone apertures~\citep{kim2024hype}, and applies hierarchical alignment to synthetic-caption detection~\citep{kong2024hyperbolic} or open-vocabulary segmentation~\citep{peng2025hyperclip}. Low-rank adaptation has also become a broad foundation-model adaptation family, including multimodal and continual settings~\citep{yang2025lorareview}; related work performs such adaptation directly on a hyperbolic manifold~\citep{yang2025hyplora}. HySAC~\citep{poppi2025hyperbolic} organizes safety concepts through a hyperbolic hierarchy, while hyperbolic multimodal unlearning studies the selective removal of concepts from MERU~\citep{vidal2025machine}. These works focus on pretraining, task-specific adaptation, safety, or removal; they do not study preservation of shared Lorentz relations throughout sequential multi-task acquisition.

\subsection{Continual Learning}
Continual learning seeks to acquire tasks in sequence without erasing capabilities learned previously~\citep{yu2024recent}. Existing methods typically rely on replay, regularization, or parameter isolation. Replay approaches retain representative examples and reuse them during later training, as in GEM~\citep{lopez2017gradient} and DER~\citep{buzzega2020dark}. Regularization methods such as EWC~\citep{kirkpatrick2017overcoming} penalize changes to parameters deemed important for earlier tasks. Parameter-isolation strategies instead reserve different model capacity for different tasks; examples include Progressive Neural Networks~\citep{rusu2016progressive}, PackNet~\citep{mallya2018packnet}, and Piggyback~\citep{mallya2018piggyback}.

Continual vision--language learning introduces an additional requirement: retention of cross-modal alignment together with modality-specific features~\citep{ni2023continual}. Unlike CLIP domain adaptation and generalization~\citep{li2025clip}, it must preserve a growing task history. Ni et al.~\citep{ni2023continual} described \emph{Spatial Disorder}, which combines within-modality rotation and between-modality deviation, as an important failure mode in continual CLIP adaptation. Later methods preserve zero-shot behavior through distillation~\citep{zheng2023preventing}, use mixture-of-experts adapters for task acquisition~\citep{yu2024boosting}, or apply representation-level contrastive regularization, as in C-FLAT~\cite{bian2024make}. Personalized LLM research likewise identifies lifelong updating, evolving user memories, and forgetting-resistant adaptation as core challenges~\citep{liu2025personalizedllmsurvey}. As a concrete parameter-efficient instance, PerFit models personalization through a shared low-rank representation shift and user-specific deviations, then intervenes directly in hidden states~\citep{liu2026perfit}. Related geometric work treats long-horizon memory as a nonuniform dynamical space~\citep{liu2026memorygeometry} and tailors Lorentz representations to heterogeneous federated clients~\citep{liu2026flatland}. Hyperbolic continual learning has also been studied for hierarchical image recognition~\citep{ayoughi2025continual} and multimodal few-shot class-incremental learning~\citep{doan2024streamlined}, but these lines do not enforce preservation of one shared Lorentz isometry across modalities and sequential tasks. Our setting also considers the semantic hierarchy encoded by radial coordinates and entailment cones. This motivates update constraints that account for hyperbolic geometry as well as cross-modal alignment.

\begin{journalnew}
\subsection{Projected Updates with Adaptive Optimizers}
Projection-based continual-learning methods often express retention as a constraint on a gradient or a descent direction~\citep{lopez2017gradient}. With scalar-step stochastic gradient descent, projecting the direction and projecting the resulting parameter displacement are equivalent up to the learning rate. AdamW breaks this equivalence through momentum, element-wise preconditioning, and decoupled decay~\citep{loshchilov2017decoupled}. Riemannian adaptive optimizers instead study parameters that themselves lie on a manifold and use manifold-wise gradients and updates~\citep{becigneul2019riemannian}. Our setting is different. The Lorentz head is Euclidean-parameterized, while the continual-learning constraint requires earlier outputs to remain on one shared Lorentz-isometry orbit.
\end{journalnew}

\section{Preliminaries}

\subsection{Hyperbolic Geometry}
\label{sec: pre-hyperbolic}
Hyperbolic space is a non-Euclidean manifold of constant negative curvature. We work with its Lorentz, or hyperboloid, realization because it supports stable numerical optimization~\citep{nickel2018learning,mishne2023numerical}. Appendix~\ref{apdx:additional_preliminaries} reviews the hierarchy motivation, Lorentz model, and Lorentz neural layers.

\textbf{Lorentz model.}
Let the curvature be $-1/K$ for $K>0$. A $d$-dimensional hyperbolic space can be represented by the upper sheet of a hyperboloid in $(d+1)$-dimensional Minkowski space. With metric tensor $\mathbf{G}=\operatorname{diag}(-1,1,\ldots,1)\in\mathbb{R}^{(d+1)\times(d+1)}$, the manifold is
\begin{equation}
    \mathbb{H}^d_K=\{\mathbf{x}\in\mathbb{R}^{d+1}\mid \mathbf{x}^{\top}\mathbf{G}\mathbf{x}=-K,\;x_0>0\}.
\end{equation}
A point $\mathbf{x}=[x_0;\mathbf{x}_{[1:d]}]^\top\in\mathbb{H}^d_K$ contains one time-like component $x_0$ and $d$ space-like components $\mathbf{x}_{[1:d]}$, consistent with the signature of the Lorentzian metric.

\textbf{Lorentzian product and distance.}
For $\mathbf{x},\mathbf{y}\in\mathbb{H}^d_K$, their Lorentzian inner product is
\begin{equation}
    \langle\mathbf{x},\mathbf{y}\rangle_{\mathcal{L}}
    =\mathbf{x}^{\top}\mathbf{G}\mathbf{y}
    =-x_0y_0+\sum_{i=1}^{d}x_i y_i.
\end{equation}
The associated geodesic distance is
\begin{equation}
    d_\mathcal{L}^K(\mathbf{x},\mathbf{y})
    =\sqrt{K}\operatorname{arcosh}\!\left(-\langle\mathbf{x},\mathbf{y}\rangle_{\mathcal{L}}/K\right)
    =\sqrt{K}\operatorname{arcosh}\!\left(-\mathbf{x}^{\top}\mathbf{G}\mathbf{y}/K\right).
    \label{equ:prelim_distance}
\end{equation}

\textbf{Lorentz transformation layer.}
Consider a Lorentz embedding $\mathbf{z}\in\mathbb{R}^{d+1}$ with positive time-like coordinate $z_0$ and spatial block $\mathbf{z}_s\in\mathbb{R}^{d}$. Following~\citep{chen2022fully}, a Lorentz transformation layer maps it as
\begin{equation}
f(\mathbf{W};\mathbf{z})=
\begin{bmatrix}
\sqrt{K+\|\mathbf{W}\mathbf{z}\|_2^2}\\
\mathbf{W}\mathbf{z}
\end{bmatrix},
\label{eq:lorentz_layer}
\end{equation}
where $\mathbf{W}\in\mathbb{R}^{d\times(d+1)}$. The reconstructed first coordinate ensures that the output remains on $\mathbb{H}^d_K$. We partition the parameters as
$
\mathbf{W}=[\,\mathbf{w}_0\;\;\mathbf{W}_s\,],
\quad
\mathbf{w}_0\in\mathbb{R}^{d\times1},\;
\mathbf{W}_s\in\mathbb{R}^{d\times d},
$
so that $\mathbf{w}_0$ and $\mathbf{W}_s$ act on the time-like and space-like inputs, respectively.

\subsection{Hyperbolic Multimodal Learning}
Let $\mathcal{X}^{m,m'}=\{(\mathbf{x}_i^m,\mathbf{x}_i^{m'})\}_{i=1}^{N}\subset\mathcal{X}^m\times\mathcal{X}^{m'}$ be $N$ semantically matched observations from modalities $m$ and $m'$. A multimodal model embeds them as matrices $\mathbf{Z}^{m},\mathbf{Z}^{m'}\in\mathbb{R}^{N\times(d+1)}$, whose $i$th rows $\mathbf{z}_i^m$ and $\mathbf{z}_i^{m'}$ lie on $\mathbb{H}^d_K$.

\textbf{Contrastive similarity.}
Pairwise similarity is defined by negative hyperbolic distance:
\begin{equation}
\mathbf{S}^{m\rightarrow m'}
=-d_\mathcal{L}^K(\mathbf{Z}^{m},\mathbf{Z}^{m'})
=-\sqrt{K}\operatorname{arcosh}\!\left(-\mathbf{Z}^{m}\mathbf{G}(\mathbf{Z}^{m'})^{\top}/K\right).
\label{equ:sim}
\end{equation}
Here $\operatorname{arcosh}$ is evaluated element-wise and $S_{ij}^{m\rightarrow m'}=-d_\mathcal{L}^K(\mathbf{z}_i^m,\mathbf{z}_j^{m'})$. The directional contrastive objective is
\begin{equation}
\mathcal{L}_{m\to m'}=-\frac{1}{N}\sum_{i=1}^{N}
\log\frac{\exp(S_{ii}^{m\rightarrow m'}/\tau)}
{\sum_{j=1}^{N}\exp(S_{ij}^{m\rightarrow m'}/\tau)},
\end{equation}
where $\tau$ is the temperature. Training uses the symmetric loss
\begin{equation}
\mathcal{L}_{\mathrm{contrast}}
=\frac{1}{2}\big(\mathcal{L}_{m\to m'}+\mathcal{L}_{m'\to m}\big).
\end{equation}

\textbf{Entailment cones.}
Hierarchy is represented by assigning each $\mathbf{z}_i^m\in\mathbb{H}_K^d$ an entailment cone whose aperture is
\begin{equation}
\operatorname{aper}(\mathbf{z}_i^m)
=\sin^{-1}\!\left(\frac{2\kappa\sqrt K}{\|(\mathbf{z}_i^m)_{[1:d]}\|_2}\right),
\end{equation}
with $\kappa=0.1$ controlling behavior near the origin. Points with a smaller spatial norm represent broader concepts and receive wider cones; larger norms correspond to more specific concepts and narrower cones. The exterior angle from $\mathbf{z}_i^m$ to $\mathbf{z}_i^{m'}$ is
\begin{equation}
\begin{aligned}
\gamma_i^{m,m'}&:=\langle\mathbf{z}_i^m,\mathbf{z}_i^{m'}\rangle_{\mathcal{L}}/K,\\
\operatorname{ext}(\mathbf{z}_i^m,\mathbf{z}_i^{m'})
&=\cos^{-1}\!\left(
\frac{(\mathbf{z}_i^{m'})_0+(\mathbf{z}_i^m)_0\gamma_i^{m,m'}}
{\|(\mathbf{z}_i^m)_{[1:d]}\|_2\sqrt{(\gamma_i^{m,m'})^2-1}}
\right).
\end{aligned}
\end{equation}
The entailment objective~\citep{ganea2018hyperbolic} penalizes a target that falls outside the source cone:
\begin{equation}
\mathcal{L}_{\mathrm{entail}}(\mathbf{z}_i^m,\mathbf{z}_i^{m'})
=\max\!\left(0,
\operatorname{ext}(\mathbf{z}_i^m,\mathbf{z}_i^{m'})
-\operatorname{aper}(\mathbf{z}_i^m)\right).
\end{equation}
This loss imposes the desired cross-modal partial order, for example by locating a general textual description above its visual instance in the hierarchy.

\section{Problem Formulation}
\label{sec:problem_formulation}

\subsection{Continual Hyperbolic Multimodal Learning}
We study a sequence of multimodal tasks whose representations share one hyperbolic space. At every stage, the learner absorbs the current task while retaining the within-modality, cross-modal, and hierarchical relations established by earlier tasks. Our setup follows the multimodal contrastive continual-learning protocol of~\cite{liu2025continual}.

For each modality $m\in\{1,\ldots,M\}$, a frozen pretrained encoder $\mathcal{E}_m:\mathcal{X}^m\to\mathbb{H}_K^d$ is followed at task $t$ by a trainable Lorentz transformation $T_m^{(t)}:\mathbb{H}_K^d\to\mathbb{H}_K^d$. Applied to data from the previous task, the encoder first produces
\begin{equation}
\mathbf{Z}_{t-1}^{m,*}=\mathcal{E}_m(X_{t-1}^m)\in\mathbb{R}^{N\times(d+1)},
\end{equation}
where every row lies on $\mathbb H_K^d$ and the superscript $*$ indicates a frozen pretrained representation. The stage-$t$ transformation then maps these representations through the trainable Lorentz head:
\begin{equation}
\mathbf{Z}_{t-1}^{m,t}
=T_m^{(t)}(\mathbf{Z}_{t-1}^{m,*})
=f\!\left(\mathbf{W}^{m,t};\mathbf{Z}_{t-1}^{m,*}\right)
\in\mathbb{R}^{N\times(d+1)}.
\label{equ: transformation layer}
\end{equation}
The same stage processes current-task observations as
\begin{equation}
\mathbf{Z}_{t}^{m,t}
=T_m^{(t)}\!\left(\mathcal{E}_m(X_t^m)\right)
\in\mathbb{R}^{N\times(d+1)}.
\end{equation}

Our notation $\mathbf{Z}_{q}^{m,s}$ therefore identifies the data task $q$, modality $m$, and processing stage $s$. For instance, $\mathbf{Z}_{t-1}^{m,*}$ contains frozen features from task $t-1$, whereas $\mathbf{Z}_{t-1}^{m,t}$ contains their Lorentz embeddings after the stage-$t$ transformation. Appendix~\ref{apdx:notation} summarizes all symbols.

\subsection{Geometric Stability and Plasticity}

\textbf{Stability.}
We describe retention of earlier hyperbolic knowledge through changes in the representation geometry. Because geodesic similarity is a monotone function of the Lorentzian inner product, we use three invariants\footnote{Since $\operatorname{arcosh}(\cdot)$ in Appendix~\ref{equ: distance} is strictly monotone, equal similarities are equivalent to equal Lorentzian inner products in Definition~\ref{def: inner product}.}. For every pair of distinct modalities $m,m'\in\{1,\ldots,M\}$:

\begin{infobox}
\begin{enumerate}[label=(\textbf{P\arabic*}),itemsep=0.35em,topsep=0pt,parsep=0pt]
\item \label{P: intra} \textbf{Intra-modal preservation:}\enspace
$\mathbf{Z}_{t-1}^{m,t}\mathbf{G}(\mathbf{Z}_{t-1}^{m,t})^{\top}
=\mathbf{Z}_{t-1}^{m,t-1}\mathbf{G}(\mathbf{Z}_{t-1}^{m,t-1})^{\top}$;
\item \label{P: inter} \textbf{Inter-modal preservation:}\enspace
$\mathbf{Z}_{t-1}^{m,t}\mathbf{G}(\mathbf{Z}_{t-1}^{m',t})^{\top}
=\mathbf{Z}_{t-1}^{m,t-1}\mathbf{G}(\mathbf{Z}_{t-1}^{m',t-1})^{\top}$;
\item \label{P: hierarchical} \textbf{Hierarchical preservation:}\enspace
$\|(\mathbf{z}_{t-1}^{m,t})_{[1:d]}\|_2
=\|(\mathbf{z}_{t-1}^{m,t-1})_{[1:d]}\|_2$.
\end{enumerate}
\end{infobox}

The first condition retains the relations internal to each modality. The second fixes cross-modal correspondences, and the third prevents a point from moving to a different radial level of the entailment hierarchy. Together, they protect semantic similarity, modality alignment, and specificity. Appendix~\ref{apdx:problem_statement} gives further justification.

\begin{figure}[!t]
\centering
\includegraphics[width=\linewidth]{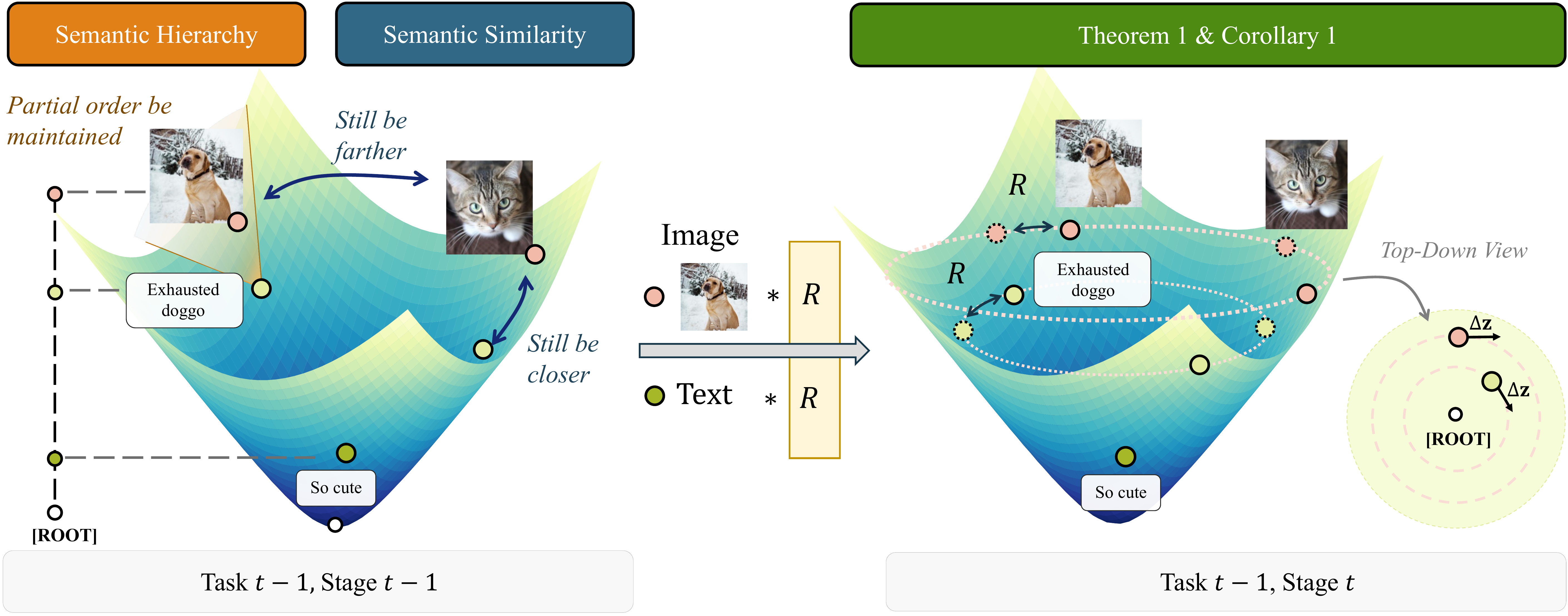}
\caption{\revisiontext{\textbf{Geometry of continual hyperbolic multimodal learning.}
Geodesic proximity encodes semantic similarity, while radial distance from the origin records specificity. Under our preservation conditions, all modalities undergo one common hyperbolic isometry. From the top-down view, first-order preservation of radial hierarchy limits a permitted update to the tangent direction.}}
\label{fig:geometric interp}
\end{figure}

\textbf{Plasticity.}
Stability also leaves enough freedom to learn the current task. At stage $t$, the embeddings $\{\mathbf{Z}_{t}^{m,t}\}_{m=1}^{M}$ fit the joint multimodal distribution $p_t^{(1,2,\ldots,M)}$. The transformation layers capture new task-specific patterns and associations without using all of their representation capacity.

\noindent\textbf{Core difficulty.}
Conditions~\textbf{\ref{P: intra}}--\textbf{\ref{P: hierarchical}} describe this balance, but applying them directly during gradient-based training is difficult. Section~\ref{sec:equivalent analysis} gives an equivalent geometric view that leads to practical constraints on parameter updates.

\section{Geometric Conditions for Admissible Updates}
\label{sec:equivalent analysis}

We next convert the preservation requirements into a geometric description of the parameter changes that are permitted during continual adaptation.

\subsection{Characterizing Preserved Representations}
\label{subsec:geometric_characterization}

The following result identifies the transformation that can relate old-task embeddings before and after a learning stage without changing the three invariants.

\begin{theorem}[Geometric Characterization of Preservation]
\label{coro:p1_p3_synergy}
Suppose the task-$t-1$ representations satisfy the full-rank condition in Appendix~\ref{apdx:proof for them1}, and the stage-to-stage correspondence follows the continuous optimizer path from the identity component. Conditions~\textbf{\ref{P: intra}}, \textbf{\ref{P: inter}}, and \textbf{\ref{P: hierarchical}} are preserved from stage $t-1$ to stage $t$ if and only if one transformation
\[
\mathbf{R}=
\begin{pmatrix}
1 & \mathbf{0}^{\top}\\
\mathbf{0} & \widetilde{\mathbf{R}}
\end{pmatrix},
\qquad \widetilde{\mathbf{R}}\in\mathrm{SO}(d),
\]
acts on every modality, so that
\[
\mathbf{Z}_{t-1}^{m,t}
=\mathbf{Z}_{t-1}^{m,t-1}\mathbf{R}^{\top},
\qquad m\in\{1,\ldots,M\}.
\]
\end{theorem}

\noindent\emph{Proof sketch.}
The intra- and inter-modal conditions preserve every required Lorentzian inner product among the task-$t-1$ embeddings. Theorem~\ref{thm:witt_multimodal} and Corollary~\ref{coro:shared L} in Appendix~\ref{apdx:proof for them1} then imply a unique Lorentz transformation $\mathbf{L}\in\mathrm{SO}^{+}(1,d)$ shared by all modalities. The forward sheet fixes the time orientation, while continuity from the identity fixes the determinant sign. Hierarchical preservation additionally fixes every spatial norm and therefore every time-like coordinate. Full rank then gives $\mathbf L^{\top}\mathbf e_0=\mathbf e_0$, leaving only the block-diagonal spatial rotation $\mathbf{R}$ shown above. Appendix~\ref{apdx:geometric_proofs} contains the complete argument.

\begin{revisionnew}
\noindent\textbf{Role of non-degeneracy.}
Full rank is the worst-case identifiability condition: it uniquely determines the shared isometry over the entire ambient space. Practical HMCL protects only the retained low-rank subspace and therefore does not require this global uniqueness; deriving the update under the maximally constrained case gives a conservative, robust admissible rule.
\end{revisionnew}

\subsection{First-Order Preservation of Hierarchy}
\label{subsec:hierarchy_preservation}

On the Lorentz manifold, the time-like coordinate is determined by the spatial block. We exploit this dependence to obtain a local constraint that prevents a continual update from changing the radial level of an old embedding.

\begin{corollary}[Updates under~\ref{P: hierarchical}]
\label{cor:timelike_stability}
Let
$\Delta\mathbf{z}=\mathbf{z}_{t-1}^{m,t}-\mathbf{z}_{t-1}^{m,t-1}$
be the change applied to an old-task embedding $\mathbf{z}_{t-1}^{m,t-1}$. If its time-like coordinate is invariant to first order, as required by~\ref{P: hierarchical}, then its spatial change obeys
\begin{equation}
\label{eq:radial_free}
\mathbf{z}_{[1{:}d]}^{\top}\Delta\mathbf{z}_{[1{:}d]}=0.
\end{equation}
\end{corollary}

\begin{infobox}
\textbf{Takeaway.}
Corollary~\ref{cor:timelike_stability} requires the spatial displacement of an old embedding to be orthogonal to its current radial direction. In other words, a first-order hierarchy-preserving change lies along the tangent direction and \revisiontext{does not introduce time-like drift and therefore avoids radial drift}.
\end{infobox}

\noindent\revisiontext{\textbf{Geometric Interpretation.}}
Figure~\ref{fig:geometric interp} illustrates two implications of the analysis. Hyperbolic distance represents semantic similarity, while radius represents hierarchical specificity. Theorem~\ref{coro:p1_p3_synergy} shows that both quantities remain unchanged across a stage transition when all modalities share the same hyperbolic isometry. From the top-down view, Corollary~\ref{cor:timelike_stability} further limits the local motion to a direction that keeps the radius fixed. A general continual-learning update is not subject to this condition and may distort the geometry even when its task loss is small.

\section{Hyperbolic Multimodal Continual Learning}

\revisiontext{The preceding results describe the admissible family in representation space. We now transfer this family to the parameters of a Lorentz transformation layer and obtain a practical training update.}

\subsection{Geometrically Admissible HMCL Updates}
\label{subsec: admissible updates}

\begin{proposition}[Admissible Parameter Changes]
\label{prop:blockwise_admissible}
Consider the Lorentz layer from Section~\ref{sec: pre-hyperbolic}, parameterized by
$\mathbf{W}^{m,t}=[\,\mathbf{w}^{m,t}_{0}\;\;\mathbf{W}^{m,t}_{s}\,]$.
Assume that Theorem~\ref{coro:p1_p3_synergy} holds at stage $t-1$. Define the applied parameter displacement as
$\Delta\mathbf{W}^{m}=\mathbf{W}^{m,t}-\mathbf{W}^{m,t-1}$.
In the infinitesimal regime $\|\Delta\mathbf{W}^{m}\|_F\to0$, the collection of modal displacements is admissible if and only if there is one skew-symmetric matrix $\boldsymbol{\Omega}$ such that, simultaneously for every modality $m$,
\begin{equation}
\mathbf{Z}^{m,*}_{t-1}
\big(\Delta\mathbf{W}^{m}\big)^{\top}
=
\mathbf{Z}^{m,t-1}_{t-1}[1\!:\!d]
\boldsymbol{\Omega}^{\top},
\qquad
\boldsymbol{\Omega}^{\top}=-\boldsymbol{\Omega}.
\label{eq:prop_exact_admissible}
\end{equation}
\end{proposition}

The left-hand side of~\eqref{eq:prop_exact_admissible} is the first-order change of an old spatial representation. The right-hand side is the tangent motion of the rotation $\exp(\boldsymbol\Omega)$. Crucially, the same $\boldsymbol\Omega$ appears for every modality. The condition therefore permits a common change of spatial coordinates, but excludes independent modal rotations that would alter cross-modal relations.

Equation~\eqref{eq:prop_exact_admissible} includes both parameter blocks. Under the subspace regularity condition in Appendix~\ref{appendix: closest_admissible}, the same output motion has a representative whose time-input column remains fixed; Appendix~\ref{appendix: blockwise} gives the precise condition. For every modality, we use the blockwise form
\begin{equation}
\begin{aligned}
\Delta\mathbf w_0^m&=\mathbf0,\\
\mathbf Z_{t-1}^{m,*}[1{:}d](\Delta\mathbf W_s^m)^{\top}
&=\mathbf Z_{t-1}^{m,t-1}[1{:}d]\boldsymbol\Omega^{\top}.
\end{aligned}
\label{eq:prop_blockwise_admissible_spatial}
\end{equation}

\begin{corollary}[HMCL-CA: Closest-Admissible Update]
\label{cor:closest_admissible}
Given candidate parameter displacements
$\{\boldsymbol\delta^m=[\,\boldsymbol\delta_0^m\;\;\boldsymbol\delta_s^m\,]\}_{m=1}^{M}$,
where $\boldsymbol\delta_0^m\in\mathbb R^{d\times1}$ is the column acting on
the time-like input coordinate and
$\boldsymbol\delta_s^m\in\mathbb R^{d\times d}$ is the block acting on the
spatial input coordinates, HMCL-CA jointly selects the admissible
displacements and their common rotation by solving
\begin{equation}
\begin{aligned}
\underset{\{\Delta\mathbf W^m\},\,\boldsymbol\Omega}{\operatorname{minimize}}
\quad&\frac12\sum_{m=1}^{M}
\|\Delta\mathbf W^m-\boldsymbol\delta^m\|_F^2\\
\text{subject to}\quad&
\boldsymbol\Omega^{\top}=-\boldsymbol\Omega,
\quad \Delta\mathbf w_0^m=\mathbf0,\\
&
\mathbf Z_{t-1}^{m,*}[1{:}d](\Delta\mathbf W_s^m)^{\top}
=\mathbf Z_{t-1}^{m,t-1}[1{:}d]\boldsymbol\Omega^{\top},\\
&m=1,\ldots,M.
\end{aligned}
\label{eq:hmcl_closest_admissible_problem}
\end{equation}
We denote the solution by $\{\boldsymbol\delta^{\mathrm{CA},m}\}_{m=1}^{M}$ and $\boldsymbol\Omega^{\star}$. Under the regularity condition in Appendix~\ref{appendix: closest_admissible}, the parameter displacements are unique and are obtained from one Lyapunov equation for $\boldsymbol\Omega^{\star}$. Thus, HMCL-CA retains the common rotation that best matches all modal candidates, rather than solving a separate rotation for each modality.
\end{corollary}

\begin{corollary}[HMCL-MR: Minimal-Rotation Update]
\label{cor:canonical_admissible}
Selecting $\boldsymbol\Omega=\mathbf0$ in~\eqref{eq:hmcl_closest_admissible_problem} gives, for each modality,
\begin{equation}
\mathbf{Z}^{m,*}_{t-1,s}
\big(\boldsymbol\delta^{\mathrm{MR},m}_{s}\big)^{\top}=\mathbf{0},
\qquad
\boldsymbol\delta^{\mathrm{MR},m}_{0}=\mathbf{0},
\label{eq:canonical_constraints}
\end{equation}
where $\mathbf{Z}^{m,*}_{t-1,s}=\mathbf{Z}^{m,*}_{t-1}[1:d]$.
For an orthogonal projector $\mathbf P_{t-1}$ onto the protected old-feature subspace, the unique closest solution is
\begin{equation}
\boldsymbol\delta^{\mathrm{MR},m}_{s}
=\boldsymbol\delta^m_s
\big(\mathbf{I}-\mathbf{P}_{t-1}\big).
\label{eq:canonical_projection}
\end{equation}
Here, $\mathbf P_{t-1}$ projects onto the protected old-feature subspace. This is the closest candidate displacement that induces no first-order motion on that subspace.
\end{corollary}

HMCL-MR is therefore the $\boldsymbol\Omega=\mathbf0$ member of the same closest-admissible problem. Appendix~\ref{appendix: blockwise} proves Proposition~\ref{prop:blockwise_admissible}; Appendices~\ref{appendix: closest_admissible} and~\ref{appendix: canonical} give the corresponding derivations.

In summary, hyperbolic preservation determines one admissible family: HMCL-CA selects its closest shared-rotation update, and HMCL-MR selects the zero-rotation member.

\begin{journalnew}
\subsection{Stepwise Correction and Task Anchoring}
\label{subsec:optimizer_consistent_hmcl}

\revisiontext{We follow the standard training paradigm of the hyperbolic multimodal backbones, which optimizes their Euclidean-parameterized Lorentz heads with AdamW. Because AdamW's moment estimation and coordinate-wise scaling make its realized parameter displacement different from the raw gradient, HMCL applies \emph{post-AdamW projection} to that realized displacement. It then uses \emph{task anchoring} to contract the corrected parameters toward the current task start. During task $t$,}
$J$ denotes the total number of optimizer steps and
$k\in\{0,\ldots,J-1\}$ indexes them, with
$\mathbf W^{m,t}_0=\mathbf W^{m,t-1}$ and
$\mathbf W^{m,t}_J=\mathbf W^{m,t}$. A first subscript $k$ denotes the
within-task optimizer step; a second subscript after a comma denotes an input
block, with $0$ for the time-like input column and $s$ for the spatial input
block. Thus, for $r\in\{\mathrm{opt},\star\}$,
\[
\mathbf W_k^{m,t}
=[\,\mathbf w_{k,0}^{m,t}\;\;\mathbf W_{k,s}^{m,t}\,],
\qquad
\boldsymbol\delta_k^{r,m,t}
=[\,\boldsymbol\delta_{k,0}^{r,m,t}\;\;
\boldsymbol\delta_{k,s}^{r,m,t}\,].
\]
In particular, the single subscript in $\mathbf W_0^{m,t}$ identifies the
full parameter matrix at optimizer step $0$; it is not a block label. With
the AdamW state left implicit, one optimizer step produces
\begin{equation*}
\begin{aligned}
\widetilde{\mathbf W}^{m,t}_{k+1}
&=\operatorname{Opt}_k^m(\mathbf W^{m,t}_{k},\mathbf g^{m,t}_{k}),\\
\boldsymbol\delta^{\mathrm{opt},m,t}_{k}
&=\widetilde{\mathbf W}^{m,t}_{k+1}-\mathbf W^{m,t}_{k}.
\end{aligned}
\end{equation*}
Here $\mathbf g_k^{m,t}$ is the raw gradient, whereas
$\boldsymbol\delta_k^{\mathrm{opt},m,t}$ is the realized parameter
displacement proposed by AdamW before HMCL correction. It already includes
the learning rate, moment estimation, coordinate-wise preconditioning, and
decoupled weight decay; it is
not a gradient.

To state the geometric correction at the level of this proposed step, let
$\boldsymbol\xi_k^{m,t}
=[\,\boldsymbol\xi_{k,0}^{m,t}\;\;\boldsymbol\xi_{k,s}^{m,t}\,]$ denote an
arbitrary parameter displacement at within-task step $k$, with the same
time-input/spatial-input partition as $\mathbf W_k^{m,t}$. For any collection
$\{\boldsymbol\xi_k^{m,t}\}_{m=1}^{M}$, define its Lorentz defect by
\begin{equation*}
\mathcal E_{\mathcal L,t-1}(\{\boldsymbol\xi_k^{m,t}\})
=\min_{\boldsymbol\Omega^{\top}=-\boldsymbol\Omega}
\sum_{m=1}^{M}
\left\|
\begin{aligned}
&\mathbf Z_{t-1}^{m,*}[1{:}d](\boldsymbol\xi_{k,s}^{m,t})^{\top}\\[-2pt]
&-\mathbf Z_{t-1}^{m,t-1}[1{:}d]\boldsymbol\Omega^{\top}
\end{aligned}
\right\|_F^2
+\sum_{m=1}^{M}
\|\boldsymbol\xi_{k,0}^{m,t}\|_2^2.
\end{equation*}
Proposition~\ref{prop:blockwise_admissible} gives
$\mathcal E_{\mathcal L,t-1}=0$ exactly for a fixed time-input column and one
spatial Lorentz rotation shared across modalities. Let
$\mathcal P^{\mathcal L}_{t-1}$ be the Frobenius closest-point map onto this
zero-defect family, denoted by $\mathcal A_{t-1}$. At optimizer step $k$, the defect is evaluated at
$\boldsymbol\xi_k^{m,t}=\boldsymbol\delta_k^{\mathrm{opt},m,t}$, whose two blocks
are $\boldsymbol\delta_{k,0}^{\mathrm{opt},m,t}$ and
$\boldsymbol\delta_{k,s}^{\mathrm{opt},m,t}$. We collect the modal
displacements as
$\boldsymbol\delta_k^{\mathrm{opt},t}
=(\boldsymbol\delta_k^{\mathrm{opt},m,t})_{m=1}^{M}$, with analogous notation
for $\boldsymbol\delta_k^{\star,t}$. For $0\leq\beta<1$, the combined update is
\begin{equation}
\color{revisionred}
\begin{aligned}
\boldsymbol\delta_k^{\star,t}
&=\mathcal P^{\mathcal L}_{t-1}
(\boldsymbol\delta_k^{\mathrm{opt},t}),\\
\mathbf W^{m,t}_{k+1}
&=\beta\mathbf W^{m,t}_{0}+(1-\beta)
\left(\mathbf W^{m,t}_{k}+\boldsymbol\delta^{\star,m,t}_{k}\right).
\end{aligned}
\label{eq:optimizer_consistent_anchored_update}
\end{equation}
For a modal tuple $\mathcal U=(\mathbf U^m)_{m=1}^{M}$, write
$\|\mathcal U\|_{\oplus F}:=(\sum_m\|\mathbf U^m\|_F^2)^{1/2}$.

\begin{theorem}[Anchored Admissibility and Bounded Accumulation]
\label{thm:optimizer_consistent_closest_update}
For the update in~\eqref{eq:optimizer_consistent_anchored_update} on a
protected representation system fixed throughout task $t$, the cumulative
displacement
$\mathbf A_J^t=(\mathbf A_J^m)_{m=1}^{M}$, where
$\mathbf A_J^m=\mathbf W_J^{m,t}-\mathbf W_0^{m,t}$, satisfies
$\{\mathbf A_J^m\}_{m=1}^{M}\in\mathcal A_{t-1}$ and
$\mathcal E_{\mathcal L,t-1}(\{\mathbf A_J^m\})=0$. Thus, the protected
representations undergo one shared first-order Lorentz rotation across all
modalities. The projection is non-expansive in the joint Frobenius norm. If
$\|\boldsymbol\delta_k^{\mathrm{opt},t}\|_{\oplus F}\leq u_{\max}$ for every $k$,
then task anchoring with $0<\beta<1$ also gives the uniform bound
$\|\mathbf A_J^t\|_{\oplus F}\leq(1-\beta)u_{\max}/\beta$.
\end{theorem}

\begin{infobox}
\begin{revisionnew}
\textbf{Takeaway.}
HMCL separates stability from plasticity at each optimizer step. Projection removes only the part of AdamW's realized displacement that violates the shared Lorentz-isometry constraint, leaving the admissible component free to learn the current task. Task anchoring then limits cumulative drift without leaving this family.
\end{revisionnew}
\end{infobox}
\end{journalnew}

\section{Experiments}
\label{sec:experiments}

Our evaluation is organized around four questions concerning continual adaptation of hyperbolic multimodal models:
\begin{itemize}
    \item \revisiontext{\textbf{RQ1:} Does HMCL improve the stability--plasticity balance, measured by final performance and backward transfer, across hyperbolic multimodal backbones?}
    \item \textbf{RQ2:} Does HMCL reduce radial, angular, cross-modal, and paired-distance drift on previously learned tasks?
    \item \textbf{RQ3:} Does the resulting representation preserve semantic hierarchies in quantitative and qualitative tests?
    \item \journaltext{\textbf{RQ4:} Within the unified HMCL update, what are the separate effects of correction placement and task anchoring?}
\end{itemize}
Accordingly, Section~\ref{sec:exp_rq1} addresses RQ1, Section~\ref{sec:exp_drift} addresses RQ2, Sections~\ref{subsec:journal_extended_results} and~\ref{sec:exp_case_study} address RQ3, and Section~\ref{subsec:hmcl_component_analysis} addresses RQ4.

\subsection{Experimental Setup}

\begin{revisionnew}
\textbf{Task Stream and Datasets.}
The continual stream combines 14 multimodal classification tasks with the COCO~\citep{lin2014microsoft} and Flickr30K~\citep{young2014image} cross-modal retrieval tasks. ImageNet-1K~\citep{deng2009imagenet} is inserted after Caltech-101, yielding 16 tasks in total. Non-replay methods never revisit earlier-task data, whereas replay baselines access it only through a bounded buffer. Appendices~\ref{apdx:datasets} and~\ref{apdx:robustness_drift} provide the complete dataset statistics, canonical order, and alternative-order protocols.

\textbf{Metrics.}
For retrieval, the implementation records image-to-text and text-to-image Recall@$k$ for $k\in\{1,5,10\}$. The main comparison uses symmetric R@5, i.e., the mean of image-to-text and text-to-image R@5, for both retrieval tasks. Classification performance is top-1 accuracy. Overall is the unweighted mean of the 14 final classification accuracies and the two final symmetric retrieval scores. We define signed backward transfer as $\mathrm{BWT}=\mathrm{Performance}_{\mathrm{final}}-\mathrm{Performance}_{\mathrm{right\ after\ task}}$; thus, a negative value denotes forgetting. The main table reports this quantity separately for classification, retrieval, and all tasks. Reported entries are means and sample standard deviations over five paired seeds.

\textbf{Baselines.}
Our comparison covers the principal families of continual-learning strategies under the same hyperbolic objective: ordinary sequential fine-tuning (Vanilla), parameter regularization (EWC), replay-based constrained optimization (GEM), vision--language distillation with flatness-aware optimization (C-FLAT), and dual-sided null-space projection (DNS)~\citep{ni2023continual,lopez2017gradient,bian2024make,liu2025continual}. This breadth tests HMCL against methods that preserve parameters, examples, representations, or update subspaces; none of them explicitly preserves a shared Lorentz isometry across modalities.

\textbf{Implementation Details.}
We follow each backbone's hyperbolic multimodal training paradigm: the pretrained encoder is frozen, continual adaptation is confined to a bias-free linear Lorentz head, and AdamW~\cite{loshchilov2017decoupled} optimizes the common objective. Classification and retrieval tasks use 10 and 15 epochs with learning rates $5\times10^{-4}$ and $5\times10^{-5}$, respectively; all use batch size 1024, curvature $-0.1$, and one accumulated AdamW step per epoch. Main HMCL runs apply post-AdamW correction followed by task anchoring. Every method shares the task order, cached 512-dimensional features, schedule, optimizer-step budget, and five paired seeds. Appendix~\ref{apdx:reproducibility_notes} gives the complete configurations and baseline memory budgets, while Appendices~\ref{apdx:pullback_sensitivity}, \ref{apdx:adamw_weight_decay}, and~\ref{apdx:clean_training_time} report sensitivity and timing details.
\end{revisionnew}

\subsection{Main Results (RQ1)}
\label{sec:exp_rq1}

To answer RQ1, we compare final task performance and signed backward transfer under the matched 16-task protocol across all three hyperbolic multimodal backbones.

\begin{table}[t]
\centering
\small
\setlength{\tabcolsep}{4.2pt}
\renewcommand{\arraystretch}{1.12}
\caption{Continual-learning results on the unified 16-task stream (mean $\pm$ sample standard deviation over five paired seeds). Classification is top-1 accuracy, retrieval is symmetric R@5, and BWT is final minus immediate performance. Light-blue rows denote the complete post-AdamW, task-pullback HMCL variants; boldface marks the best result in each backbone block. The $\Delta$ rows give $\mathrm{HMCL}_{\mathrm{CA}}$'s relative improvement over Vanilla, using the reduction in negative magnitude for BWT. All methods share the task order, cached features, training schedule, and optimizer-step budget.}
\label{tab:main_results16}
\begin{adjustbox}{max width=\textwidth}
\begin{tabular}{l l c c c c c c}
\toprule
& & \multicolumn{2}{c}{\textbf{Classification}} & \multicolumn{2}{c}{\textbf{Retrieval}} & \multicolumn{2}{c}{\textbf{Overall}} \\
\cmidrule(lr){3-4}\cmidrule(lr){5-6}\cmidrule(lr){7-8}
\textbf{Backbone} & \textbf{Method} & \textbf{Acc} $\uparrow$ & \(\mathbf{BWT}_{\mathrm{A}}\uparrow\) & \textbf{R@5} $\uparrow$ & \(\mathbf{BWT}_{\mathrm{R}5}\uparrow\) & \textbf{Performance} $\uparrow$ & \textbf{BWT} $\uparrow$ \\
\midrule
\multirow{8}{*}{MERU-L} & Vanilla & $40.251\pm0.202$ & $-5.461\pm0.180$ & $29.625\pm0.110$ & $-2.479\pm0.123$ & $38.922\pm0.165$ & $-5.088\pm0.145$ \\
 & EWC & $40.321\pm0.187$ & $-5.467\pm0.213$ & $30.215\pm0.146$ & $-2.096\pm0.158$ & $39.058\pm0.153$ & $-5.046\pm0.174$ \\
 & GEM & $41.613\pm0.288$ & $-3.742\pm0.284$ & $29.220\pm0.299$ & $-2.306\pm0.418$ & $40.064\pm0.276$ & $-3.563\pm0.276$ \\
 & C-FLAT & $40.450\pm0.171$ & $-3.945\pm0.108$ & $29.446\pm0.174$ & $-2.032\pm0.115$ & $39.075\pm0.143$ & $-3.706\pm0.086$ \\
 & DNS & $40.904\pm0.149$ & $-4.912\pm0.174$ & $29.716\pm0.170$ & $-2.455\pm0.100$ & $39.505\pm0.126$ & $-4.605\pm0.149$ \\
\cmidrule(lr){2-8}
 & \cellcolor{myblue!20}HMCL$_{\mathrm{MR}}$ & \cellcolor{myblue!20}$43.891\pm0.085$ & \cellcolor{myblue!20}$-1.316\pm0.086$ & \cellcolor{myblue!20}$\mathbf{36.300\pm0.114}$ & \cellcolor{myblue!20}$\mathbf{-1.921\pm0.057}$ & \cellcolor{myblue!20}$42.942\pm0.070$ & \cellcolor{myblue!20}$\mathbf{-1.392\pm0.070}$ \\
 & \cellcolor{myblue!20}HMCL$_{\mathrm{CA}}$ & \cellcolor{myblue!20}$\mathbf{43.987\pm0.087}$ & \cellcolor{myblue!20}$\mathbf{-1.305\pm0.090}$ & \cellcolor{myblue!20}$36.011\pm0.136$ & \cellcolor{myblue!20}$-2.046\pm0.099$ & \cellcolor{myblue!20}$\mathbf{42.990\pm0.073}$ & \cellcolor{myblue!20}$-1.398\pm0.078$ \\
 & \textcolor{darkgreen}{$\boldsymbol{\Delta}$ vs. Vanilla} & \textcolor{darkgreen}{\textbf{+9.3\%}} & \textcolor{darkgreen}{\textbf{+76.1\%}} & \textcolor{darkgreen}{\textbf{+21.6\%}} & \textcolor{darkgreen}{\textbf{+17.5\%}} & \textcolor{darkgreen}{\textbf{+10.5\%}} & \textcolor{darkgreen}{\textbf{+72.5\%}} \\
\midrule
\multirow{8}{*}{MERU-B} & Vanilla & $43.326\pm0.243$ & $-6.617\pm0.244$ & $30.535\pm0.093$ & $-4.610\pm0.120$ & $41.727\pm0.222$ & $-6.366\pm0.218$ \\
 & EWC & $43.472\pm0.210$ & $-6.490\pm0.217$ & $30.965\pm0.086$ & $-4.291\pm0.172$ & $41.909\pm0.190$ & $-6.215\pm0.189$ \\
 & GEM & $44.880\pm0.272$ & $-3.803\pm0.093$ & $32.010\pm0.373$ & $-3.009\pm0.238$ & $43.271\pm0.228$ & $-3.704\pm0.056$ \\
 & C-FLAT & $42.561\pm0.236$ & $-4.700\pm0.146$ & $31.459\pm0.161$ & $-4.300\pm0.237$ & $41.174\pm0.211$ & $-4.650\pm0.124$ \\
 & DNS & $43.769\pm0.163$ & $-6.265\pm0.225$ & $31.423\pm0.073$ & $-4.138\pm0.193$ & $42.226\pm0.151$ & $-5.999\pm0.217$ \\
\cmidrule(lr){2-8}
 & \cellcolor{myblue!20}HMCL$_{\mathrm{MR}}$ & \cellcolor{myblue!20}$45.736\pm0.042$ & \cellcolor{myblue!20}$-3.448\pm0.112$ & \cellcolor{myblue!20}$36.456\pm0.090$ & \cellcolor{myblue!20}$-1.392\pm0.162$ & \cellcolor{myblue!20}$44.576\pm0.036$ & \cellcolor{myblue!20}$-3.191\pm0.089$ \\
 & \cellcolor{myblue!20}HMCL$_{\mathrm{CA}}$ & \cellcolor{myblue!20}$\mathbf{45.746\pm0.078}$ & \cellcolor{myblue!20}$\mathbf{-3.159\pm0.134}$ & \cellcolor{myblue!20}$\mathbf{36.522\pm0.083}$ & \cellcolor{myblue!20}$\mathbf{-1.229\pm0.116}$ & \cellcolor{myblue!20}$\mathbf{44.593\pm0.071}$ & \cellcolor{myblue!20}$\mathbf{-2.918\pm0.111}$ \\
 & \textcolor{darkgreen}{$\boldsymbol{\Delta}$ vs. Vanilla} & \textcolor{darkgreen}{\textbf{+5.6\%}} & \textcolor{darkgreen}{\textbf{+52.3\%}} & \textcolor{darkgreen}{\textbf{+19.6\%}} & \textcolor{darkgreen}{\textbf{+73.3\%}} & \textcolor{darkgreen}{\textbf{+6.9\%}} & \textcolor{darkgreen}{\textbf{+54.2\%}} \\
\midrule
\multirow{8}{*}{HyCoCLIP-B} & Vanilla & $40.988\pm0.160$ & $-4.421\pm0.097$ & $62.034\pm0.182$ & $-2.544\pm0.172$ & $43.618\pm0.152$ & $-4.186\pm0.087$ \\
 & EWC & $41.121\pm0.129$ & $-4.385\pm0.122$ & $63.367\pm0.148$ & $-1.423\pm0.091$ & $43.901\pm0.118$ & $-4.014\pm0.107$ \\
 & GEM & $42.111\pm0.318$ & $-3.236\pm0.353$ & $62.586\pm0.125$ & $-1.449\pm0.121$ & $44.670\pm0.268$ & $-3.013\pm0.315$ \\
 & C-FLAT & $40.951\pm0.116$ & $-3.699\pm0.097$ & $62.010\pm0.042$ & $-1.839\pm0.078$ & $43.583\pm0.104$ & $-3.466\pm0.093$ \\
 & DNS & $41.356\pm0.105$ & $-4.065\pm0.076$ & $62.536\pm0.189$ & $-2.068\pm0.169$ & $44.004\pm0.108$ & $-3.815\pm0.077$ \\
\cmidrule(lr){2-8}
 & \cellcolor{myblue!20}HMCL$_{\mathrm{MR}}$ & \cellcolor{myblue!20}$44.828\pm0.064$ & \cellcolor{myblue!20}$-0.635\pm0.069$ & \cellcolor{myblue!20}$\mathbf{72.801\pm0.087}$ & \cellcolor{myblue!20}$\mathbf{-0.879\pm0.091}$ & \cellcolor{myblue!20}$48.325\pm0.053$ & \cellcolor{myblue!20}$-0.666\pm0.069$ \\
 & \cellcolor{myblue!20}HMCL$_{\mathrm{CA}}$ & \cellcolor{myblue!20}$\mathbf{45.108\pm0.031}$ & \cellcolor{myblue!20}$\mathbf{-0.388\pm0.061}$ & \cellcolor{myblue!20}$72.772\pm0.073$ & \cellcolor{myblue!20}$-0.937\pm0.054$ & \cellcolor{myblue!20}$\mathbf{48.566\pm0.024}$ & \cellcolor{myblue!20}$\mathbf{-0.456\pm0.051}$ \\
 & \textcolor{darkgreen}{$\boldsymbol{\Delta}$ vs. Vanilla} & \textcolor{darkgreen}{\textbf{+10.1\%}} & \textcolor{darkgreen}{\textbf{+91.2\%}} & \textcolor{darkgreen}{\textbf{+17.3\%}} & \textcolor{darkgreen}{\textbf{+63.2\%}} & \textcolor{darkgreen}{\textbf{+11.3\%}} & \textcolor{darkgreen}{\textbf{+89.1\%}} \\
\bottomrule
\end{tabular}
\end{adjustbox}
\end{table}

\begin{revisionnew}
\textbf{HMCL gives the strongest stability--plasticity balance across backbones.}
Table~\ref{tab:main_results16} shows that both complete HMCL variants outperform every external baseline on all three backbones in final classification and retrieval performance as well as in the Overall score. Their BWT values also move substantially toward zero, so higher final scores accompany stronger retention rather than sacrificing earlier tasks. $\mathrm{HMCL}_{\mathrm{CA}}$ gives the highest Overall score in every backbone block and consistently exceeds task-balanced GEM. DNS improves only modestly over Vanilla, suggesting that Euclidean null-space protection alone does not fully capture the constraints of hyperbolic representations. The agreement across classification and retrieval is informative because the former tests category discrimination, whereas the latter depends on preserving correspondence between modalities. Improvements in both settings are consistent with protecting their shared representation geometry, rather than benefiting only one prediction rule. Appendix~\ref{apdx:imagenet_wordnet_protocol} provides the complete per-task results.

\textbf{CA and MR expose complementary members of the admissible family.}
On the unified 16-task stream, CA generally gives higher classification and Overall scores, while MR is stronger on retrieval for MERU-L and HyCoCLIP-B and gives the best Overall BWT on MERU-L. This division is not universal: in the modality-extended stream below, CA also gives the stronger retrieval result on all three backbones. The evidence therefore supports a task-dependent stability--plasticity trade-off within one admissible family, rather than assigning a fixed empirical role to either variant.
\end{revisionnew}

\begin{journalnew}
\begin{figure}[!t]
\centering
\includegraphics[width=0.99\linewidth]{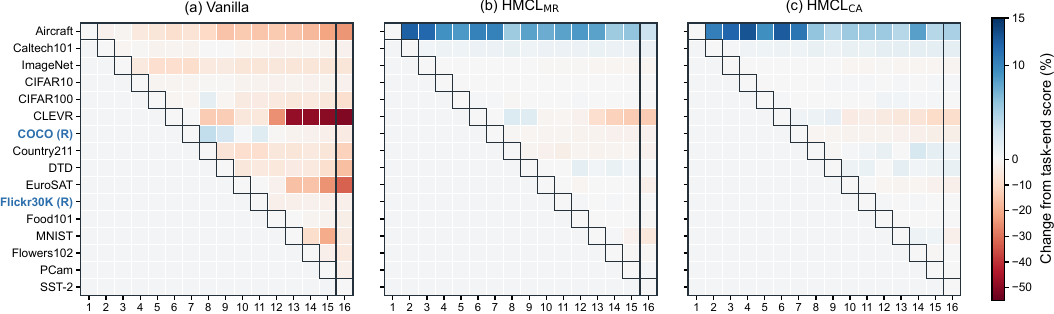}
\caption{Stage-wise relative task-performance change for HyCoCLIP-B on seed 42 in the unified 16-task stream. Columns correspond to the completed training stage $t$, and rows correspond to evaluated task $i$ in stream order. Each observed cell reports $100(P_{i,t}-P_{i,i})/P_{i,i}$, where $P_{i,i}$ is the score immediately after task $i$ is learned; thus, the outlined diagonal is zero, red denotes relative degradation, and blue denotes improvement. The gray upper triangle contains tasks not yet introduced. All panels share one color scale, the outlined rightmost column gives the final-stage changes, and (R) marks retrieval tasks evaluated with symmetric R@5; the remaining tasks use top-1 accuracy.}
\label{fig:stage_task_relative_change}
\end{figure}

\begin{revisionnew}
\textbf{HMCL remains stable throughout the task stream.}
Figure~\ref{fig:stage_task_relative_change} normalizes each task by its own task-end score. On the displayed paired seed, Vanilla exhibits pronounced degradation on earlier tasks, whereas both HMCL variants keep the trajectories closer to their task-end reference. The contrast is visible across intermediate stages as well as at the final checkpoint, suggesting that the retention gains reflect sustained protection during adaptation rather than recovery only at the end of the stream. These old-task trajectories provide a stage-wise view of the five-seed BWT gains in Table~\ref{tab:main_results16}; the displayed seed illustrates the temporal pattern rather than replacing the multi-seed comparison.
\end{revisionnew}
\end{journalnew}

\begin{journalnew}

\begin{revisionnew}
\Needspace{3\baselineskip}
\textbf{The gains persist under alternative task orders.}
Appendix~\ref{apdx:robustness_drift} evaluates two shuffled HyCoCLIP-B streams with the main-table configurations fixed and no order-specific retuning. Across the paired seeds, both HMCL variants retain higher final classification, retrieval, and Overall scores and less negative BWT than Vanilla under either permutation. Their relative ranking changes with the order: MR leads in Overall on one permutation, whereas CA leads on the other. These results support robustness beyond the canonical sequence while reinforcing that the preferred admissible representative depends on the task stream.

\Needspace{22\baselineskip}
\begin{wraptable}{r}{0.58\textwidth}
\vspace{-0.75\baselineskip}
\color{journalblue}
\centering
\scriptsize
\setlength{\tabcolsep}{1.6pt}
\renewcommand{\arraystretch}{1.02}
\captionsetup{font=footnotesize}
\caption{Robustness to new modalities in a separate protocol with one audio and one thermal retrieval task (mean $\pm$ standard deviation over three paired seeds). Cls. is top-1 accuracy, Ret. is symmetric R@5, Overall averages all final task scores, and BWT is final minus immediate performance. Both HMCL variants use after-AdamW correction and step pullback ($\beta=0.015$). Boldface marks the best result per backbone and metric.}
\label{tab:backbone_robustness17}
\begin{adjustbox}{max width=\linewidth}
\begin{tabular}{@{}llccc@{}}
\toprule
\textbf{Backbone} & \textbf{Metric} & \textbf{Vanilla} & \textbf{HMCL$_{\rm MR}$} & \textbf{HMCL$_{\rm CA}$} \\
\midrule
\multirow{4}{*}{MERU-L}
 & Cls. & $42.514\pm0.092$ & $\mathbf{43.894\pm0.130}$ & $43.188\pm0.046$ \\
 & Ret. & $16.221\pm0.048$ & $16.278\pm0.105$ & $\mathbf{18.224\pm0.077}$ \\
 & Overall & $36.327\pm0.064$ & $\mathbf{37.396\pm0.099}$ & $37.315\pm0.026$ \\
 & BWT & $-4.876\pm0.054$ & $-1.650\pm0.038$ & $\mathbf{-1.110\pm0.036}$ \\
\midrule
\multirow{4}{*}{MERU-B}
 & Cls. & $44.156\pm0.190$ & $\mathbf{46.646\pm0.213}$ & $45.815\pm0.050$ \\
 & Ret. & $14.868\pm0.131$ & $17.000\pm0.076$ & $\mathbf{18.482\pm0.053}$ \\
 & Overall & $37.265\pm0.116$ & $\mathbf{39.670\pm0.145}$ & $39.384\pm0.028$ \\
 & BWT & $-7.293\pm0.191$ & $-4.354\pm0.120$ & $\mathbf{-2.871\pm0.067}$ \\
\midrule
\multirow{4}{*}{HyCoCLIP-B}
 & Cls. & $42.869\pm0.154$ & $44.563\pm0.180$ & $\mathbf{45.348\pm0.077}$ \\
 & Ret. & $32.970\pm0.123$ & $33.506\pm0.123$ & $\mathbf{35.407\pm0.081}$ \\
 & Overall & $40.540\pm0.143$ & $41.961\pm0.155$ & $\mathbf{43.009\pm0.072}$ \\
 & BWT & $-3.859\pm0.062$ & $-2.603\pm0.056$ & $\mathbf{-1.658\pm0.017}$ \\
\bottomrule
\end{tabular}
\end{adjustbox}
\vspace{-0.5\baselineskip}
\end{wraptable}

\textbf{HMCL remains effective when the stream introduces new modalities.}
Table~\ref{tab:backbone_robustness17} reports a separate protocol that adds one audio and one thermal retrieval task to the original stream. Both complete HMCL variants improve the Overall score and BWT over matched Vanilla training on every backbone. CA consistently improves retrieval R@5 while substantially reducing forgetting across the backbones. MR gives a slightly higher Overall score on the two MERU backbones, whereas CA is strongest on HyCoCLIP-B. The result shows that the shared geometric constraint remains useful when previously unseen modality types enter the stream. This extension is a distinct stress test: adaptation now accommodates additional cross-modal relations while retaining those learned earlier. The consistent retrieval gains suggest that geometric protection remains useful beyond the original image--text setting, although the changing MR--CA ranking cautions against treating either realization as uniformly preferable. Separate evaluation of earlier image--text tasks distinguishes retention of existing correspondences from new-modality gains.
\end{revisionnew}

\begin{revisionnew}
\Needspace{3\baselineskip}
\textbf{The retrieval gains persist at fine granularity.}
Figure~\ref{fig:retrieval_finegrained17} shows final COCO and Flickr30K retrieval across backbones, directions, and recall cutoffs; Appendix~\ref{apdx:finegrained_retrieval17} gives the complete numerical results. Under the same modality-extension protocol, HMCL$_{\mathrm{CA}}$ exceeds Vanilla and gives the highest score in all 36 backbone--dataset--direction--cutoff comparisons. Its advantage therefore spans both retrieval directions and every recall cutoff, not only symmetric R@5.
\end{revisionnew}

\begin{figure}[!t]
\centering
\includegraphics[width=0.99\linewidth]{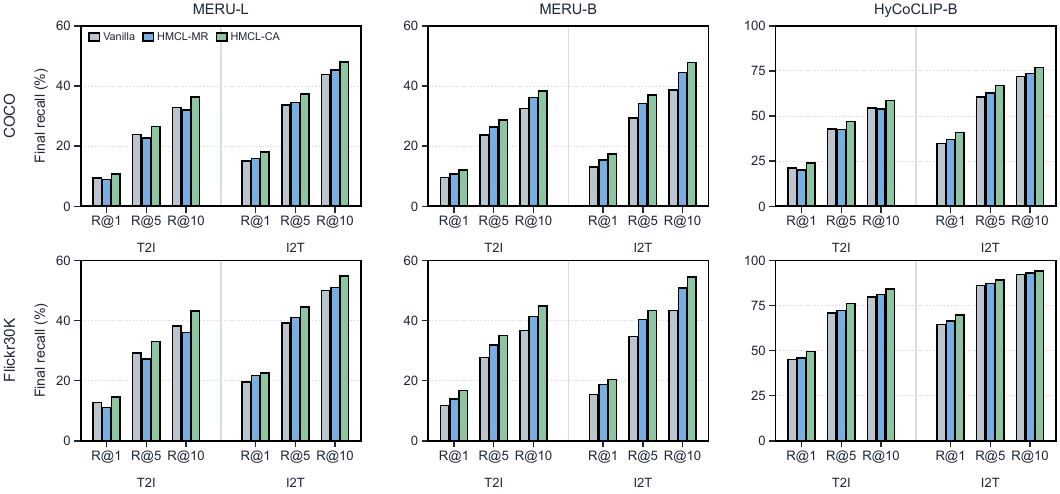}
\caption{\revisiontext{Fine-grained final retrieval performance on COCO and Flickr30K in the separate stream augmented with audio and thermal tasks. Each panel reports the mean over three paired seeds for one backbone; T2I and I2T denote text-to-image and image-to-text Recall@$\{1,5,10\}$, respectively. Gray, blue, and green bars denote Vanilla, HMCL$_{\mathrm{MR}}$, and HMCL$_{\mathrm{CA}}$. Appendix~\ref{apdx:finegrained_retrieval17} reports exact means and standard deviations. Higher values indicate better retrieval.}}
\label{fig:retrieval_finegrained17}
\end{figure}
\end{journalnew}

\subsection{Quantitative Representation Drift (RQ2)}
\label{sec:exp_drift}

\begin{figure}[!t]
\centering
\includegraphics[width=0.99\linewidth]{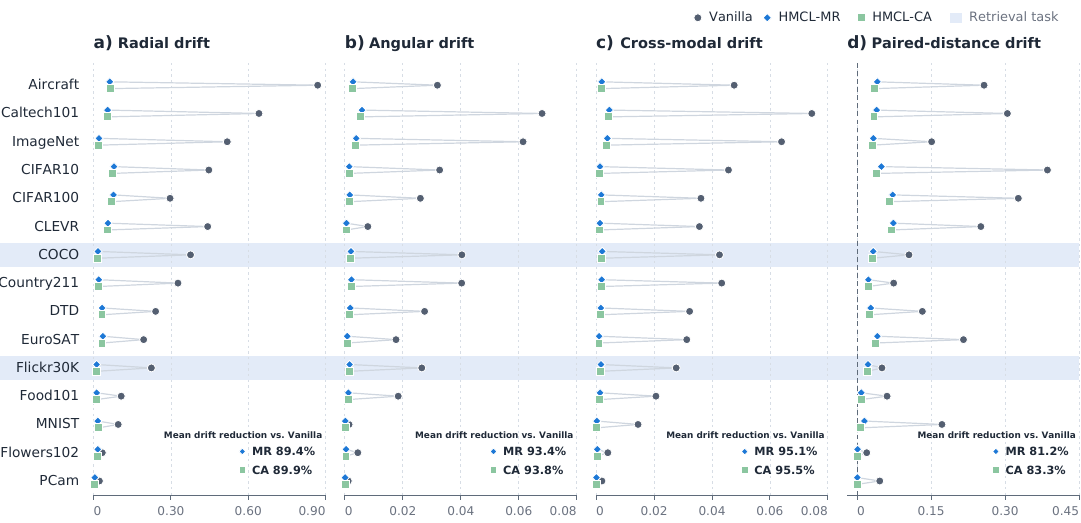}
\caption{\revisiontext{Dataset-wise drift for MERU-L (seed 1024) on the unified 16-task stream, measured from each task-end checkpoint to the final checkpoint using 50 fixed test anchors per old task. The 15 plotted rows exclude SST-2 because it is the final task and hence has no post-task drift interval. Gray circles, blue diamonds, and green squares denote Vanilla, $\mathrm{HMCL}_{\mathrm{MR}}$, and $\mathrm{HMCL}_{\mathrm{CA}}$, respectively; shaded rows identify retrieval tasks. Lower values indicate better preservation. In-panel percentages report each HMCL variant's relative reduction in mean drift over the 15 plotted tasks compared with Vanilla.}}
\label{fig:drift_analysis}
\end{figure}

To answer RQ2, we go beyond task-level BWT, which records predictive change but does not identify which parts of the hyperbolic representation have moved. We compare each old task at its task-end checkpoint and at the final checkpoint using four complementary drift measures:
\begin{itemize}[leftmargin=1.35em,topsep=1pt,itemsep=1pt,parsep=0pt]
    \item \textbf{Radial drift} is the mean absolute change of the time-like coordinate, averaged over image and text anchors. It tests hierarchical preservation in Condition~\ref{P: hierarchical}.
    \item \textbf{Angular drift} is the mean absolute change of the normalized spatial Gram matrix, averaged over the image and text blocks. It removes magnitude effects and tracks the intra-modal relations in Condition~\ref{P: intra}.
    \item \textbf{Cross-modal drift} is the mean absolute change of the normalized spatial image--text Gram block and therefore measures the inter-modal relations in Condition~\ref{P: inter}.
    \item \textbf{Paired-distance drift} is the mean change in Lorentz geodesic distance between matched image--text anchors. It tests cross-modal alignment in Condition~\ref{P: inter} while remaining sensitive to hierarchy-related radial changes in Condition~\ref{P: hierarchical}.
\end{itemize}

\begin{revisionnew}
Figure~\ref{fig:drift_analysis} shows lower drift for both HMCL variants across all 15 old tasks and all four measures. Both variants substantially suppress radial, angular, cross-modal, and paired-distance drift, with CA giving slightly lower mean drift throughout. For both variants, every paired comparison favors HMCL and is significant under an exact two-sided Wilcoxon signed-rank test ($p=6.1\times10^{-5}$; Bonferroni-adjusted $p=2.4\times10^{-4}$ over four measures). The agreement between MR and CA shows that the stability gain is not specific to the zero-rotation realization: both retain radial hierarchy, within-modality relations, and cross-modal geometry throughout the stream.

The joint behavior of these measures is more informative than any one measure alone. Small angular changes would not exclude radial contraction, and stable within-modality relations would not establish that image--text correspondence is retained. Lower paired-distance drift complements the normalized Gram measurements by tracking matched examples in the original Lorentz geometry. Taken together, the observations support preservation of both hierarchical and relational structure. They are consistent with the proposed mechanism, while remaining an empirical representation-space diagnostic rather than a proof that drift reduction alone causes the task-level gains.
\end{revisionnew}

\begin{journalnew}
\subsection{ImageNet--WordNet Hierarchy (RQ3)}
\label{subsec:journal_extended_results}

\noindent\revisiontext{\textbf{WordNet Evaluation Protocol.} To answer RQ3 quantitatively, ImageNet measures new-task capacity and WordNet~\citep{miller1995wordnet} measures whether the learned semantic hierarchy survives later tasks. Following the HyCoCLIP protocol~\citep{palcompositional} and standard hierarchical-classification evaluation~\citep{kosmopoulos2015evaluation}, we report top-1 accuracy, tree-induced error (TIE), lowest-common-ancestor distance (LCA), ancestor Jaccard, and hierarchical precision and recall. Lower TIE and LCA are better; higher values are better for all other metrics. These metrics separate prediction correctness from semantic error severity: top-1 counts every mistake equally, whereas TIE and LCA distinguish nearby errors from cross-branch confusions; ancestor Jaccard measures shared ancestry, and hierarchical precision and recall summarize the correctness and coverage of predicted ancestor paths. Their joint use prevents an apparent hierarchy gain from being attributed to a change in accuracy or only one aspect of the tree path. We therefore compute the hierarchy scores from the same final predictions as top-1 and interpret improvement only when predictive retention and semantic proximity move consistently. Appendix~\ref{apdx:imagenet_wordnet_protocol} gives the split, hierarchy mapping, retained ranks, and configuration details.}

\begin{figure}[!t]
\centering
\includegraphics[width=0.99\linewidth]{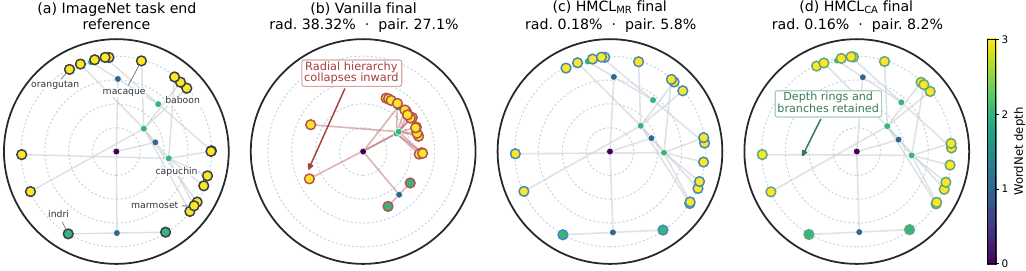}
\caption{Radial geometry of the complete 20-class ImageNet primate WordNet subtree for HyCoCLIP-B (seed 42). (a) Geometry immediately after learning ImageNet; (b--d) geometry at the final checkpoint of the 16-task stream. Final representations are globally aligned to the task-end reference by a proper spatial rotation before display, so the remaining change reflects radial and relational distortion rather than an arbitrary global orientation. Node color denotes WordNet depth, and edges show the fixed subtree relations. ``rad.'' is the mean absolute radial change normalized by the task-end mean radius; ``pair.'' is the mean absolute change in pairwise Lorentz distance normalized by the task-end mean distance. Both quantities are computed in the original representation space rather than from the two-dimensional drawing.}
\label{fig:imagenet_wordnet_radial_geometry}
\end{figure}

\Needspace{18\baselineskip}
\begin{wraptable}{r}{0.45\textwidth}
\vspace{-0.75\baselineskip}
\color{journalblue}
\centering
\scriptsize
\setlength{\tabcolsep}{1.5pt}
\renewcommand{\arraystretch}{1.05}
\captionsetup{font=footnotesize}
\caption{Final ImageNet accuracy and WordNet hierarchy metrics on the 16-task HyCoCLIP-B stream (five-seed means). HMCL uses the configurations in Table~\ref{tab:main_results16}; shaded rows identify HMCL.}
\label{tab:wordnet_hierarchy}
\begin{adjustbox}{max width=\linewidth}
\begin{tabular}{l c c c c c c}
\toprule
\textbf{Method} & \textbf{Top-1 $\uparrow$} & \textbf{TIE $\downarrow$} & \textbf{LCA $\downarrow$} & \textbf{Jacc. $\uparrow$} & \textbf{H-Prec. $\uparrow$} & \textbf{H-Rec. $\uparrow$} \\
\midrule
Vanilla & 39.404 & 3.598 & 2.217 & 0.7869 & 0.8539 & 0.8560 \\
EWC & 40.045 & 3.549 & 2.199 & 0.7901 & 0.8560 & 0.8583 \\
C-FLAT & 38.934 & 3.634 & 2.237 & 0.7849 & 0.8521 & 0.8552 \\
DNS & 39.494 & 3.594 & 2.222 & 0.7875 & 0.8538 & 0.8568 \\
\midrule
\cellcolor{myblue!20}HMCL$_{\rm MR}$ & \cellcolor{myblue!20}$45.700$ & \cellcolor{myblue!20}$3.137$ & \cellcolor{myblue!20}$2.042$ & \cellcolor{myblue!20}$0.8162$ & \cellcolor{myblue!20}$0.8743$ & \cellcolor{myblue!20}$0.8757$ \\
\cellcolor{myblue!20}HMCL$_{\rm CA}$ & \cellcolor{myblue!20}$45.590$ & \cellcolor{myblue!20}$3.156$ & \cellcolor{myblue!20}$2.051$ & \cellcolor{myblue!20}$0.8152$ & \cellcolor{myblue!20}$0.8735$ & \cellcolor{myblue!20}$0.8751$ \\
\bottomrule
\end{tabular}
\end{adjustbox}
\vspace{-0.5\baselineskip}
\end{wraptable}

\noindent\textbf{HMCL preserves the ImageNet hierarchy.}\revisiontext{\hspace{0.25em}Table~\ref{tab:wordnet_hierarchy} reports the final HyCoCLIP-B checkpoints from Table~\ref{tab:main_results16}. Both HMCL variants outperform all external baselines in top-1 accuracy and every hierarchy metric. Lower TIE and LCA together with higher ancestor Jaccard, hierarchical precision, and hierarchical recall show that their remaining errors preserve more of the correct WordNet ancestry. These concurrent gains show that hierarchy retention is not achieved by sacrificing classification performance. MR performs best on the ImageNet--WordNet measures, whereas CA gives the highest Overall score on the full stream, indicating a modest trade-off between hierarchy retention and aggregate performance. Thus, selecting a model solely by Overall can obscure differences in how well semantic ancestry survives continual adaptation.}

\Needspace{5\baselineskip}
\textbf{The radial view makes hierarchy retention visible.}
Figure~\ref{fig:imagenet_wordnet_radial_geometry} shows a representative checkpoint trajectory alongside the five-seed metrics. Vanilla contracts the depth rings and disrupts branch layout, whereas both HMCL variants stay close to the task-end geometry. Their smaller radial and pairwise-distance errors indicate preservation of both depth organization and relations among concepts. Because the views are aligned by a global spatial rotation, this contrast is not simply a difference in drawing orientation. The visualization complements the hierarchy metrics in Table~\ref{tab:wordnet_hierarchy}: the former reveals changes in representation geometry, whereas the latter evaluates semantic ancestry in the final predictions. Appendix~\ref{apdx:norm_contraction} reports modality-specific norm distributions over all 15 old tasks.

\end{journalnew}

\subsection{Representation-Space Case Study (RQ3)}
\label{sec:exp_case_study}

\definecolor{headergray}{RGB}{215, 221, 229}
\definecolor{rulegray}{RGB}{166, 176, 187}
\definecolor{greendark}{RGB}{36, 107, 75}
\definecolor{darkgreen}{RGB}{196, 223, 206}
\definecolor{middlegreen}{RGB}{217, 234, 223}
\definecolor{lightgreen}{RGB}{232, 243, 236}

\begin{figure}[!t]
    \footnotesize
    \centering
    \renewcommand{\arraystretch}{1.10}
    \arrayrulecolor{rulegray}
    \setlength{\lightrulewidth}{0.35pt}
    \setlength{\fboxsep}{0pt}
    \setlength{\fboxrule}{0.35pt}
    \setlength{\tabcolsep}{2pt}
    \newcolumntype{Z}{>{\centering\arraybackslash}X}

\begin{minipage}[t]{0.30\linewidth}
        \centering
        \fcolorbox{rulegray}{white}{\includegraphics[width=0.72\linewidth]{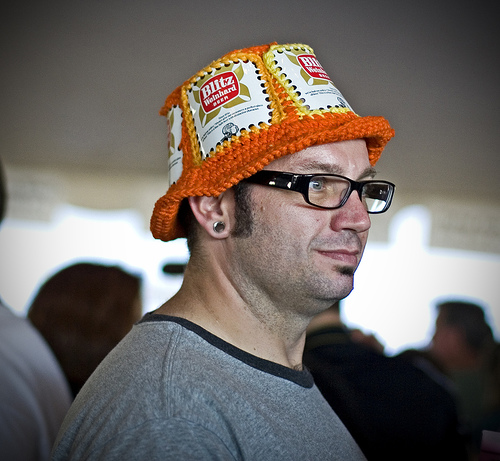}}

        \vspace{1mm}
        {\color{greendark}\rule{\linewidth}{0.65pt}}
        \vspace{0.35mm}

        \begin{tabularx}{\linewidth}{ZZ}
            \rowcolor{headergray} \textbf{HMCL (ours)} & \textbf{Vanilla} \\
            \midrule
            \rowcolor{darkgreen} \emph{hat} & \emph{volleyball} \\
            \midrule
            \rowcolor{middlegreen} \emph{fashion} & \emph{interest} \\
            \midrule
            \rowcolor{lightgreen} \emph{style} & \emph{fashion} \\
            \midrule
            \texttt{\textbf{[ROOT]}} & \texttt{\textbf{[ROOT]}} \\
        \end{tabularx}
    \end{minipage}
    \hfill
\begin{minipage}[t]{0.30\linewidth}
        \centering
        \fcolorbox{rulegray}{white}{\includegraphics[width=0.72\linewidth]{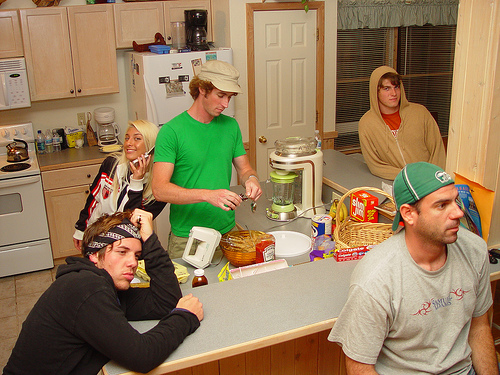}}

        \vspace{1mm}
        {\color{greendark}\rule{\linewidth}{0.65pt}}
        \vspace{0.35mm}

        \begin{tabularx}{\linewidth}{ZZ}
            \rowcolor{headergray} \textbf{HMCL (ours)} & \textbf{Vanilla} \\
            \midrule
            \rowcolor{darkgreen} \emph{relaxation} & \emph{relaxation} \\
            \midrule
            \rowcolor{middlegreen} \emph{vacation} & \emph{aloof} \\
            \midrule
            \rowcolor{lightgreen} \emph{peace} & \emph{sunken} \\
            \midrule
            \texttt{\textbf{[ROOT]}} & \texttt{\textbf{[ROOT]}} \\
        \end{tabularx}
    \end{minipage}
    \hfill
\begin{minipage}[t]{0.30\linewidth}
        \centering
        \fcolorbox{rulegray}{white}{\includegraphics[width=0.72\linewidth]{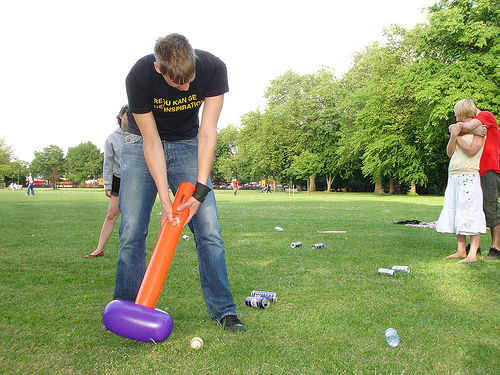}}

        \vspace{1mm}
        {\color{greendark}\rule{\linewidth}{0.65pt}}
        \vspace{0.35mm}

        \begin{tabularx}{\linewidth}{ZZ}
            \rowcolor{headergray} \textbf{HMCL (ours)} & \textbf{Vanilla} \\
            \midrule
            \rowcolor{darkgreen} \emph{defeat} & \emph{workout} \\
            \midrule
            \rowcolor{middlegreen} \emph{workout} & \emph{pilates} \\
            \midrule
            \rowcolor{lightgreen} \emph{peace} & \emph{race} \\
            \midrule
            \texttt{\textbf{[ROOT]}} & \texttt{\textbf{[ROOT]}} \\
        \end{tabularx}
    \end{minipage}
     \caption{\textbf{Hierarchical concept retrieval along image-to-origin geodesics.}
    For each Flickr30K example, we move from the image representation toward the manifold origin [ROOT] and display the nearest text concept along the resulting specific-to-generic path.
    HMCL gives more coherent sequences in these examples, such as \emph{hat} $\to$ \emph{fashion} $\to$ \emph{style}, while Vanilla more often shifts to less related concepts after sequential training.}
    \label{fig:case_study}
\end{figure}

\textbf{Construction of the Concept Set.}
To complement the quantitative answer to RQ3, we study hierarchy retention through image traversals on Flickr30K. From the original captions, nouns and adjectives are collected to form a text vocabulary with different levels of semantic abstraction. Appendix~\ref{apdx:case_study} details how the concept set is prepared.

\textbf{Traversal Protocol.}
For each image, we trace its embedding toward the manifold origin [ROOT], which represents the generic concept. We sample 20 locations in the origin's tangent space, map them back to the Lorentz hyperboloid, and retrieve the nearest text representation at each location.

\revisiontext{\textbf{Qualitative Finding.}
Figure~\ref{fig:case_study} shows a recurring contrast in the displayed examples. HMCL maintains local semantic continuity along the geodesic---for example, \emph{hat} $\to$ \emph{fashion} $\to$ \emph{style} and \emph{relaxation} $\to$ \emph{vacation} $\to$ \emph{peace}---whereas Vanilla introduces abrupt off-path transitions such as \emph{volleyball} $\to$ \emph{interest} and \emph{aloof} $\to$ \emph{sunken}. Because the HMCL paths become progressively more generic before reaching [ROOT], the result suggests preservation of intermediate abstraction ordering rather than only convergence to the shared endpoint. This qualitative evidence complements the WordNet and radial-geometry results in Section~\ref{subsec:journal_extended_results}, thereby providing a representation-level answer to RQ3. Appendix~\ref{apdx:case_study} provides additional Flickr30K and COCO traversals.}

\begin{journalnew}
\subsection{Ablation Study (RQ4)}
\label{subsec:hmcl_component_analysis}

To answer RQ4, we isolate where the geometric correction is applied and whether task anchoring is added, while keeping the HyCoCLIP-B stream and evaluation metrics fixed.

\Needspace{18\baselineskip}
\begin{wraptable}{r}{0.45\textwidth}
\vspace{-0.75\baselineskip}
\color{journalblue}
\centering
\scriptsize
\setlength{\tabcolsep}{1.5pt}
\renewcommand{\arraystretch}{1.06}
\arrayrulecolor{black}
\setlength{\heavyrulewidth}{0.8pt}
\setlength{\lightrulewidth}{0.5pt}
\captionsetup{font=footnotesize}
\caption{Focused update-placement ablation on the 16-task HyCoCLIP-B stream (five-seed mean $\pm$ sample standard deviation). Each placement reports both MR and CA. Pre and Post apply the correction to the raw gradient and realized AdamW displacement, respectively; Post+Pullback adds task anchoring.}
\label{tab:optimizer_placement}
\begin{adjustbox}{max width=\linewidth}
\begin{tabular}{@{}l c c c c@{}}
\toprule
\textbf{Update} & \textbf{Corr.} & \textbf{Overall $\uparrow$} & \textbf{BWT $\uparrow$} & \textbf{ImageNet $\uparrow$} \\
\midrule
Vanilla & -- & $43.618\pm0.152$ & $-4.186\pm0.087$ & $39.404\pm0.058$ \\
\midrule
\multirow{2}{*}{Pre} & MR & $46.397\pm0.124$ & $-2.907\pm0.070$ & $41.303\pm0.095$ \\
& CA & $47.172\pm0.129$ & $-2.932\pm0.058$ & $40.871\pm0.007$ \\
\midrule
\multirow{2}{*}{Post} & MR & $47.780\pm0.106$ & $-1.548\pm0.073$ & $43.760\pm0.032$ \\
& CA & $47.785\pm0.102$ & $-1.532\pm0.075$ & $43.759\pm0.043$ \\
\midrule
\multirow{2}{*}{\makecell[l]{Post +\\Pullback}} & MR & $48.053\pm0.072$ & $\mathbf{-0.239}\pm0.057$ & $\mathbf{46.257}\pm0.033$ \\
& CA & $\mathbf{48.566}\pm0.024$ & $-0.456\pm0.051$ & $45.585\pm0.052$ \\
\bottomrule
\end{tabular}
\end{adjustbox}
\vspace{-0.5\baselineskip}
\end{wraptable}

\begin{revisionnew}
Table~\ref{tab:optimizer_placement} shows that all corrected configurations outperform Vanilla in the Overall score, BWT, and ImageNet accuracy. The consistent advantage of Post over Pre for both MR and CA shows that the constraint is most effective when applied to AdamW's realized displacement rather than to the raw gradient. Pullback then complements the geometric correction by limiting cumulative within-task deviation, improving retention without removing the flexibility needed for the current task. Post+Pullback CA attains the highest Overall score, whereas Post+Pullback MR gives the strongest BWT and ImageNet retention.

The placement comparison separates optimizer alignment from the choice of admissible representative. Both MR and CA benefit from correcting the realized displacement, consistent with the fact that AdamW's momentum and adaptive scaling can change the direction proposed by the raw gradient. Task anchoring addresses a different source of deviation: locally corrected updates can still accumulate change over a task. The further retention gains with Pullback therefore support its complementary role within the integrated update, rather than an alternative to geometric correction.

The MR ablation uses stronger task anchoring than its main-table configuration. As detailed in Appendix~\ref{apdx:pullback_sensitivity}, this favors retention, whereas the main-table setting favors the Overall score; the configurations therefore represent different stability--plasticity trade-offs within the same update.
\end{revisionnew}
\end{journalnew}

\Needspace{8\baselineskip}
\section{Conclusion}

\begin{revisionnew}
We presented a continual-learning formulation for hyperbolic multimodal representations and characterized the updates that retain within-modal relations, cross-modal alignment, and semantic hierarchy. HMCL translates these representation-space conditions into a shared admissible family with two complementary realizations: CA permits a common rotation, whereas MR selects the identity representative. Across backbones, both variants improve final performance and retention over the external baselines; CA achieves the highest Overall scores, while MR gives the best ImageNet--WordNet hierarchy scores. The stage-wise, drift, modality-extension, and hierarchy analyses show that these gains coincide with more stable radial and relational structure rather than reflecting one aggregate metric alone. The component study further shows that geometric correction and task anchoring address complementary sources of change. Together, the results support shared geometric admissibility as a practical stability--plasticity principle for continual adaptation in non-Euclidean representation spaces.
\end{revisionnew}

\FloatBarrier

\clearpage
\appendix
\begin{center}
  {\Huge\bfseries\color{arxivblue} Appendix}\par
  \vspace{0.35em}
  {\large\color{black!70}Supplementary derivations, implementation details, and extended results}
\end{center}
\vspace{1.5em}

\begin{tcolorbox}[
  enhanced,
  breakable,
  colback=arxivlight,
  colframe=arxivblue!45,
  boxrule=0.55pt,
  arc=1.5pt,
  left=12pt,right=12pt,top=10pt,bottom=8pt
]
{\large\bfseries\color{arxivblue}Contents}\par
\vspace{0.35em}
\appendixcontents
\end{tcolorbox}

\vfill

\appendixsection{Notation Table}
\label{apdx:notation}

Table~\ref{tab:notation} collects the symbols and conventions used in the main paper and this supplement.

\begingroup
\footnotesize
\setlength{\LTleft}{\fill}
\setlength{\LTright}{\fill}
\setlength{\tabcolsep}{5pt}
\renewcommand{\arraystretch}{1.02}
\begin{longtable}{@{}p{0.18\textwidth}p{0.74\textwidth}@{}}
\caption{Summary of the notation used in this work.}
\label{tab:notation}\\
\toprule
\textbf{Symbol} & \textbf{Description} \\
\midrule
\endfirsthead
\multicolumn{2}{c}{\tablename\ \thetable\ (continued)}\\
\toprule
\textbf{Symbol} & \textbf{Description} \\
\midrule
\endhead
\midrule
\multicolumn{2}{r}{Continued on next page}\\
\endfoot
\bottomrule
\endlastfoot

\multicolumn{2}{@{}l}{\textbf{Hyperbolic geometry and the Lorentz model}}\\[1pt]
$d$ & Intrinsic hyperbolic dimension; the Lorentz ambient space has dimension $d+1$. \\
$K$ & Squared hyperboloid radius ($K>0$); the sectional curvature is $-1/K$. \\
$\mathbb{H}^d_K$ & $d$-dimensional hyperbolic space with curvature $-1/K$. \\
$\mathbf{G}$ & Minkowski metric tensor, $\mathbf{G}=\operatorname{diag}(-1,1,\ldots,1)\in\mathbb{R}^{(d+1)\times(d+1)}$. \\
$\langle\cdot,\cdot\rangle_{\mathcal L}$ & Lorentzian inner product, $\langle\mathbf{x},\mathbf{y}\rangle_{\mathcal L}=\mathbf{x}^{\top}\mathbf{G}\mathbf{y}$. \\
$d_{\mathcal L}^K(\mathbf{x},\mathbf{y})$ & Geodesic distance, $\sqrt{K}\operatorname{arcosh}(-\langle\mathbf{x},\mathbf{y}\rangle_{\mathcal L}/K)$. \\
$\mathbf{o}$ & Origin of the Lorentz model, $(\sqrt{K},0,\ldots,0)^\top\in\mathbb{H}^d_K$. \\
$x_0,\mathbf{x}_{[1:d]}$ & Time-like coordinate and $d$ space-like coordinates of $\mathbf x$. \\
$\mathrm{SO}^+(1,d),\mathrm{SO}(d)$ & Proper orthochronous Lorentz group and spatial rotation group. \\
\addlinespace[3pt]

\multicolumn{2}{@{}l}{\textbf{Multimodal learning}}\\[1pt]
$M$ & Number of modalities. \\
$m,m'$ & Modality indices, $m,m'\in\{1,2,\ldots,M\}$. \\
$N$ & Number of paired samples in the current context. \\
$\mathcal{X}^m$ & Data space of modality $m$. \\
$\mathcal{X}^{m,m'}$ & Paired dataset $\{(\mathbf{x}_i^m,\mathbf{x}_i^{m'})\}_{i=1}^N$. \\
$\mathbf{x}_i^m,\mathbf{z}_i^m$ & The $i$-th input and its hyperbolic embedding for modality $m$. \\
\addlinespace[3pt]

\multicolumn{2}{@{}l}{\textbf{Continual learning framework}}\\[1pt]
$t$ & Task or stage index. \\
$J$ & Total number of optimizer steps within a task. \\
$k$ & Within-task optimizer-step index, $k\in\{0,\ldots,J-1\}$. \\
$T_m^{(t)}$ & Lorentz transformation for modality $m$ at task $t$. \\
$\mathbf{W}^{m,t}$ & Final Lorentz-head parameters at task $t$. \\
\journaltext{$\mathbf W_{k}^{m,t}=[\,\mathbf w_{k,0}^{m,t}\;\;\mathbf W_{k,s}^{m,t}\,]$} & \journaltext{Parameters after step $k$, partitioned into the time-like input column and spatial input block; $\mathbf W_0^{m,t}=\mathbf W^{m,t-1}$ and $\mathbf W_J^{m,t}=\mathbf W^{m,t}$.} \\
\journaltext{$\widetilde{\mathbf W}^{m,t}_{k+1}$} & \journaltext{Optimizer-proposed parameter candidate before HMCL correction.} \\
\journaltext{$\mathbf g_k^{m,t}$} & \journaltext{Raw gradient at step $k$.} \\
$X_t^m$ & Input data for task $t$ and modality $m$. \\
$\mathbf{Z}_{q}^{m,*}$ & Frozen pretrained Lorentz representations of task-$q$ data from modality $m$. \\
$\mathbf{Z}_{q}^{m,t}$ & The same data after the stage-$t$ Lorentz head. \\
$f(\mathbf{W};\cdot)$ & Lorentz layer in~\eqref{eq:supp_lorentz_layer}. \\
\addlinespace[3pt]

\multicolumn{2}{@{}l}{\textbf{Similarity measures and objectives}}\\[1pt]
$\mathbf{S}^{m\rightarrow m'}$ & Similarity matrix from modality $m$ to $m'$, with $\mathbf{S}^{m\rightarrow m'}\in\mathbb{R}^{N\times N}$. \\
$S_{ij}^{m\rightarrow m'}$ & Entry $(i,j)$ of $\mathbf S^{m\rightarrow m'}$, equal to $-d_{\mathcal L}^K(\mathbf{z}_i^m,\mathbf{z}_j^{m'})$. \\
$\tau$ & Temperature parameter for contrastive learning. \\
$\mathcal{L}_{m\rightarrow m'}$ & Unidirectional contrastive loss from modality $m$ to $m'$. \\
$\mathcal{L}_{\mathrm{contrast}}$ & Symmetric contrastive loss $\tfrac12(\mathcal{L}_{m\rightarrow m'}+\mathcal{L}_{m'\rightarrow m})$. \\
$\operatorname{aper}(\mathbf{z})$ & Entailment-cone aperture at embedding $\mathbf{z}$. \\
$\operatorname{ext}(\mathbf{z},\mathbf{z}')$ & Exterior angle between embeddings $\mathbf{z}$ and $\mathbf{z}'$. \\
$\kappa$ & Boundary constant for entailment cones (typically $\kappa=0.1$). \\
$\mathcal{L}_{\mathrm{entail}}$ & Entailment loss for hierarchical relationships. \\
\addlinespace[3pt]

\multicolumn{2}{@{}l}{\textbf{Geometric analysis and parameter updates}}\\[1pt]
$\mathbf{L}$ & General Lorentz transformation, $\mathbf{L}\in\mathrm{SO}^+(1,d)$. \\
$\mathbf{R}$ & Spatial Lorentz rotation $\operatorname{diag}(1,\widetilde{\mathbf{R}})$. \\
$\widetilde{\mathbf{R}}$ & Spatial rotation matrix in $\mathrm{SO}(d)$. \\
$\boldsymbol{\delta}^m=[\,\boldsymbol\delta_0^m\;\;\boldsymbol\delta_s^m\,]$ & Candidate parameter displacement, partitioned by input coordinate. \\
\journaltext{$\boldsymbol{\delta}^{\mathrm{opt},m,t}_{k}$} & \journaltext{Optimizer-proposed displacement $\widetilde{\mathbf W}_{k+1}^{m,t}-\mathbf W_k^{m,t}$ before HMCL correction.} \\
\journaltext{$\boldsymbol{\delta}^{\star,m,t}_{k}$} & \journaltext{Closest admissible displacement obtained from $\boldsymbol\delta_k^{\mathrm{opt},m,t}$.} \\
\journaltext{$\mathbf A_k^m$} & \journaltext{Cumulative displacement $\mathbf W_k^{m,t}-\mathbf W_0^{m,t}$.} \\
\journaltext{$\mathcal E_{\mathcal L,t-1}$} & \journaltext{Defect from a shared first-order Lorentz rotation on protected representations.} \\
\journaltext{$\mathcal A_{t-1},\mathcal P^{\mathcal L}_{t-1}$} & \journaltext{Zero-defect admissible family and its Frobenius closest-point map.} \\
\journaltext{$\mathbf V_{t-1},\mathbf P_{t-1}$} & \journaltext{Protected-subspace basis and projector, with $\mathbf P_{t-1}=\mathbf V_{t-1}\mathbf V_{t-1}^{\top}$.} \\
\journaltext{$\beta$} & \journaltext{Pullback coefficient toward the task-start parameters.} \\
\end{longtable}
\endgroup

\appendixsection{Additional Preliminaries}
\label{apdx:additional_preliminaries}

\appendixsubsection{Why Hyperbolic Geometry Represents Hierarchies}

Hyperbolic space is a homogeneous geometry with constant negative curvature. Its volume expands exponentially with radius: in $n$ dimensions and curvature $-1$, a radius-$r$ ball has volume
\[
V_{\mathbb{H}^n}(r)\propto\int_0^r\sinh^{n-1}(t)\,dt,
\]
which is asymptotically proportional to $e^{(n-1)r}$. Euclidean volume, in contrast, grows only as $r^n$. In the two-dimensional case, for instance, the hyperbolic disk area is $2\pi(\cosh r-1)$ rather than the Euclidean value $\pi r^2$. This exponential capacity closely matches the growth of tree-structured data, which is why hyperbolic embeddings can represent taxonomies, trees, and knowledge graphs with low distortion~\citep{nickel2017poincare,sarkar2011low}.

\appendixsubsection{Formal Description of the Lorentz Model}
\label{apdx:lorentz_manifold}

The Lorentz model embeds a $d$-dimensional hyperbolic manifold as one sheet of a hyperboloid in $(d+1)$-dimensional Minkowski space.

\begin{definition}[Minkowski Space and Metric]
\label{def: minkowski space}
Minkowski space is $\mathbb{R}^{d+1}$ equipped with the indefinite metric
\[
\mathbf{G}=\operatorname{diag}(-1,1,\ldots,1)\in\mathbb{R}^{(d+1)\times(d+1)}\revisiontext{.}
\]
\revisiontext{Here $K>0$ denotes the squared hyperboloid radius used below, so the corresponding sectional curvature is $-1/K$.} Its signature $(-,+,\ldots,+)$ distinguishes one time-like direction from $d$ space-like directions.
\end{definition}

\begin{definition}[Lorentzian Inner Product]
\label{def: inner product}
For $\mathbf{x},\mathbf{y}\in\mathbb{R}^{d+1}$, their Lorentzian (Minkowski) inner product is
\[
\langle\mathbf{x},\mathbf{y}\rangle_{\mathcal L}
=\mathbf{x}^{\top}\mathbf{G}\mathbf{y}
=-x_0y_0+\sum_{i=1}^{d}x_i y_i.
\]
\end{definition}

\begin{definition}[Lorentz Manifold]
\label{def: lorentz manifold}
The Lorentz realization of $d$-dimensional hyperbolic space with curvature $-1/K$ is
\[
\mathbb{H}_K^d
=\left\{\mathbf{x}\in\mathbb{R}^{d+1}:
\mathbf{x}^{\top}\mathbf{G}\mathbf{x}=-K,\;x_0>0\right\}.
\]
We write $\mathcal{L}_K^d=(\mathbb{H}_K^d,\mathbf{G})$ when referring to the manifold together with its metric.
\end{definition}

\begin{remark}
Expanding the manifold constraint gives
\[
-x_0^2+\sum_{i=1}^{d}x_i^2=-K,
\qquad
x_0^2-\sum_{i=1}^{d}x_i^2=K.
\]
This equation describes a two-sheeted hyperboloid; $x_0>0$ selects its connected upper sheet. For any $\mathbf{x}\in\mathbb{H}_K^d$,
\[
x_0=\sqrt{K+\|\mathbf{x}_{[1:d]}\|^2}\geq\sqrt K,
\]
where $\mathbf{x}_{[1:d]}=(x_1,\ldots,x_d)^{\top}$ contains the space-like coordinates.
\end{remark}

\begin{definition}[Lorentzian Distance]
\label{def: lorentz distance}
The length of the geodesic between $\mathbf{x},\mathbf{y}\in\mathcal{L}_K^d$ is
\begin{equation}
d_{\mathcal L}^{K}(\mathbf{x},\mathbf{y})
=\sqrt{K}\operatorname{arcosh}\!\left(
-\frac{\langle\mathbf{x},\mathbf{y}\rangle_{\mathcal L}}{K}
\right)
=\sqrt{K}\operatorname{arcosh}\!\left(
-\frac{\mathbf{x}^{\top}\mathbf{G}\mathbf{y}}{K}
\right).
\label{equ: distance}
\end{equation}
\end{definition}

\begin{remark}
For points on the upper hyperboloid, $\langle\mathbf{x},\mathbf{y}\rangle_{\mathcal L}\leq-K$, with equality if and only if $\mathbf{x}=\mathbf{y}$. The argument of $\operatorname{arcosh}$ is therefore at least one, so~\eqref{equ: distance} is well defined. Substituting $\mathbf{x}=\mathbf{y}$ gives zero distance.
\end{remark}

\begin{remark}[Relation to Ambient Euclidean Distance]
Far from the origin, geodesic distance increases approximately logarithmically with ambient Euclidean distance. This growth allows hyperbolic space to represent tree-like structures with low distortion.
\end{remark}

The linear symmetries of the Lorentz model form a group of transformations that leave the Lorentzian inner product unchanged.

\begin{definition}[Proper Orthochronous Lorentz Group]
\label{def:lorentz_group}
The proper orthochronous Lorentz group is
\[
\mathrm{SO}^{+}(1,d)
=\left\{\mathbf{L}\in\mathbb{R}^{(d+1)\times(d+1)}:
\mathbf{L}^{\top}\mathbf{G}\mathbf{L}=\mathbf{G},\;
\det(\mathbf{L})=1,\;L_{00}\geq1\right\}.
\]
\end{definition}

\begin{remark}
The metric constraint makes $\mathbf L$ an isometry, the determinant condition excludes orientation-reversing reflections, and \revisiontext{$L_{00}\geq1$ preserves the time orientation of the upper hyperboloid.} The group can also map any point on $\mathbb{H}_K^d$ to any other point: for $\mathbf{x},\mathbf{y}\in\mathbb{H}_K^d$, some $\mathbf{L}\in\mathrm{SO}^{+}(1,d)$ satisfies $\mathbf{L}\mathbf{x}=\mathbf{y}$.
\end{remark}

\appendixsubsection{Lorentz Neural Layers}

\begin{definition}[Lorentz Transformation Layer~\citep{chen2022fully}]
Let $\mathbf{z}\in\mathbb{R}^{d+1}$ be a Lorentz representation, with time-like component $z_0>0$ and spatial block $\mathbf{z}_s\in\mathbb{R}^{d}$. A Lorentz transformation layer is defined by
\begin{equation}
f(\mathbf{W};\mathbf{z})
=
\begin{bmatrix}
\sqrt{K+\|\mathbf{W}\mathbf{z}\|_2^2}\\
\mathbf{W}\mathbf{z}
\end{bmatrix},
\qquad
\mathbf{W}\in\mathbb{R}^{d\times(d+1)}.
\label{eq:supp_lorentz_layer}
\end{equation}
\end{definition}

\begin{remark}
The time-like output in~\eqref{eq:supp_lorentz_layer} is reconstructed from the spatial output so that the result satisfies the Lorentz manifold constraint.
\end{remark}

For the block-wise analysis, we partition the weights as
\begin{equation}
\mathbf{W}=[\,\mathbf{w}_0\;\;\mathbf{W}_s\,],
\qquad
\mathbf{w}_0\in\mathbb{R}^{d\times1},
\quad
\mathbf{W}_s\in\mathbb{R}^{d\times d},
\label{eq:W_block}
\end{equation}
where $\mathbf{w}_0$ multiplies the time-like input and $\mathbf{W}_s$ acts on the space-like input coordinates.

\appendixsection{Rationale for the Preservation Objectives}
\label{apdx:problem_statement}

Continual learning aims to incorporate the current task while retaining information acquired earlier. For hyperbolic representations, that information is represented by geometry: pairwise relations encode semantic similarity, while radial position encodes semantic specificity and hierarchy~\cite{desai2023hyperbolic}. We therefore use the following three requirements for representations of earlier data:

\begin{enumerate}[label=(\textbf{P\arabic*})]
\item \textbf{Intra-modal preservation}:
\[
\mathbf{Z}_{t-1}^{m,t}\mathbf{G}(\mathbf{Z}_{t-1}^{m,t})^{\top}
=\mathbf{Z}_{t-1}^{m,t-1}\mathbf{G}(\mathbf{Z}_{t-1}^{m,t-1})^{\top},
\quad \forall m\in\{1,\ldots,M\};
\]
\item \textbf{Inter-modal preservation}:
\[
\mathbf{Z}_{t-1}^{m,t}\mathbf{G}(\mathbf{Z}_{t-1}^{m',t})^{\top}
=\mathbf{Z}_{t-1}^{m,t-1}\mathbf{G}(\mathbf{Z}_{t-1}^{m',t-1})^{\top},
\quad \forall m\neq m';
\]
\item \textbf{Hierarchical preservation}:
\[
\| (\mathbf{z}_{t-1}^{m,t})_{[1:d]}\|
=\|(\mathbf{z}_{t-1}^{m,t-1})_{[1:d]}\|,
\quad \forall m\in\{1,\ldots,M\}.
\]
\end{enumerate}

\begin{table}[htbp]
\centering
\caption{Links between the desired semantic properties and the hyperbolic quantities that encode them.}
\label{tab:preservation_correspondence}
\begin{tabular}{lcc}
\toprule[1.2pt]
\textbf{Semantic Property} & \textbf{Geometric Quantity} & \textbf{Condition} \\
\midrule[1pt]
\rowcolor{myblue!20}\multicolumn{3}{l}{\textbf{$-$ Directly constrained}} \\
Semantic similarity & $d_{\mathcal L}(\mathbf{x},\mathbf{y})\propto\langle\mathbf{x},\mathbf{y}\rangle_{\mathcal L}$ & \ref{P: intra}, \ref{P: inter} \\
Semantic specificity & $d_{\mathcal L}(\mathbf{x},\mathbf{o})\propto\|\mathbf{x}_{[1:d]}\|$ & \ref{P: hierarchical} \\
\midrule
\rowcolor{myblue!20}\multicolumn{3}{l}{$-$ \textbf{Implied consequence} (Proposition~\ref{prop:partial_order_invariance})} \\
Concept partial order & $\operatorname{aper}(\mathbf{x}),\operatorname{ext}(\mathbf{x},\mathbf{y})$ & \textit{Implied by the conditions} \\
\bottomrule[1.2pt]
\end{tabular}
\end{table}

Table~\ref{tab:preservation_correspondence} links each preservation goal to the geometric quantity that represents it. Conditions~\ref{P: intra} and~\ref{P: inter} fix Lorentzian inner products. Since the distance in~\eqref{equ: distance} is a monotone function of these products, the associated semantic similarities remain unchanged. Condition~\ref{P: hierarchical} fixes the spatial norm. On the hyperboloid,
\[
d_{\mathcal L}^{K}(\mathbf{x},\mathbf{o})
=\sqrt K\operatorname{arcosh}\!\left(\sqrt{1+\|\mathbf{x}_{[1:d]}\|_2^2/K}\right).
\]
Preserving the spatial norm therefore keeps a concept at the same level of specificity.

Together, these invariants are also sufficient to retain the entailment relation used to encode concept order.

\begin{thmbox}
\begin{proposition}[Invariance of the Concept Partial Order]
\label{prop:partial_order_invariance}
Let $\mathbf{z}_i^m,\mathbf{z}_i^{m'}\in\mathbb{H}_K^d$ be hyperbolic embeddings. If $\|(\mathbf{z}_i^m)_{[1:d]}\|_2$ and $\langle\mathbf{z}_i^m,\mathbf{z}_i^{m'}\rangle_{\mathcal L}$ do not change, then the entailment relation $\mathbf{z}_i^m\sqsupseteq\mathbf{z}_i^{m'}$ is unchanged as well.
\end{proposition}
\end{thmbox}

\begin{proof}
Entailment holds when $\mathcal{L}_{\mathrm{entail}}(\mathbf{z}_i^m,\mathbf{z}_i^{m'})=0$, or equivalently
\[
\operatorname{ext}(\mathbf{z}_i^m,\mathbf{z}_i^{m'})
\leq\operatorname{aper}(\mathbf{z}_i^m).
\]
The two angles are
\begin{align}
\operatorname{aper}(\mathbf{z}_i^m)
&=\sin^{-1}\!\left(\frac{2\kappa\sqrt K}{\|(\mathbf{z}_i^m)_{[1:d]}\|}\right),\\
\operatorname{ext}(\mathbf{z}_i^m,\mathbf{z}_i^{m'})
&=\cos^{-1}\!\left(
\frac{(\mathbf{z}_i^{m'})_0+(\mathbf{z}_i^m)_0
\langle\mathbf{z}_i^m,\mathbf{z}_i^{m'}\rangle_{\mathcal L}/K}
{\|(\mathbf{z}_i^m)_{[1:d]}\|
\sqrt{\big(\langle\mathbf{z}_i^m,\mathbf{z}_i^{m'}\rangle_{\mathcal L}/K\big)^2-1}}
\right).
\end{align}
The aperture depends only on the spatial norm, while the exterior angle depends on that norm and the Lorentzian inner product. The time-like coordinate follows from the manifold constraint. If these quantities are preserved, neither side of the entailment inequality changes, so the corresponding partial-order relation is invariant.
\end{proof}

\appendixsection{Proofs for the Geometric Characterization}
\label{apdx:geometric_proofs}

\begin{definition}[Joint Embedding Matrix]
\label{def:joint_emb_matrix}
Consider $N$ paired examples from task $t\in\mathcal T$, evaluated at stage $s\in\mathcal S$ for modalities $m_1$ and $m_2$. Given $\mathbf{Z}_{t}^{m,s}\in\mathbb{R}^{N\times(d+1)}$ with rows $\mathbf{z}_{t,i}^{m,s}\in\mathbb{H}_K^d$, define
\[
\mathbf{Z}_{t,\mathrm{all}}^{s}
:=
\begin{pmatrix}
\mathbf{Z}_{t}^{m_1,s}\\[2pt]
\mathbf{Z}_{t}^{m_2,s}
\end{pmatrix}
\in\mathbb{R}^{2N\times(d+1)}.
\]
The ordering in this vertical concatenation retains the pairing between the two modalities.
\end{definition}

\begin{definition}[Extended Lorentz Gram Matrix]
\label{def: extended lorentz gram matrix}
For $\mathbf{G}=\operatorname{diag}(-1,1,\ldots,1)$, the extended Gram matrix of the joint embeddings is
\begin{align}
\mathbf{G}_{t,\mathrm{ext}}^{s}
&:=\mathbf{Z}_{t,\mathrm{all}}^{s}\mathbf{G}(\mathbf{Z}_{t,\mathrm{all}}^{s})^{\top}\\
&=\begin{pmatrix}
\mathbf{Z}_{t}^{m_1,s}\mathbf{G}(\mathbf{Z}_{t}^{m_1,s})^{\top}
&\mathbf{Z}_{t}^{m_1,s}\mathbf{G}(\mathbf{Z}_{t}^{m_2,s})^{\top}\\[2pt]
\mathbf{Z}_{t}^{m_2,s}\mathbf{G}(\mathbf{Z}_{t}^{m_1,s})^{\top}
&\mathbf{Z}_{t}^{m_2,s}\mathbf{G}(\mathbf{Z}_{t}^{m_2,s})^{\top}
\end{pmatrix}.
\end{align}
Its diagonal blocks contain within-modality Lorentzian products, while the off-diagonal blocks record the cross-modal geometry.
\end{definition}

\begin{definition}[Lorentz Boost]
\label{def: lorentz boost}
For $\mathbf{v}\in\mathbb{R}^{d}$ with $\|\mathbf{v}\|<1$ and $\gamma=(1-\|\mathbf{v}\|^2)^{-1/2}$, the boost associated with $\mathbf v$ is
\[
\mathbf{B}=
\begin{bmatrix}
\gamma&-\gamma\mathbf{v}^{\top}\\
-\gamma\mathbf{v}&
\mathbf{I}+\dfrac{\gamma^2}{1+\gamma}\mathbf{v}\mathbf{v}^{\top}
\end{bmatrix}.
\]
It mixes the time-like coordinate with the direction $\mathbf v$ without independently rotating the spatial axes.
\end{definition}

\begin{definition}[Lorentz Rotation]
\label{def: lorentz rotation}
A spatial Lorentz rotation has block form
\[
\mathbf{R}=
\begin{bmatrix}
1&\mathbf{0}^{\top}\\
\mathbf{0}&\widetilde{\mathbf{R}}
\end{bmatrix},
\qquad
\widetilde{\mathbf{R}}^{\top}\widetilde{\mathbf{R}}=\mathbf I,
\quad
\det(\widetilde{\mathbf R})=1.
\]
Thus $\widetilde{\mathbf R}\in\mathrm{SO}(d)$ rotates only the spatial coordinates.
\end{definition}

\appendixsubsection{Shared Isometry Induced by Gram Preservation}
\label{apdx:proof for them1}

\begin{lemma}[Witt's Extension Theorem~\cite{lam2005introduction}]
\label{lem:witt}
Let $(\mathcal V,Q)$ be a non-degenerate quadratic space. Any isometry $\phi:\mathcal U\rightarrow\mathcal V'$ between subspaces $\mathcal U,\mathcal V'\subseteq\mathcal V$ extends to an isometry of all of $\mathcal V$.
\end{lemma}

\begin{remark}
In matrix notation, if $\mathbf{Z}_1\mathbf{Q}\mathbf{Z}_1^{\top}=\mathbf{Z}_2\mathbf{Q}\mathbf{Z}_2^{\top}$ for a non-degenerate symmetric $\mathbf Q$, then an element $\mathbf T$ of
\[
O(\mathbf Q)=\{\mathbf T:\mathbf T^{\top}\mathbf Q\mathbf T=\mathbf Q\}
\]
maps the first set of rows to the second, i.e., $\mathbf Z_2=\mathbf Z_1\mathbf T^{\top}$.
\end{remark}

\begin{lemma}[Hyperboloid-Preserving Isometries]
\label{lem:hyperboloid_isometry}
Let $\mathbf{L}\in O(1,d)$ be an orientation-preserving isometry that maps the upper hyperboloid
\[
\mathbb{H}_K^d=\{\mathbf{x}:\mathbf{x}^{\top}\mathbf G\mathbf{x}=-K,\;x_0>0\}
\]
onto itself. Then $\mathbf L\in\mathrm{SO}^{+}(1,d)$.
\end{lemma}

\begin{proof}
Because the upper sheet is invariant, applying $\mathbf L$ to $\mathbf o=(\sqrt K,0,\ldots,0)^{\top}$ gives $(\mathbf L\mathbf o)_0=\sqrt K L_{00}>0$. The identity $\mathbf L^{\top}\mathbf G\mathbf L=\mathbf G$ further implies
\[
-L_{00}^{2}+\sum_{i=1}^{d}L_{i0}^{2}=-1,
\]
so $L_{00}^{2}=1+\sum_{i=1}^{d}L_{i0}^{2}\geq1$ and therefore $L_{00}\geq1$. Orientation preservation gives $\det(\mathbf L)=1$. Together with $\mathbf L\in O(1,d)$, these are the defining conditions of $\mathrm{SO}^{+}(1,d)$.
\end{proof}

\begin{thmbox}
\begin{theorem}[Uniqueness of the Lorentzian Isometry]
\label{thm:witt_multimodal}
Let $\mathbf{Z}_{t,\mathrm{all}}^{s},\mathbf{Z}_{t',\mathrm{all}}^{s'}\in\mathbb{R}^{2N\times(d+1)}$. Suppose
\begin{enumerate}[label=(\arabic*)]
\item their extended Gram matrices agree, $\mathbf{G}_{t,\mathrm{ext}}^{s}=\mathbf{G}_{t',\mathrm{ext}}^{s'}$; and
\item both joint matrices have full column rank $d+1$; and
\item the stage-to-stage correspondence is induced by a continuous path of full-column-rank configurations with the same extended Gram matrix, beginning at the first configuration.
\end{enumerate}
Then there is a unique $\mathbf L\in\mathrm{SO}^{+}(1,d)$ satisfying
\[
\mathbf{Z}_{t',\mathrm{all}}^{s'}
=\mathbf{Z}_{t,\mathrm{all}}^{s}\mathbf L^{\top}.
\]
\end{theorem}
\end{thmbox}

\begin{proof}
Full column rank means that the rows of both joint matrices span the ambient Lorentzian space. Associate each row $\mathbf z_{t,i}^{m,s}$ with the corresponding row $\mathbf z_{t',i}^{m,s'}$. Equality of the extended Gram matrices gives
\[
\langle\mathbf z_{t,i}^{m,s},\mathbf z_{t,j}^{m',s}\rangle_{\mathcal L}
=
\langle\mathbf z_{t',i}^{m,s'},\mathbf z_{t',j}^{m',s'}\rangle_{\mathcal L}
\]
for every pair of samples and modalities. This correspondence is therefore an isometry on a generating set. Lemma~\ref{lem:witt} extends it to a global element $\mathbf L\in O(1,d)$ such that
\[
\mathbf{Z}_{t',\mathrm{all}}^{s'}
=\mathbf{Z}_{t,\mathrm{all}}^{s}\mathbf L^{\top}.
\]
It remains to identify the connected component. Let $\mathbf Z(\lambda)$ denote the path in condition~(3), with $\mathbf Z(0)=\mathbf Z_{t,\mathrm{all}}^s$. Full rank makes the corresponding isometry unique, and
\[
\mathbf L(\lambda)^{\top}
=\big(\mathbf Z(0)^{\top}\mathbf Z(0)\big)^{-1}
\mathbf Z(0)^{\top}\mathbf Z(\lambda)
\]
makes its continuity explicit. Thus $\mathbf L(0)=\mathbf I$ and $\mathbf L(1)=\mathbf L$. Since the determinant of an element of $O(1,d)$ is either $+1$ or $-1$, continuity gives $\det\mathbf L(\lambda)=+1$ throughout the path. All configurations remain on the forward sheet, so applying $\mathbf L(\lambda)$ to the origin gives $L_{00}(\lambda)>0$; the Lorentz constraint then gives $L_{00}(\lambda)\geq1$. Hence $\mathbf L\in\mathrm{SO}^{+}(1,d)$.

For uniqueness, assume that $\mathbf L_1$ and $\mathbf L_2$ both satisfy the relation. Then
\[
\mathbf{Z}_{t,\mathrm{all}}^{s}
(\mathbf L_1^{\top}-\mathbf L_2^{\top})=\mathbf0.
\]
Since $\mathbf{Z}_{t,\mathrm{all}}^{s}$ has column rank $d+1$, its right null space contains only zero; hence $\mathbf L_1=\mathbf L_2$.
\end{proof}

\begin{corollary}[A Common Isometry for All Modalities]
\label{coro:shared L}
Under Theorem~\ref{thm:witt_multimodal}, the same $\mathbf L\in\mathrm{SO}^{+}(1,d)$ acts on both modalities:
\[
\mathbf Z_{t'}^{m_1,s'}=\mathbf Z_t^{m_1,s}\mathbf L^{\top},
\qquad
\mathbf Z_{t'}^{m_2,s'}=\mathbf Z_t^{m_2,s}\mathbf L^{\top}.
\]
If, in addition, each modality-specific block has full column rank, then any modality-specific maps $\mathbf L_1$ and $\mathbf L_2$ satisfying these two relations must obey $\mathbf L_1=\mathbf L_2=\mathbf L$.
\end{corollary}

\begin{proof}
Stacking the two modality-specific relations would produce
\[
\mathbf{Z}_{t',\mathrm{all}}^{s'}=
\begin{pmatrix}
\mathbf Z_t^{m_1,s}\mathbf L_1^{\top}\\[2pt]
\mathbf Z_t^{m_2,s}\mathbf L_2^{\top}
\end{pmatrix}.
\]
Theorem~\ref{thm:witt_multimodal} gives the unique transformation for the full stacked matrix, so both blocks admit the same restriction. Under the additional blockwise full-rank condition, subtracting the common relation from each modality-specific relation gives $\mathbf Z_t^{m_i,s}(\mathbf L_i^{\top}-\mathbf L^{\top})=\mathbf0$. The right null space is then trivial, and hence $\mathbf L_i=\mathbf L$ for $i\in\{1,2\}$.
\end{proof}

\begin{remark}
Conditions~\ref{P: intra} and~\ref{P: inter} are jointly equivalent to
\[
\mathbf{G}_{t-1,\mathrm{ext}}^{t}
=\mathbf{G}_{t-1,\mathrm{ext}}^{t-1}.
\]
Theorem~\ref{thm:witt_multimodal} and Corollary~\ref{coro:shared L} therefore show that preserving both within-modal and cross-modal relations determines one Lorentz transformation shared by all modalities.
\end{remark}

\appendixsubsection{Removing the Boost Component}

\begin{lemma}[Generation of $\mathrm{SO}^{+}(1,d)$~\cite{moretti2002interplay}]
\label{lem:SOplus_generation}
Every element of $\mathrm{SO}^{+}(1,d)$ can be written as a finite composition of Lorentz rotations and boosts.
\end{lemma}

\begin{remark}
Boosts and spatial rotations are intrinsic linear transformations of the Lorentz model: if $\mathbf x\in\mathbb H_K^d$, then both $\mathbf B\mathbf x$ and $\mathbf R\mathbf x$ remain on $\mathbb H_K^d$.
\end{remark}

\begin{thmbox}
\begin{theorem}[Elimination of Boosts; Restatement of Theorem~\ref{coro:p1_p3_synergy}]
Suppose $\mathbf L\in\mathrm{SO}^{+}(1,d)$ maps every old representation according to $\mathbf z_{t-1}^{m,t}=\mathbf L\mathbf z_{t-1}^{m,t-1}$, and the stacked old-representation matrix has full column rank. If Condition~\ref{P: hierarchical} holds for all such representations, then
\[
\mathbf L=
\begin{bmatrix}
1&\mathbf0^{\top}\\
\mathbf0&\widetilde{\mathbf R}
\end{bmatrix},
\qquad
\widetilde{\mathbf R}\in\mathrm{SO}(d),
\]
i.e., $\mathbf L$ contains no nontrivial boost.
\end{theorem}
\end{thmbox}

\begin{proof}
For every old representation on $\mathbb H_K^d$, Condition~\ref{P: hierarchical} fixes the spatial norm. The manifold constraint
\[
z_0=\sqrt{K+\|\mathbf z_{[1:d]}\|_2^2}
\]
then also fixes the time-like coordinate. In row-matrix form,
\[
\mathbf Z_{t-1,\mathrm{all}}^{t-1}
(\mathbf L^{\top}\mathbf e_0-\mathbf e_0)=\mathbf0,
\qquad \mathbf e_0=(1,0,\ldots,0)^{\top}.
\]
Full column rank makes the right null space trivial, so $\mathbf L^{\top}\mathbf e_0=\mathbf e_0$. Writing $\mathbf L$ in time--space blocks and using $\mathbf L^{\top}\mathbf G\mathbf L=\mathbf G$ therefore gives
\[
\mathbf L=\operatorname{diag}(1,\widetilde{\mathbf R}),
\qquad
\widetilde{\mathbf R}^{\top}\widetilde{\mathbf R}=\mathbf I.
\]
Finally, $\det\mathbf L=1$ implies $\det\widetilde{\mathbf R}=1$. Thus $\widetilde{\mathbf R}\in\mathrm{SO}(d)$ and no boost component remains.
\end{proof}

\appendixsubsection{First-Order Hierarchy Preservation}

The next lemma relates perturbations of the spatial and time-like coordinates on the Lorentz hyperboloid and leads directly to Corollary~\ref{cor:timelike_stability}.

\begin{thmbox}
\begin{lemma}[First-Order Change of the Time-Like Coordinate]
\label{lem:timelike_variation}
Let $\mathbf z\in\mathbb{R}^{d+1}$ obey
$z_0^2-\|\mathbf z_{[1:d]}\|_2^2=K$ with $z_0>0$. For an infinitesimal perturbation $\Delta\mathbf z$,
\begin{equation}
\Delta z_0
=\left(\frac{\mathbf z_{[1:d]}^{\top}}{z_0}\right)
\Delta\mathbf z_{[1:d]}.
\label{eq:timelike_first_order}
\end{equation}
\end{lemma}
\end{thmbox}

\begin{proof}
Define the manifold constraint
\begin{equation}
\phi(\mathbf z)=z_0^2-\|\mathbf z_{[1:d]}\|_2^2-K=0.
\label{eq:proof_constraint}
\end{equation}
For $\widetilde{\mathbf z}=\mathbf z+\Delta\mathbf z$, \revisiontext{a first-order Taylor expansion} gives
\[
\phi(\widetilde{\mathbf z})
=\phi(\mathbf z)+\nabla\phi(\mathbf z)^{\top}\Delta\mathbf z
+O(\|\Delta\mathbf z\|_2^2).
\]
Both $\mathbf z$ and $\widetilde{\mathbf z}$ lie on the manifold to first order, so
\begin{equation}
\nabla\phi(\mathbf z)^{\top}\Delta\mathbf z=0.
\label{eq:phi_linear_zero}
\end{equation}
Since
\[
\nabla\phi(\mathbf z)=
\begin{bmatrix}
2z_0\\
-2\mathbf z_{[1:d]}
\end{bmatrix},
\]
Equation~\eqref{eq:phi_linear_zero} becomes
$2z_0\Delta z_0-2\mathbf z_{[1:d]}^{\top}\Delta\mathbf z_{[1:d]}=0$. Dividing by $2z_0>0$ proves~\eqref{eq:timelike_first_order}.
\end{proof}

\begin{thmbox}
\begin{corollary}[Restatement of Corollary~\ref{cor:timelike_stability}]
Let $\Delta\mathbf z=\mathbf z_{t-1}^{m,t}-\mathbf z_{t-1}^{m,t-1}$ denote the update of an old-task embedding. If Condition~\ref{P: hierarchical} makes its time-like coordinate invariant to first order, then
\begin{equation}
\mathbf z_{[1:d]}^{\top}\Delta\mathbf z_{[1:d]}=0.
\label{eq:supp_radial_free}
\end{equation}
\end{corollary}
\end{thmbox}

\begin{proof}
Condition~\ref{P: hierarchical} sets $\Delta z_0=0$ to first order. Lemma~\ref{lem:timelike_variation} simultaneously requires
\[
\Delta z_0
=\frac{\mathbf z_{[1:d]}^{\top}\Delta\mathbf z_{[1:d]}}{z_0}.
\]
Because $z_0>0$, combining the two statements yields~\eqref{eq:supp_radial_free}.
\end{proof}

\appendixsection{Derivations for the HMCL Update}

\appendixsubsection{Proof of Proposition~\ref{prop:blockwise_admissible}}
\label{appendix: blockwise}

\begin{thmbox}
\begin{proposition}[Restatement: Admissible Parameter Changes]
Consider the Lorentz layer in Section~\ref{sec: pre-hyperbolic}, with
$\mathbf W^{m,t}=[\,\mathbf w_0^{m,t}\;\;\mathbf W_s^{m,t}\,]$. Assume that Theorem~\ref{coro:p1_p3_synergy} holds at stage $t-1$. Define
\[
\Delta\mathbf W^m=\mathbf W^{m,t}-\mathbf W^{m,t-1},
\qquad \|\Delta\mathbf W^m\|_F\rightarrow0.
\]
The collection of modal displacements is admissible to first order if and only if one skew-symmetric $\boldsymbol\Omega$ satisfies, simultaneously for every modality,
\begin{equation}
\mathbf Z_{t-1}^{m,*}(\Delta\mathbf W^m)^{\top}
=\mathbf Z_{t-1}^{m,t-1}[1\!:\!d]\boldsymbol\Omega^{\top},
\qquad
\boldsymbol\Omega^{\top}=-\boldsymbol\Omega.
\label{eq:supp_prop_exact_admissible}
\end{equation}
\end{proposition}
\end{thmbox}

\begin{proof}
The spatial output of the Lorentz layer is linear in its parameters. Its block form is
\begin{equation}
f_{[1:d]}(\mathbf W^{m,t};\mathbf Z_{t-1}^{m,*})
=\mathbf Z_{t-1}^{m,*}[0](\mathbf w_0^{m,t})^{\top}
+\mathbf Z_{t-1}^{m,*}[1{:}d](\mathbf W_s^{m,t})^{\top}.
\label{eq:supp_block_spatial_form}
\end{equation}
Consequently, the applied displacement changes the old spatial output by
\begin{align}
\mathbf Z_{t-1}^{m,t}[1\!:\!d]
&=\mathbf Z_{t-1}^{m,t-1}[1\!:\!d]
+\mathbf Z_{t-1}^{m,*}[0](\Delta\mathbf w_0^m)^{\top}
+\mathbf Z_{t-1}^{m,*}[1{:}d](\Delta\mathbf W_s^m)^{\top}.
\label{eq:block_taylor}
\end{align}

Combining the two update terms gives the compact form
\begin{equation}
\mathbf Z_{t-1}^{m,t}[1{:}d]
=\mathbf Z_{t-1}^{m,t-1}[1{:}d]
+\mathbf Z_{t-1}^{m,*}(\Delta\mathbf W^m)^{\top}.
\label{eq:supp_compact_spatial_change}
\end{equation}

Theorem~\ref{coro:p1_p3_synergy} also states that an admissible change of the old representations is a spatial rotation. In a neighborhood of the identity, write
\[
\mathbf R=\mathbf I+\boldsymbol\Omega+\mathcal O(\|\boldsymbol\Omega\|_F^2),
\qquad
\boldsymbol\Omega^{\top}=-\boldsymbol\Omega.
\]
It follows that
\begin{align}
\mathbf Z_{t-1}^{m,t}[1\!:\!d]
&=\mathbf Z_{t-1}^{m,t-1}[1\!:\!d]\mathbf R^{\top}\notag\\
&=\mathbf Z_{t-1}^{m,t-1}[1\!:\!d]
+\mathbf Z_{t-1}^{m,t-1}[1\!:\!d]\boldsymbol\Omega^{\top}
+\mathcal O(\|\boldsymbol\Omega\|_F^2).
\label{eq:block_theorem1}
\end{align}
Equating the first-order terms of~\eqref{eq:supp_compact_spatial_change} and~\eqref{eq:block_theorem1} yields
\begin{equation}
\mathbf Z_{t-1}^{m,*}(\Delta\mathbf W^m)^{\top}
=\mathbf Z_{t-1}^{m,t-1}[1\!:\!d]\boldsymbol\Omega^{\top},
\qquad
\boldsymbol\Omega^{\top}=-\boldsymbol\Omega.
\label{eq:block_admissible_spatial}
\end{equation}
The same $\boldsymbol\Omega$ follows from Theorem~\ref{coro:p1_p3_synergy} for all modalities. Conversely,~\eqref{eq:block_admissible_spatial} matches the tangent motion of that common rotation. Its spatial norm is unchanged to first order, and the reconstructed time-like coordinate in~\eqref{eq:lorentz_layer} is therefore unchanged as well.
\end{proof}

The implementation fixes the time-input column and realizes the admissible output motion through $\Delta\mathbf W_s^m$. Such a representative exists exactly when
\begin{equation}
\operatorname{col}\!\left(
\mathbf Z_{t-1}^{m,t-1}[1{:}d]\boldsymbol\Omega^{\top}
\right)
\subseteq
\operatorname{col}\!\left(\mathbf Z_{t-1}^{m,*}[1{:}d]\right).
\label{eq:supp_blockwise_representability}
\end{equation}
The regularity condition in Appendix~\ref{appendix: closest_admissible} makes the two column spaces equal, so~\eqref{eq:supp_blockwise_representability} holds for every admissible rotation. This gives the blockwise form in~\eqref{eq:prop_blockwise_admissible_spatial}.

\begin{journalnew}
\appendixsubsection{Derivation of Corollary~\ref{cor:closest_admissible}}
\label{appendix: closest_admissible}

\begin{thmbox}
\begin{corollary}[Restatement: Closest-Admissible Update]
Given candidate modal displacements
$\{\boldsymbol\delta^m=[\,\boldsymbol\delta_0^m\;\;\boldsymbol\delta_s^m\,]\}_{m=1}^{M}$,
HMCL-CA jointly selects their closest admissible displacements and one shared
rotation by solving
\begin{equation}
\begin{aligned}
\underset{\{\Delta\mathbf W^m\},\,\boldsymbol\Omega}{\operatorname{minimize}}
\quad&\frac12\sum_{m=1}^{M}
\|\Delta\mathbf W^m-\boldsymbol\delta^m\|_F^2\\
\text{subject to}\quad&
\boldsymbol\Omega^{\top}=-\boldsymbol\Omega,
\quad \Delta\mathbf w_0^m=\mathbf0,\\
&\mathbf Z_{t-1}^{m,*}[1{:}d](\Delta\mathbf W_s^m)^{\top}
=\mathbf Z_{t-1}^{m,t-1}[1{:}d]\boldsymbol\Omega^{\top},
\quad m=1,\ldots,M.
\end{aligned}
\label{eq:supp_hmcl_closest_admissible_problem}
\end{equation}
Under the regularity condition below, the modal displacements are unique and
are obtained from one Lyapunov equation for the common
$\boldsymbol\Omega^{\star}$.
\end{corollary}
\end{thmbox}

We derive the joint HMCL-CA solution and make the shared rotation explicit. For each modality, let
\begin{equation}
\mathbf X_m=\mathbf Z^{m,*}_{t-1}[1{:}d],
\qquad
\mathbf Y_m=\mathbf Z^{m,t-1}_{t-1}[1{:}d].
\end{equation}
Assume that $\mathbf Y_m$ has full column rank and that
$\operatorname{col}(\mathbf X_m)=\operatorname{col}(\mathbf Y_m)$.
Then
\begin{equation}
\mathbf X_m=\mathbf Y_m\mathbf C_m,
\qquad
\mathbf C_m=\mathbf Y_m^{\dagger}\mathbf X_m,
\label{eq:supp_ca_reparameterization}
\end{equation}
where $\mathbf C_m$ is invertible. Proposition~\ref{prop:blockwise_admissible} now gives
\begin{equation}
\mathbf C_m(\Delta\mathbf W_s^m)^{\top}
=\boldsymbol\Omega^{\top}
\quad\Longleftrightarrow\quad
\Delta\mathbf W_s^m
=\boldsymbol\Omega\mathbf C_m^{-\top}.
\label{eq:supp_ca_modal_update}
\end{equation}
The same $\boldsymbol\Omega$ occurs for every $m$. Substituting~\eqref{eq:supp_ca_modal_update} into the closest-admissible problem yields
\begin{equation}
\min_{\boldsymbol\Omega^{\top}=-\boldsymbol\Omega}
\frac12\sum_{m=1}^{M}
\left\|\boldsymbol\Omega\mathbf C_m^{-\top}
-\boldsymbol\delta_s^m\right\|_F^2.
\label{eq:supp_ca_rotation_problem}
\end{equation}
Define
\begin{equation}
\mathbf H=\sum_{m=1}^{M}\mathbf C_m^{-\top}\mathbf C_m^{-1},
\qquad
\mathbf F=\sum_{m=1}^{M}\boldsymbol\delta_s^m\mathbf C_m^{-1}.
\label{eq:supp_ca_lyapunov_matrices}
\end{equation}
Since every $\mathbf C_m$ is invertible, $\mathbf H$ is positive definite. Stationarity of~\eqref{eq:supp_ca_rotation_problem} over the skew-symmetric matrices gives the Lyapunov equation
\begin{equation}
\boldsymbol\Omega^{\star}\mathbf H
+\mathbf H\boldsymbol\Omega^{\star}
=2\operatorname{Skew}(\mathbf F),
\qquad
\operatorname{Skew}(\mathbf F)=\frac12(\mathbf F-\mathbf F^{\top}).
\label{eq:supp_ca_lyapunov}
\end{equation}
Positive definiteness of $\mathbf H$ gives a unique skew-symmetric solution. The closest admissible displacements are therefore
\begin{equation}
\boldsymbol\delta_s^{\mathrm{CA},m}
=\boldsymbol\Omega^{\star}\mathbf C_m^{-\top},
\qquad
\boldsymbol\delta_0^{\mathrm{CA},m}=\mathbf0,
\qquad m=1,\ldots,M.
\label{eq:ca_poststep_solution}
\end{equation}
This construction solves for one rotation from all modal candidates. In the optimizer-consistent update, each $\boldsymbol\delta_s^m$ is the spatial block of the displacement proposed by the chosen optimizer.
\end{journalnew}

\appendixsubsection{Proof of Corollary~\ref{cor:canonical_admissible}}
\label{appendix: canonical}

\begin{thmbox}
\begin{corollary}[HMCL-MR: Minimal-Rotation Update (Restatement)]
Let $\boldsymbol\delta_s^m$ be an unconstrained candidate displacement and impose Proposition~\ref{prop:blockwise_admissible} while penalizing the induced rotation. In the minimum-rotation limit, $\boldsymbol\Omega^{\star}=\mathbf0$ and
\begin{equation}
\mathbf Z_{t-1}^{m,*}[1{:}d](\Delta\mathbf W_s^m)^{\top}=\mathbf0,
\qquad
\Delta\mathbf w_0^m=\mathbf0.
\label{eq:cor_optimal_null}
\end{equation}
The canonical solution is the null-space projection
\begin{equation}
\Delta\mathbf W_s^m
=\boldsymbol\delta_s^m(\mathbf I-\mathbf P_{t-1}),
\qquad
\mathbf P_{t-1}
=(\mathbf Z_{t-1}^{m,*}[1{:}d])^{\top}
\left(\mathbf Z_{t-1}^{m,*}[1{:}d](\mathbf Z_{t-1}^{m,*}[1{:}d])^{\top}\right)^{\dagger}
\mathbf Z_{t-1}^{m,*}[1{:}d].
\label{eq:optimal_projection}
\end{equation}
\end{corollary}
\end{thmbox}

\begin{proof}
Begin with the general admissible family
\begin{equation}
\mathbf Z_{t-1}^{m,*}[1{:}d](\Delta\mathbf W_s^m)^{\top}
=\mathbf Z_{t-1}^{m,t-1}[1\!:\!d]\boldsymbol\Omega^{\top},
\qquad
\boldsymbol\Omega^{\top}=-\boldsymbol\Omega.
\label{eq:app_admissible_general}
\end{equation}
Among these displacements, consider the rotation-regularized closest point to $\boldsymbol\delta_s^m$:
\begin{equation}
\min_{\Delta\mathbf W_s^m,\boldsymbol\Omega}
\frac12\|\Delta\mathbf W_s^m-\boldsymbol\delta_s^m\|_F^2
+\frac{\lambda}{2}\|\boldsymbol\Omega\|_F^2
\quad\text{s.t.}\quad\eqref{eq:app_admissible_general}.
\label{eq:app_opt_problem}
\end{equation}
With multiplier $\boldsymbol\Lambda$, the Lagrangian is
\begin{align}
\mathcal L
&=\frac12\|\Delta\mathbf W_s^m-\boldsymbol\delta_s^m\|_F^2
+\frac{\lambda}{2}\|\boldsymbol\Omega\|_F^2\notag\\
&\quad+\left\langle\boldsymbol\Lambda,
\mathbf Z_{t-1}^{m,*}[1{:}d](\Delta\mathbf W_s^m)^{\top}
-\mathbf Z_{t-1}^{m,t-1}[1\!:\!d]\boldsymbol\Omega^{\top}
\right\rangle.
\end{align}
Stationarity gives
\begin{align}
\Delta\mathbf W_s^m
&=\boldsymbol\delta_s^m-\boldsymbol\Lambda^{\top}\mathbf Z_{t-1}^{m,*}[1{:}d],
\label{eq:app_stationary_W}\\
\boldsymbol\Omega
&=\frac1\lambda\operatorname{Skew}\!\left(
(\mathbf Z_{t-1}^{m,t-1}[1\!:\!d])^{\top}\boldsymbol\Lambda
\right),
\label{eq:app_stationary_Omega}
\end{align}
where $\operatorname{Skew}(\mathbf A)=(\mathbf A-\mathbf A^{\top})/2$. Substituting~\eqref{eq:app_stationary_W} into~\eqref{eq:app_admissible_general} yields
\begin{equation}
\mathbf Z_{t-1}^{m,*}[1{:}d](\mathbf Z_{t-1}^{m,*}[1{:}d])^{\top}\boldsymbol\Lambda
=\mathbf Z_{t-1}^{m,*}[1{:}d](\boldsymbol\delta_s^m)^{\top}
+\mathcal O(\lambda^{-1}).
\label{eq:app_lambda_eq}
\end{equation}
In the limit $\lambda\rightarrow\infty$, the rotation term vanishes and
\begin{equation}
\boldsymbol\Lambda
=\left(\mathbf Z_{t-1}^{m,*}[1{:}d](\mathbf Z_{t-1}^{m,*}[1{:}d])^{\top}\right)^{\dagger}
\mathbf Z_{t-1}^{m,*}[1{:}d](\boldsymbol\delta_s^m)^{\top}.
\label{eq:app_lambda_sol}
\end{equation}
Inserting this expression into~\eqref{eq:app_stationary_W} produces
\[
\Delta\mathbf W_s^m
=\boldsymbol\delta_s^m\left[
\mathbf I-(\mathbf Z_{t-1}^{m,*}[1{:}d])^{\top}
\left(\mathbf Z_{t-1}^{m,*}[1{:}d](\mathbf Z_{t-1}^{m,*}[1{:}d])^{\top}\right)^{\dagger}
\mathbf Z_{t-1}^{m,*}[1{:}d]
\right].
\]
Equation~\eqref{eq:app_stationary_Omega} simultaneously gives $\boldsymbol\Omega^{\star}=\mathbf0$. Combining this spatial projection with $\Delta\mathbf w_0^m=\mathbf0$ proves the stated canonical update.
\end{proof}

\begin{journalnew}
\appendixsection{Optimizer-Consistent HMCL: Derivations}
\label{apdx:optimizer_consistent}

This appendix derives the direction--displacement mismatch, Lorentz defect,
finite-step preservation result, feasible accumulation, and low-rank residual
bound for the unified HMCL update. Throughout a fixed task, we omit the stage
index $t$ and retain the optimizer-step index $k$; in $(k,0)$ and $(k,s)$, the
second entry \revisiontext{identifies the time-like input column ($0$) or the spatial input block ($s$)},
whereas $\mathbf W_0^{m,t}$ denotes the task-start parameters. As established
in the main paper, a common spatial rotation preserves the three HMCL
conditions: CA estimates its skew generator $\boldsymbol\Omega$, while MR
selects the identity member $\boldsymbol\Omega=\mathbf0$. Both corrections act
on the fully formed candidate displacement.

\appendixsubsection{A Concrete Adaptive-Optimizer Mismatch}

Let $\mathcal S\subseteq\mathbb R^q$ be the linear subspace of admissible vectorized parameter displacements and let $\boldsymbol\Pi$ be its orthogonal projector. For a positive diagonal matrix $\mathbf D_{\mathrm{opt}}$, consider the preconditioned component of an AdamW step,
\begin{equation}
\mathbf d_{\mathrm{pre}}=-\eta\mathbf D_{\mathrm{opt}}\mathbf u,
\qquad \mathbf u\in\mathcal S.
\label{eq:supp_preconditioned_step}
\end{equation}

\begin{proposition}[Direction--Displacement Mismatch]
\label{prop:optimizer_mismatch}
Suppose AdamW maps an admissible first-moment direction $\mathbf u\in\mathcal S$ to
\begin{equation}
\mathbf d^{\mathrm{AdamW}}=-\eta\mathbf D_{\mathrm{opt}}\mathbf u-\eta\omega\boldsymbol\theta,
\end{equation}
where $\mathbf D_{\mathrm{opt}}$ is a positive diagonal preconditioner and $\omega$ is the decoupled weight-decay coefficient. The preconditioned component remains admissible for every $\mathbf u\in\mathcal S$ if and only if
\begin{equation}
(\mathbf I-\boldsymbol\Pi)\mathbf D_{\mathrm{opt}}\boldsymbol\Pi=\mathbf0.
\label{eq:supp_commuting_condition}
\end{equation}
The decay component is admissible exactly when $\omega=0$ or
$\boldsymbol\theta\in\mathcal S$.
\end{proposition}

\begin{proof}[Proof of Proposition~\ref{prop:optimizer_mismatch}]
Because $\boldsymbol\Pi$ projects onto $\mathcal S$, every admissible direction can be written as $\mathbf u=\boldsymbol\Pi\mathbf a$ for some $\mathbf a$. The preconditioned direction remains in $\mathcal S$ for every admissible input if and only if its component in $\mathcal S^{\perp}$ vanishes:
\begin{align}
(\mathbf I-\boldsymbol\Pi)\mathbf D_{\mathrm{opt}}\mathbf u
&=(\mathbf I-\boldsymbol\Pi)\mathbf D_{\mathrm{opt}}\boldsymbol\Pi\mathbf a
=\mathbf0
\quad\text{for every }\mathbf a.
\end{align}
This is equivalent to
$(\mathbf I-\boldsymbol\Pi)\mathbf D_{\mathrm{opt}}\boldsymbol\Pi=\mathbf0$.
The decoupled decay term is $-\eta\omega\boldsymbol\theta$. It lies in $\mathcal S$ exactly when $\omega=0$ or $\boldsymbol\theta\in\mathcal S$. Adding either non-admissible component to an admissible one does not restore the constraint in general.
\end{proof}

\medskip\noindent\textbf{A two-dimensional counterexample.}
Let $\mathcal S=\operatorname{span}([1,1]^{\top})$ and choose the admissible direction $\mathbf u=[1,1]^{\top}$. With $\mathbf D_{\mathrm{opt}}=\operatorname{diag}(1,2)$ and no decay, AdamW-like scaling gives
\[
\mathbf d_{\mathrm{pre}}=-\eta[1,2]^{\top},
\]
which is not in $\mathcal S$. Equivalently, the normal vector
$\mathbf v_{\perp}=[1,-1]^{\top}$ satisfies $\mathbf v_{\perp}^{\top}\mathbf u=0$ but
$\mathbf v_{\perp}^{\top}\mathbf d_{\mathrm{pre}}=\eta\ne0$. The example is intentionally small: the same non-commutativity occurs between a low-rank representation projector and AdamW's element-wise preconditioner in the Lorentz head.

\appendixsubsection{Lorentz Strain of a Realized Optimizer Candidate}

We now specialize the mismatch to the Lorentz representation geometry. For modality $m$, let $\mathbf Y^m\in\mathbb R^{N_m\times d}$ contain the old spatial coordinates as rows, and let $\Delta\mathbf Y^m$ contain their first-order changes under the realized optimizer candidate. Define
\begin{equation}
\begin{aligned}
\mathbf B^m
&=\arg\min_{\mathbf B}
\|\Delta\mathbf Y^m-\mathbf Y^m\mathbf B^{\top}\|_F^2,\\
\mathbf E^m
&=\Delta\mathbf Y^m-\mathbf Y^m(\mathbf B^m)^{\top}.
\end{aligned}
\label{eq:lorentz_generator_fit}
\end{equation}
Thus
$\Delta\mathbf Y^m=\mathbf Y^m(\mathbf B^m)^{\top}+\mathbf E^m$.
Write $\mathbf B^m=\mathbf S^m+\boldsymbol\Omega^m$, where
$\mathbf S^m=\operatorname{Sym}(\mathbf B^m)$ and
$\boldsymbol\Omega^m=\operatorname{Skew}(\mathbf B^m)$.
For a row $i$, we use the equivalent column-vector form
$\Delta\mathbf y_i^m=\mathbf B^m\mathbf y_i^m+\mathbf e_i^m$.

\begin{proposition}[Lorentz Defect of a Realized Step]
Under the rank condition of Theorem~\ref{coro:p1_p3_synergy}, a realized first-order change preserves Conditions~\textbf{\ref{P: intra}}--\textbf{\ref{P: hierarchical}} if and only if
\begin{equation}
\mathbf E^m=\mathbf0,
\qquad
\mathbf S^m=\mathbf0,
\qquad
\boldsymbol\Omega^m=\boldsymbol\Omega^{m'}
\quad\text{for every }m,m'.
\label{eq:lorentz_strain_zero}
\end{equation}
Equivalently, the change is induced by one shared skew-symmetric generator.
\end{proposition}

\begin{proof}
Suppose first that the realized change preserves Conditions~\textbf{\ref{P: intra}}--\textbf{\ref{P: hierarchical}} to first order. Proposition~\ref{prop:blockwise_admissible} then gives one skew-symmetric generator $\boldsymbol\Omega$ shared by all modalities, so
\begin{equation}
\Delta\mathbf Y^m=\mathbf Y^m\boldsymbol\Omega^{\top}
\qquad\text{for every }m.
\label{eq:supp_shared_skew}
\end{equation}
Under the stated rank condition, the least-squares generator in~\eqref{eq:lorentz_generator_fit} is unique. Hence
$\mathbf E^m=\mathbf0$, $\mathbf B^m=\boldsymbol\Omega$, and
$\operatorname{Sym}(\mathbf B^m)=\mathbf0$ for every modality.

Conversely, suppose~\eqref{eq:lorentz_strain_zero} holds. Then all modalities satisfy~\eqref{eq:supp_shared_skew} for the same \revisiontext{skew-symmetric matrix} $\boldsymbol\Omega$. For an old spatial coordinate $\mathbf y$, its induced time-like derivative under the reconstruction in~\eqref{eq:lorentz_layer} is
\begin{equation}
\dot z_0
=\frac{\mathbf y^{\top}\boldsymbol\Omega\mathbf y}
{\sqrt{K+\|\mathbf y\|_2^2}}=0.
\end{equation}
The first-order change of a spatial inner product is likewise
\begin{equation}
(\boldsymbol\Omega\mathbf y_i)^{\top}\mathbf y_j
+\mathbf y_i^{\top}(\boldsymbol\Omega\mathbf y_j)
=\mathbf y_i^{\top}
(\boldsymbol\Omega^{\top}+\boldsymbol\Omega)\mathbf y_j=0.
\end{equation}
The same cancellation applies across modalities because the generator is shared. The Lorentz Gram matrices and spatial radii are therefore stationary to first order, establishing Conditions~\textbf{\ref{P: intra}}--\textbf{\ref{P: hierarchical}}.
\end{proof}

A compact measure of the departure from this shared tangent direction is
\begin{equation}
\mathcal D_{\mathcal L}
=\sum_m\big(\|\mathbf S^m\|_F^2+\|\mathbf E^m\|_F^2\big)
+\sum_{m<m'}\|\boldsymbol\Omega^m-\boldsymbol\Omega^{m'}\|_F^2.
\label{eq:lorentz_admissibility_defect}
\end{equation}
It vanishes exactly for a shared first-order rotation under the same assumptions.

\medskip\noindent\textbf{Radial and cone-aperture drift.}
Let $r=\|\mathbf y\|_2$ and
$\mathbf a=\mathbf S\mathbf y+\boldsymbol\Omega\mathbf y+\mathbf e$.
A Taylor expansion and
$\mathbf y^{\top}\boldsymbol\Omega\mathbf y=0$ give
\begin{equation}
\Delta r
=\frac{\mathbf y^{\top}(\mathbf S\mathbf y+\mathbf e)}{r}
+\mathcal O(\|\mathbf a\|_2^2).
\label{eq:optimizer_radial_drift}
\end{equation}
For $h(r)=\sin^{-1}(2\kappa/r)$,
\begin{equation}
h'(r)=-\frac{2\kappa}{r\sqrt{r^2-4\kappa^2}},
\end{equation}
and therefore
\begin{equation}
\Delta\operatorname{aper}(\mathbf y)
=-\frac{2\kappa}{r\sqrt{r^2-4\kappa^2}}\Delta r
+\mathcal O((\Delta r)^2).
\label{eq:optimizer_aperture_drift}
\end{equation}
In particular, decoupled decay has local generator
$\mathbf B_{\mathrm{decay}}=-\eta\omega\mathbf I$. It produces
$\Delta r=-\eta\omega r+\mathcal O((\eta\omega)^2)$ and therefore increases the cone aperture to first order. This calculation is a geometric interpretation of the decay term, not an explanation of the main-table results, which use zero weight decay; Appendix~\ref{apdx:adamw_weight_decay} separately evaluates nonzero decay.

\medskip\noindent\textbf{Lorentz Gram drift.}
Let
$\mathbf z_i=[z_{i,0};\mathbf y_i]$ and
$\dot{\mathbf z}_i=[\dot z_{i,0};\mathbf a_i]$. The first variation of an intra- or inter-modal Lorentz product is
\begin{equation}
\frac{\mathrm d}{\mathrm d\epsilon}
\left.
\big(\mathbf z_i+\epsilon\dot{\mathbf z}_i\big)^{\top}
\mathbf G
\big(\mathbf z_j+\epsilon\dot{\mathbf z}_j\big)
\right|_{\epsilon=0}
=\dot{\mathbf z}_i^{\top}\mathbf G\mathbf z_j
+\mathbf z_i^{\top}\mathbf G\dot{\mathbf z}_j.
\label{eq:supp_lorentz_gram_variation}
\end{equation}
For one shared skew generator, both time-like derivatives vanish and the spatial terms cancel. Symmetric strain, residual motion, or different modality generators remove this cancellation. Equation~\eqref{eq:supp_lorentz_gram_variation} therefore links the optimizer defect in~\eqref{eq:lorentz_admissibility_defect} directly to the similarity matrices used by the contrastive objective.

\appendixsubsection{Closest Projection of the Realized Step}

For one modality, omit the superscript $m$ and split the optimizer candidate
at step $k$ into the time-like input column
$\boldsymbol\delta^{\mathrm{opt}}_{k,0}$ and spatial input block
$\boldsymbol\delta^{\mathrm{opt}}_{k,s}$. Let
$\mathbf P=\mathbf V\mathbf V^{\top}$ be an orthogonal projector and
$\mathbf Q=\mathbf I-\mathbf P$. We write
$\boldsymbol\delta_k=[\,\boldsymbol\delta_{k,0}\;\;
\boldsymbol\delta_{k,s}\,]$ for a generic displacement at the same step.
On the upper sheet of $\mathbb H_K^d$, zero Lorentz geodesic distance is
equivalent to equality of the two points. Because the spatial block
of~\eqref{eq:lorentz_layer} is linear in its input and the time-input
parameter block is fixed, the canonical zero-motion condition used
in~\eqref{eq:canonical_projection} is equivalent to
$\boldsymbol\delta_{k,s}\mathbf P=\mathbf0$ and
$\boldsymbol\delta_{k,0}=\mathbf0$. The following theorem and corollary establish
the closest-point and finite-step preservation claims used in
Theorem~\ref{thm:optimizer_consistent_closest_update}.

\begin{theorem}[Closest Identity-Isometry Representative]
\label{thm:closest_poststep}
Among displacements satisfying
$\boldsymbol\delta_{k,s}\mathbf P=\mathbf0$ and
$\boldsymbol\delta_{k,0}=\mathbf0$, the unique displacement closest to the realized optimizer candidate is
\begin{equation}
\boldsymbol\delta^{\star}_{k,s}
=\boldsymbol\delta^{\mathrm{opt}}_{k,s}(\mathbf I-\mathbf P),
\qquad
\boldsymbol\delta^{\star}_{k,0}=\mathbf0.
\label{eq:poststep_closed_form}
\end{equation}
\end{theorem}

\begin{proof}[Proof of Theorem~\ref{thm:closest_poststep}]
Every spatial matrix has the orthogonal decomposition
\begin{equation}
\boldsymbol\delta^{\mathrm{opt}}_{k,s}
=\boldsymbol\delta^{\mathrm{opt}}_{k,s}\mathbf P
+\boldsymbol\delta^{\mathrm{opt}}_{k,s}\mathbf Q.
\label{eq:supp_candidate_decomposition}
\end{equation}
The two terms are orthogonal in the Frobenius inner product because
$\mathbf P\mathbf Q=\mathbf0$. A feasible spatial displacement $\mathbf D_{k,s}$ satisfies $\mathbf D_{k,s}\mathbf P=\mathbf0$ and therefore $\mathbf D_{k,s}=\mathbf D_{k,s}\mathbf Q$. Hence
\begin{align}
\|\mathbf D_{k,s}-\boldsymbol\delta^{\mathrm{opt}}_{k,s}\|_F^2
&=\|\mathbf D_{k,s}-\boldsymbol\delta^{\mathrm{opt}}_{k,s}\mathbf Q\|_F^2
+\|\boldsymbol\delta^{\mathrm{opt}}_{k,s}\mathbf P\|_F^2.
\end{align}
The second term is independent of $\mathbf D_{k,s}$, while the first is uniquely minimized by
$\mathbf D_{k,s}=\boldsymbol\delta^{\mathrm{opt}}_{k,s}\mathbf Q$. The time-like constraint has the unique closest value $\mathbf d_{k,0}=\mathbf0$. Combining the two blocks gives~\eqref{eq:poststep_closed_form}.
\end{proof}

This proof is independent of how the optimizer constructs $\boldsymbol\delta^{\mathrm{opt}}_k$. Learning rates, moment estimates, preconditioners, and decay affect the candidate, while the Lorentz-derived projection solves the same closest-point problem for the displacement actually proposed.

\begin{corollary}[Exact Protected Lorentz Invariants]
\label{cor:exact_protected_lorentz}
If an old frozen feature $\mathbf x$ lies in the range of $\mathbf P_{t-1}$, the update in~\eqref{eq:poststep_closed_form} leaves its complete Lorentz representation unchanged after each finite step:
\begin{equation}
\mathbf z_{k+1}^{m}(\mathbf x)=\mathbf z_{k}^{m}(\mathbf x).
\label{eq:exact_protected_lorentz}
\end{equation}
Hence its hyperbolic relations, spatial radius, and entailment-cone aperture are unchanged.
\end{corollary}

\begin{proof}[Proof of Corollary~\ref{cor:exact_protected_lorentz}]
Let $\mathbf x=\mathbf P_{t-1}\mathbf x$. Equation~\eqref{eq:poststep_closed_form} gives
\begin{equation}
\boldsymbol\delta^{\star}_{k,s}\mathbf x
=\boldsymbol\delta^{\mathrm{opt}}_{k,s}
(\mathbf I-\mathbf P_{t-1})\mathbf P_{t-1}\mathbf x
=\mathbf0,
\end{equation}
and the time-like input block is fixed. The spatial output of the Lorentz layer is therefore identical before and after the corrected step. Its reconstructed first coordinate, which is a deterministic function of the spatial norm, is identical as well. Thus each protected Lorentz point is unchanged. Every intra- and inter-modal Lorentz Gram entry, hyperbolic distance, spatial radius, and cone aperture formed from these points is consequently unchanged.
\end{proof}

\appendixsubsection{Task-Anchored Feasible Contraction}

For a protected representation system fixed throughout task $t$, let
$\mathcal A_{t-1}$ be the joint admissible family in
\eqref{eq:hmcl_closest_admissible_problem}. It is linear because its defining
constraints are linear in the modal displacements and their shared skew
generator.

For a modal tuple $\mathcal U=(\mathbf U^m)_{m=1}^{M}$, define
$\|\mathcal U\|_{\oplus F}
=\big(\sum_m\|\mathbf U^m\|_F^2\big)^{1/2}$.
\revisiontext{Here $u_{\max}$ bounds the realized AdamW displacement rather than the raw gradient. Its matrix-wide value can grow with the number of modal parameters even when the per-parameter step is small; the result requires only a finite bound, not $u_{\max}\leq1$, although a larger value makes the numerical accumulation bound looser.}

\begin{corollary}[Within-Stage Admissibility and Bounded Accumulation]
\label{cor:poststep_accumulation}
Let $\mathbf A_0^m=\mathbf0$ for every modality $m$ denote the initial
displacement from the task-start parameters. Given optimizer candidates
$\boldsymbol\delta_k^{\mathrm{opt},t}$, let
$\mathbf D_k^t=\mathcal P^{\mathcal L}_{t-1}
(\boldsymbol\delta_k^{\mathrm{opt},t})=(\mathbf D_k^m)_{m=1}^{M}$ and define
\begin{equation}
\mathbf A_{k+1}^m=(1-\beta)(\mathbf A_k^m+\mathbf D_k^m),
\qquad 0\leq\beta<1.
\label{eq:supp_anchored_recurrence}
\end{equation}
Then $\{\mathbf A_k^m\}_{m=1}^{M}\in\mathcal A_{t-1}$ for every $k$.
Relative to the uncontracted candidate, the task-start distance is reduced by
$1-\beta$. The projection is non-expansive:
$\|\mathbf D_k^t\|_{\oplus F}
\leq\|\boldsymbol\delta_k^{\mathrm{opt},t}\|_{\oplus F}$.
If the latter is at most $u_{\max}$ for all $k$, then
\begin{equation}
\begin{aligned}
\|\mathbf A_J^t\|_{\oplus F}
&\leq\sum_{j=0}^{J-1}(1-\beta)^{J-j}\|\mathbf D_j^t\|_{\oplus F}\\
&\leq
\begin{cases}
Ju_{\max}, & \beta=0,\\[2pt]
\dfrac{(1-\beta)[1-(1-\beta)^J]}{\beta}\,u_{\max},
&0<\beta<1,
\end{cases}\\[-1pt]
&\leq\dfrac{1-\beta}{\beta}\,u_{\max} \qquad (0<\beta<1).
\end{aligned}
\label{eq:supp_anchored_bound}
\end{equation}
\end{corollary}

\begin{proof}[Proof of Corollary~\ref{cor:poststep_accumulation}]
The set $\mathcal A_{t-1}$ is a linear subspace, so projection places every
$\mathbf D_k^t$ in it and induction keeps every $\mathbf A_k^t$ in it. As the
orthogonal projection onto this subspace, $\mathcal P^{\mathcal L}_{t-1}$ is
non-expansive in the joint Frobenius norm. Unrolling the recurrence gives
\[
\mathbf A_J^t=\sum_{j=0}^{J-1}(1-\beta)^{J-j}\mathbf D_j^t.
\]
The triangle inequality gives the data-dependent bound in
\eqref{eq:supp_anchored_bound}; substituting the proposal bound $u_{\max}$ and summing
the geometric series gives its two stated cases.
\end{proof}

\appendixsubsection{Residual Drift under Low-Rank PCA}

Let $\mathbf X\in\mathbb R^{n\times p}$ contain the old frozen features, and let
$\boldsymbol\Sigma=\mathbf X^{\top}\mathbf X/n$ have eigenvalues
$\lambda_1\ge\cdots\ge\lambda_p\ge0$. Let $\mathbf P_r$ be the orthogonal projector onto the top-$r$ eigenspace and $\mathbf Q_r=\mathbf I-\mathbf P_r$. For the zero-rotation restriction, $\mathbf D_{k,s}=\mathbf D_{k,s}\mathbf Q_r$, so
\begin{align}
\frac1n\|\mathbf X\mathbf D_{k,s}^{\top}\|_F^2
&=\frac1n\|\mathbf X\mathbf Q_r\mathbf D_{k,s}^{\top}\|_F^2 \\
&\le \|\mathbf D_{k,s}\|_2^2
\frac1n\|\mathbf X\mathbf Q_r\|_F^2 \\
&=\|\mathbf D_{k,s}\|_2^2\sum_{i>r}\lambda_i.
\label{eq:pca_tail_bound}
\end{align}
Thus the update is exact on the retained subspace, and its \revisiontext{mean-squared effect} outside that subspace is bounded by the PCA spectral tail. This is an average-energy statement. It does not bound the worst example or the worst earlier task.

The task-anchored update retains the same subspace property. The joint bound
also applies to every modal block; in particular,
$\mathbf A_{J,s}=\mathbf A_{J,s}\mathbf Q_r$. Combining
~\eqref{eq:supp_anchored_bound} with~\eqref{eq:pca_tail_bound} gives, for
$0<\beta<1$,
\begin{equation}
\frac1n\|\mathbf X\mathbf A_{J,s}^{\top}\|_F^2
\leq
\left(\frac{1-\beta}{\beta}u_{\max}\right)^2
\sum_{i>r}\lambda_i.
\label{eq:anchored_pca_tail_bound}
\end{equation}
The geometric correction controls the retained subspace exactly, whereas the
anchor and the spectral tail jointly bound the average residual motion outside
that subspace.

\appendixsubsection{Pseudocode, Optimizer State, and Complexity}

Algorithms~\ref{alg:optimizer_consistent_mr}
and~\ref{alg:optimizer_consistent_ca} instantiate the same post-optimizer,
task-anchored update. They differ only in the closest-admissible correction:
HMCL-MR fixes the shared rotation to zero, whereas HMCL-CA estimates one
rotation jointly from all modal candidates.

\begin{algorithm}[H]
\caption{Optimizer-Consistent HMCL-MR}
\label{alg:optimizer_consistent_mr}
\KwIn{Current-task data $\mathcal D_t$; task-start Lorentz heads
$\{\mathbf W^{m,t}_{0}\}_{m=1}^{M}$; protected projector
$\mathbf P_{t-1}$; optimizer $\operatorname{Opt}$; anchoring coefficient
$\beta$.}
\KwOut{Final task parameters $\{\mathbf W^{m,t}_{J}\}_{m=1}^{M}$.}

\For{optimizer step $k=0,\ldots,J-1$}{
Compute the HMCL loss on $\mathcal D_t$ and gradients
$\{\mathbf g^{m,t}_{k}\}_{m=1}^{M}$\;
Let $\operatorname{Opt}$ update its state and propose
$\{\widetilde{\mathbf W}^{m,t}_{k+1}\}_{m=1}^{M}$\;
\For{$m=1,\ldots,M$}{
$\boldsymbol\delta^{\mathrm{opt},m,t}_{k}\leftarrow
\widetilde{\mathbf W}^{m,t}_{k+1}-\mathbf W^{m,t}_{k}$\;
$\boldsymbol\delta^{\mathrm{MR},m,t}_{k,0}\leftarrow\mathbf0$,
$\boldsymbol\delta^{\mathrm{MR},m,t}_{k,s}\leftarrow
\boldsymbol\delta^{\mathrm{opt},m,t}_{k,s}
(\mathbf I-\mathbf P_{t-1})$\;
$\widehat{\mathbf W}^{m,t}_{k+1}\leftarrow
\mathbf W^{m,t}_{k}+\boldsymbol\delta^{\mathrm{MR},m,t}_{k}$\;
$\mathbf W^{m,t}_{k+1}\leftarrow
\mathbf W^{m,t}_{0}+(1-\beta)
(\widehat{\mathbf W}^{m,t}_{k+1}-\mathbf W^{m,t}_{0})$\;
}
}
\Return $\{\mathbf W^{m,t}_{J}\}_{m=1}^{M}$.
\end{algorithm}

\begin{algorithm}[H]
\caption{Optimizer-Consistent HMCL-CA}
\label{alg:optimizer_consistent_ca}
\KwIn{Current-task data $\mathcal D_t$; task-start Lorentz heads
$\{\mathbf W^{m,t}_{0}\}_{m=1}^{M}$; protected matrices
$\{\mathbf X_m,\mathbf Y_m\}_{m=1}^{M}$ from
Appendix~\ref{appendix: closest_admissible}; optimizer
$\operatorname{Opt}$; anchoring coefficient $\beta$.}
\KwOut{Final task parameters $\{\mathbf W^{m,t}_{J}\}_{m=1}^{M}$.}

\For{$m=1,\ldots,M$}{
$\mathbf C_m\leftarrow\mathbf Y_m^{\dagger}\mathbf X_m$\;
}
$\mathbf H\leftarrow
\sum_{m=1}^{M}\mathbf C_m^{-\top}\mathbf C_m^{-1}$\;
\For{optimizer step $k=0,\ldots,J-1$}{
Compute the HMCL loss on $\mathcal D_t$ and gradients
$\{\mathbf g^{m,t}_{k}\}_{m=1}^{M}$\;
Let $\operatorname{Opt}$ update its state and propose
$\{\widetilde{\mathbf W}^{m,t}_{k+1}\}_{m=1}^{M}$\;
\For{$m=1,\ldots,M$}{
$\boldsymbol\delta^{\mathrm{opt},m,t}_{k}\leftarrow
\widetilde{\mathbf W}^{m,t}_{k+1}-\mathbf W^{m,t}_{k}$\;
}
$\mathbf F_k\leftarrow
\sum_{m=1}^{M}\boldsymbol\delta^{\mathrm{opt},m,t}_{k,s}
\mathbf C_m^{-1}$\;
Solve $\boldsymbol\Omega_k^{\star}\mathbf H
+\mathbf H\boldsymbol\Omega_k^{\star}
=2\operatorname{Skew}(\mathbf F_k)$\;
\For{$m=1,\ldots,M$}{
$\boldsymbol\delta^{\mathrm{CA},m,t}_{k,0}\leftarrow\mathbf0$,
$\boldsymbol\delta^{\mathrm{CA},m,t}_{k,s}\leftarrow
\boldsymbol\Omega_k^{\star}\mathbf C_m^{-\top}$\;
$\widehat{\mathbf W}^{m,t}_{k+1}\leftarrow
\mathbf W^{m,t}_{k}+\boldsymbol\delta^{\mathrm{CA},m,t}_{k}$\;
$\mathbf W^{m,t}_{k+1}\leftarrow
\mathbf W^{m,t}_{0}+(1-\beta)
(\widehat{\mathbf W}^{m,t}_{k+1}-\mathbf W^{m,t}_{0})$\;
}
}
\Return $\{\mathbf W^{m,t}_{J}\}_{m=1}^{M}$.
\end{algorithm}

For $\mathbf W_{k,s}\in\mathbb R^{d\times p}$, the HMCL-MR correction
$\boldsymbol\delta_{k,s}\leftarrow
\boldsymbol\delta_{k,s}-(\boldsymbol\delta_{k,s}\mathbf V)\mathbf V^{\top}$
costs $\mathcal O(dpr)$ and stores $\mathcal O(pr)$ values. The dense HMCL-CA
solver adds one $d\times d$ Lyapunov equation per step, with
$\mathcal O(d^3)$ time; a rank-$r$ realization reduces this solve to
$\mathcal O(r^3)$. The PCA basis and the task-level CA factors are estimated
once per stage. \revisiontext{The optimizer state remains the state produced by the raw
gradients; for AdamW, this state includes the first- and second-moment estimates.}
The guarantees concern the realized parameter trajectory rather than a
transported optimizer state.
\end{journalnew}

\definecolor{headergray}{RGB}{215, 221, 229}
\definecolor{rulegray}{RGB}{166, 176, 187}
\definecolor{bluedark}{RGB}{15, 77, 146}
\definecolor{greendark}{RGB}{36, 107, 75}

\definecolor{darkblue}{RGB}{201, 223, 245}
\definecolor{middleblue}{RGB}{221, 234, 248}
\definecolor{lightblue}{RGB}{234, 241, 248}

\definecolor{darkgreen}{RGB}{196, 223, 206}
\definecolor{middlegreen}{RGB}{217, 234, 223}
\definecolor{lightgreen}{RGB}{232, 243, 236}

\appendixsection{Additional Implementation Details}

\appendixsubsection{Dataset Details and Task Stream}
\label{apdx:datasets}

The unified journal benchmark contains 16 datasets and mixes image--text classification with bidirectional cross-modal retrieval.

\subsubsection{Classification Datasets}
The classification portion includes CIFAR-10 and CIFAR-100~\citep{krizhevsky2009learning}, Caltech-101~\citep{fei2004learning}, ImageNet-1K~\citep{deng2009imagenet}, Food-101~\citep{bossard2014food}, Flowers-102~\citep{nilsback2008flowers}, EuroSAT~\citep{helber2019eurosat}, DTD~\citep{cimpoi2013dtd}, FGVC Aircraft~\citep{maji2013finegrainedaircraft}, MNIST~\citep{lecun1998mnist}, PCAM~\citep{veeling2018pcam}, Country211~\citep{radford2021learning}, CLEVR~\citep{johnson2016clevr}, and SST-2~\citep{socher2013sst}. Collectively, these benchmarks cover coarse object categories, fine-grained recognition, textures, scenes, and language sentiment. We regard each dataset as a separate task and evaluate it using the standard image--text classification protocol.

\subsubsection{Retrieval Datasets}
COCO~\citep{lin2014microsoft} and Flickr30K~\citep{young2014image} provide the retrieval stages. Both contain paired images and captions and are evaluated in the image-to-text and text-to-image directions.

\subsubsection{Task Stream Construction}
The datasets follow one fixed sequential order shared by every method and random seed. At a given stage, non-replay methods optimize only on the current dataset and do not revisit examples from completed tasks. Replay-based baselines additionally access earlier tasks through their explicitly bounded memory buffers, whose capacities are reported with the method configurations. Classification and retrieval are interleaved so that the stream requires the model to alternate between different multimodal objectives.

\subsubsection{Dataset Statistics}
Table~\ref{tab:datasets} reports the train/test sizes, number of categories where applicable, and the evaluation measure for every stage.

\begin{table}[H]
\centering
\scriptsize
\setlength{\tabcolsep}{5pt}
\renewcommand{\arraystretch}{1.03}
\setlength{\abovecaptionskip}{4pt}
\setlength{\belowcaptionskip}{4pt}
\caption{Composition and statistics of the 16-stage continual multimodal benchmark.}
\label{tab:datasets}
\begin{tabular}{l l c c c l}
\toprule
\textbf{Dataset} & \textbf{Task} & \textbf{Classes} & \textbf{Train} & \textbf{Test} & \textbf{Evaluation} \\
\midrule
FGVC Aircraft~\citep{maji2013finegrainedaircraft} & Classification & 100 & 3{,}334   & 3{,}333   & Accuracy \\
Caltech-101~\citep{fei2004learning} & Classification & 102 & 2{,}448   & 6{,}084   & Accuracy \\
ImageNet-1K~\citep{deng2009imagenet} & Classification & 1{,}000 & 1{,}281{,}167 & 50{,}000 & Accuracy \\
CIFAR-10~\citep{krizhevsky2009learning} & Classification & 10  & 45{,}000  & 10{,}000  & Accuracy \\
CIFAR-100~\citep{krizhevsky2009learning} & Classification & 100 & 45{,}000  & 10{,}000  & Accuracy \\
CLEVR~\citep{johnson2016clevr} & Classification & 8   & 4{,}500   & 5{,}000   & Accuracy \\
Country211~\citep{radford2021learning} & Classification & 211 & 31{,}650  & 21{,}100  & Accuracy \\
DTD~\citep{cimpoi2013dtd} & Classification & 47  & 1{,}880   & 1{,}880   & Accuracy \\
EuroSAT~\citep{helber2019eurosat} & Classification & 10  & 5{,}000   & 5{,}000   & Accuracy \\
Flowers-102~\citep{nilsback2008flowers} & Classification & 102 & 1{,}020   & 6{,}149   & Accuracy \\
Food-101~\citep{bossard2014food} & Classification & 101 & 68{,}175  & 25{,}250  & Accuracy \\
MNIST~\citep{lecun1998mnist} & Classification & 10  & 48{,}000  & 10{,}000  & Accuracy \\
PCAM~\citep{veeling2018pcam} & Classification & 2   & 262{,}144 & 32{,}768  & Accuracy \\
SST-2~\citep{socher2013sst} & Classification & 2   & 6{,}920   & 1{,}821   & Accuracy \\
\midrule
COCO~\citep{lin2014microsoft} & Retrieval & -- & 118{,}287 & 5{,}000 & Recall@K \\
Flickr30K~\citep{young2014image} & Retrieval & -- & 29{,}000  & 1{,}000 & Recall@K \\
\bottomrule
\end{tabular}
\end{table}

\appendixsubsection{Task-Order Robustness}
\label{apdx:robustness_drift}

We permute all tasks in the unified 16-task benchmark and keep the HyCoCLIP-B backbone, optimizer, training schedule, and main-table HMCL configurations fixed. The task-order RNG is independent of the five paired training seeds, and no hyperparameter is selected on either shuffled stream. The two orders, generated once using order seeds 2608311 and 2608312, are:

\noindent\textbf{seq2:} Flickr30K $\rightarrow$ MNIST $\rightarrow$ PCAM $\rightarrow$ Food-101 $\rightarrow$ EuroSAT $\rightarrow$ Caltech-101 $\rightarrow$ SST-2 $\rightarrow$ COCO $\rightarrow$ Flowers-102 $\rightarrow$ CIFAR-10 $\rightarrow$ CLEVR $\rightarrow$ Aircraft $\rightarrow$ DTD $\rightarrow$ CIFAR-100 $\rightarrow$ Country211 $\rightarrow$ ImageNet.

\noindent\textbf{seq3:} DTD $\rightarrow$ ImageNet $\rightarrow$ CLEVR $\rightarrow$ CIFAR-10 $\rightarrow$ Flickr30K $\rightarrow$ PCAM $\rightarrow$ Flowers-102 $\rightarrow$ EuroSAT $\rightarrow$ COCO $\rightarrow$ CIFAR-100 $\rightarrow$ Country211 $\rightarrow$ Aircraft $\rightarrow$ MNIST $\rightarrow$ Food-101 $\rightarrow$ Caltech-101 $\rightarrow$ SST-2.

Table~\ref{tab:task_order_stability} reports the direct five-seed results without relative-difference rows. On seq2, HMCL$_{\mathrm{MR}}$ gives the highest Overall score and BWT, while HMCL$_{\mathrm{CA}}$ gives the highest final retrieval score. On seq3, HMCL$_{\mathrm{CA}}$ gives the highest Overall score, while HMCL$_{\mathrm{MR}}$ gives the strongest retrieval score and marginally better Overall BWT. Both complete variants retain a clear advantage over Vanilla under each permutation, showing that the main result is not tied to the canonical task order. The separate AdamW weight-decay sweep is reported in Appendix~\ref{apdx:adamw_weight_decay}.

\begin{table}[H]
\centering
\scriptsize
\setlength{\tabcolsep}{2.4pt}
\renewcommand{\arraystretch}{1.13}
\caption{Task-order robustness on HyCoCLIP-B using two fixed permutations of the unified 16-task stream (mean $\pm$ sample standard deviation over five paired seeds). Classification uses top-1 accuracy, retrieval uses symmetric R@5, and BWT is final minus immediate performance. Both HMCL variants use their fixed main-table configurations; no hyperparameter is retuned for either order.}
\label{tab:task_order_stability}
\begin{adjustbox}{max width=\textwidth}
\begin{tabular}{@{}l l c c c c c c@{}}
\toprule[1.2pt]
& & \multicolumn{2}{c}{\textbf{Classification}}
& \multicolumn{2}{c}{\textbf{Retrieval}}
& \multicolumn{2}{c}{\textbf{Overall}} \\
\cmidrule(lr){3-4}\cmidrule(lr){5-6}\cmidrule(lr){7-8}
\textbf{Order} & \textbf{Method}
& \textbf{Acc} $\uparrow$ & $\mathbf{BWT}_{\mathrm{A}}\uparrow$
& \textbf{R@5} $\uparrow$ & $\mathbf{BWT}_{\mathrm{R}5}\uparrow$
& \textbf{Overall} $\uparrow$ & \textbf{BWT} $\uparrow$ \\
\midrule
seq2 & Vanilla
& $42.095\pm0.128$ & $-3.890\pm0.223$
& $62.673\pm0.095$ & $-10.002\pm0.092$
& $44.668\pm0.104$ & $-4.654\pm0.187$ \\
\rowcolor{myblue!20}
seq2 & HMCL$_{\mathrm{MR}}$
& $\mathbf{45.441\pm0.049}$ & $\mathbf{-0.198\pm0.054}$
& $71.263\pm0.082$ & $\mathbf{-3.997\pm0.101}$
& $\mathbf{48.669\pm0.047}$ & $\mathbf{-0.673\pm0.051}$ \\
\rowcolor{myblue!20}
seq2 & HMCL$_{\mathrm{CA}}$
& $45.222\pm0.020$ & $-0.324\pm0.058$
& $\mathbf{71.351\pm0.071}$ & $-4.139\pm0.057$
& $48.488\pm0.016$ & $-0.801\pm0.054$ \\
\midrule
seq3 & Vanilla
& $41.299\pm0.168$ & $-4.686\pm0.051$
& $62.551\pm0.090$ & $-5.551\pm0.193$
& $43.956\pm0.147$ & $-4.794\pm0.050$ \\
\rowcolor{myblue!20}
seq3 & HMCL$_{\mathrm{MR}}$
& $45.278\pm0.062$ & $-0.733\pm0.088$
& $\mathbf{72.613\pm0.081}$ & $\mathbf{-2.141\pm0.117}$
& $48.695\pm0.053$ & $\mathbf{-0.909\pm0.065}$ \\
\rowcolor{myblue!20}
seq3 & HMCL$_{\mathrm{CA}}$
& $\mathbf{45.430\pm0.070}$ & $\mathbf{-0.684\pm0.124}$
& $72.397\pm0.052$ & $-2.579\pm0.034$
& $\mathbf{48.801\pm0.064}$ & $-0.920\pm0.113$ \\
\bottomrule[1.2pt]
\end{tabular}
\end{adjustbox}
\end{table}

\appendixsubsection{Supplementary Materials for Case Study}
\label{apdx:case_study}

We further visualize how continual adaptation changes the visual--semantic hierarchy learned on Flickr30K. Nouns and adjectives extracted from the dataset captions augment the candidate text pool with concepts at different abstraction levels. Starting from an image embedding, we sample 20 locations on the geodesic leading to [ROOT], the manifold origin, and compare the nearest concepts recovered by HMCL and Vanilla.

Examples (1)--(3) search a joint pool of captions, nouns, and adjectives; examples (4)--(6) use captions alone. Blue rows show traversals immediately after the Flickr30K stage, while green rows show the same analysis after the full continual sequence. In these examples, HMCL gives a clearer movement from image-specific descriptions toward more general concepts. Vanilla shows less consistent results after later tasks, which is consistent with greater semantic and hierarchical drift.

\begin{figure}[!p]
    \footnotesize
    \newcolumntype{Z}{>{\centering\arraybackslash}X}
    \renewcommand{\arraystretch}{1.03}
    \arrayrulecolor{rulegray}
    \setlength{\lightrulewidth}{0.35pt}
    \setlength{\fboxsep}{0pt}
    \setlength{\fboxrule}{0.35pt}
    \setlength{\tabcolsep}{2pt}
    \centering
    {\textcolor{bluedark}{\rule{1.2em}{1.3pt}}\enspace\textbf{After Flickr30K}
    \hspace{2.5em}
    \textcolor{greendark}{\rule{1.2em}{1.3pt}}\enspace\textbf{After the full stream}}\par
    \vspace{4pt}

\begin{minipage}[t]{\linewidth}
    \adjustbox{valign=t}{
        \begin{minipage}[t]{0.18\linewidth}
            \centering
            \fcolorbox{rulegray}{white}{\includegraphics[width=\dimexpr\linewidth-0.7pt\relax]{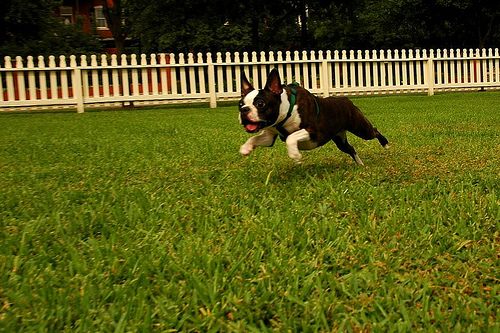}}
            (1)
        \end{minipage}
    }
    \hfill
    \begin{minipage}[t]{0.8\linewidth}
\begin{tabularx}{\linewidth}[t]{ZZ}
            \rowcolor{headergray} \textbf{HMCL (ours)} & \textbf{Vanilla} \\
            \midrule
            \rowcolor{darkblue} \emph{A dog runs on the green grass near a wooden fence .} & \emph{A black and white dog is running in a grassy garden surrounded by a white fence .} \\
            \midrule
            \rowcolor{middleblue} \emph{dog} & \emph{handsome} \\
            \midrule
            \rowcolor{lightblue} \emph{sincerity} & \emph{fondness} \\
            \midrule
            \texttt{\textbf{[ROOT]}} & \texttt{\textbf{[ROOT]}} \\
        \end{tabularx}
\vspace{3pt}
\begin{tabularx}{\linewidth}[t]{ZZ}
            \rowcolor{headergray} \textbf{HMCL (ours)} & \textbf{Vanilla} \\
            \midrule
            \rowcolor{darkgreen} \emph{A dog runs on the green grass near a wooden fence .} & \emph{A black and white dog is running in a grassy garden surrounded by a white fence .} \\
            \midrule
            \rowcolor{middlegreen} \emph{dog} & \emph{aloof} \\
            \midrule
            \rowcolor{lightgreen} \emph{sincerity} & \emph{pleasure} \\
            \midrule
            \texttt{\textbf{[ROOT]}} & \texttt{\textbf{[ROOT]}} \\
        \end{tabularx}
    \end{minipage}
\end{minipage}

\vspace{3pt}   

\begin{minipage}[t]{\linewidth}
    \adjustbox{valign=t}{
        \begin{minipage}[t]{0.18\linewidth}
            \centering
            \fcolorbox{rulegray}{white}{\includegraphics[width=\dimexpr\linewidth-0.7pt\relax]{Figures/case_flickr30k_1007129816.jpg}}
            (2)
        \end{minipage}
    }
    \hfill
    \begin{minipage}[t]{0.8\linewidth}
\begin{tabularx}{\linewidth}[t]{ZZ}
            \rowcolor{headergray} \textbf{HMCL (ours)} & \textbf{Vanilla} \\
            \midrule
            \rowcolor{darkblue} \emph{The man with pierced ears is wearing glasses and an orange hat .} & \emph{The man with pierced ears is wearing glasses and an orange hat .} \\
            \midrule
            \rowcolor{middleblue} \emph{hat} & \emph{fashion} \\
            \midrule
            \rowcolor{lightblue} \emph{fashion} & \emph{style} \\
            \midrule
            \texttt{\textbf{[ROOT]}} & \texttt{\textbf{[ROOT]}} \\
        \end{tabularx}
\vspace{3pt}
\begin{tabularx}{\linewidth}[t]{ZZ}
            \rowcolor{headergray} \textbf{HMCL (ours)} & \textbf{Vanilla} \\
            \midrule
            \rowcolor{darkgreen} \emph{The man with pierced ears is wearing glasses and an orange hat .} & \emph{The man with pierced ears is wearing glasses and an orange hat .} \\
            \midrule
            \rowcolor{middlegreen} \emph{fashion} & \emph{kind} \\
            \midrule
            \rowcolor{lightgreen} \emph{cousin} & \emph{new} \\
            \midrule
            \texttt{\textbf{[ROOT]}} & \texttt{\textbf{[ROOT]}} \\
        \end{tabularx}
    \end{minipage}
    
\end{minipage}

\vspace{3pt}   

\begin{minipage}[t]{\linewidth}
    \adjustbox{valign=t}{
        \begin{minipage}[t]{0.18\linewidth}
            \centering
            \fcolorbox{rulegray}{white}{\includegraphics[width=\dimexpr\linewidth-0.7pt\relax]{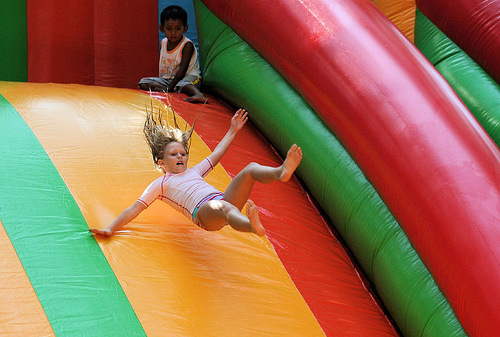}}
            (3)
        \end{minipage}
    }
    \hfill
    \begin{minipage}[t]{0.8\linewidth}
\begin{tabularx}{\linewidth}[t]{ZZ}
            \rowcolor{headergray} \textbf{HMCL (ours)} & \textbf{Vanilla} \\
            \midrule
            \rowcolor{darkblue} \emph{A girl in a pink shirt slides down an inflatable fun slide .} & \emph{Young girl sliding down an inflated slide .} \\
            \midrule
            \rowcolor{middleblue} \emph{summer} & \emph{summer} \\
            \midrule
            \rowcolor{lightblue} \emph{conventional} & \emph{quality} \\
            \midrule
            \texttt{\textbf{[ROOT]}} & \texttt{\textbf{[ROOT]}} \\
        \end{tabularx}
\vspace{3pt}
\begin{tabularx}{\linewidth}[t]{ZZ}
            \rowcolor{headergray} \textbf{HMCL (ours)} & \textbf{Vanilla} \\
            \midrule
            \rowcolor{darkgreen} \emph{A girl in a pink shirt slides down an inflatable fun slide .} & \emph{Young girl sliding down an inflated slide .} \\
            \midrule
            \rowcolor{middlegreen} \emph{summer} & \emph{summer} \\
            \midrule
            \rowcolor{lightgreen} \emph{quick} & \emph{material} \\
            \midrule
            \texttt{\textbf{[ROOT]}} & \texttt{\textbf{[ROOT]}} \\
        \end{tabularx}
    \end{minipage}
\end{minipage}

\end{figure}

\begin{figure}[!p]
    \footnotesize
    \newcolumntype{Z}{>{\centering\arraybackslash}X}
    \renewcommand{\arraystretch}{1.03}
    \arrayrulecolor{rulegray}
    \setlength{\lightrulewidth}{0.35pt}
    \setlength{\fboxsep}{0pt}
    \setlength{\fboxrule}{0.35pt}
    \setlength{\tabcolsep}{2pt}
    \centering
    {\textcolor{bluedark}{\rule{1.2em}{1.3pt}}\enspace\textbf{After Flickr30K}
    \hspace{2.5em}
    \textcolor{greendark}{\rule{1.2em}{1.3pt}}\enspace\textbf{After the full stream}}\par
    \vspace{4pt}
\begin{minipage}[t]{\linewidth}
    \adjustbox{valign=t}{
        \begin{minipage}[t]{0.18\linewidth}
            \centering
            \fcolorbox{rulegray}{white}{\includegraphics[width=\dimexpr\linewidth-0.7pt\relax]{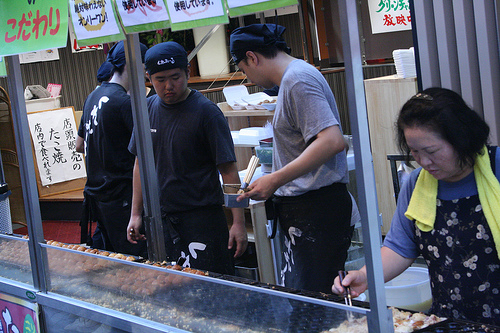}}
            (4)
        \end{minipage}
    }
    \hfill
    \begin{minipage}[t]{0.8\linewidth}
\begin{tabularx}{\linewidth}[t]{ZZ}
            \rowcolor{headergray} \textbf{HMCL (ours)} & \textbf{Vanilla} \\
            \midrule
            \rowcolor{darkblue} \emph{Numerous Asian people working behind a counter .} & \emph{Numerous Asian people working behind a counter .} \\
            \midrule
            \rowcolor{middleblue} \emph{An Asian food market where a woman is picking out food .} & \emph{An Asian food market where a woman is picking out food .} \\
            \midrule
            \rowcolor{lightblue} \emph{Employees at a sushi restaurant prepare for dinner time rush .} & \emph{A cyclist wearing a black helmet is riding by some black vans .} \\
            \midrule
            \texttt{\textbf{[ROOT]}} & \texttt{\textbf{[ROOT]}} \\
        \end{tabularx}
\vspace{3pt}
\begin{tabularx}{\linewidth}[t]{ZZ}
            \rowcolor{headergray} \textbf{HMCL (ours)} & \textbf{Vanilla} \\
            \midrule
            \rowcolor{darkgreen} \emph{Numerous Asian people working behind a counter .} & \emph{Employees at a sushi restaurant prepare for dinner time rush .} \\
            \midrule
            \rowcolor{middlegreen} \emph{An Asian food market where a woman is picking out food .} & \emph{A cyclist wearing a black helmet is riding by some black vans .} \\
            \midrule
            \rowcolor{lightgreen} \emph{A cyclist wearing a black helmet is riding by some black vans .} & \emph{Two wet dogs run into the surf at sunset .} \\
            \midrule
            \texttt{\textbf{[ROOT]}} & \texttt{\textbf{[ROOT]}} \\
        \end{tabularx}
    \end{minipage}
\end{minipage}

\vspace{3pt}   

\begin{minipage}[t]{\linewidth}
    \adjustbox{valign=t}{
        \begin{minipage}[t]{0.18\linewidth}
            \centering
            \fcolorbox{rulegray}{white}{\includegraphics[width=\dimexpr\linewidth-0.7pt\relax]{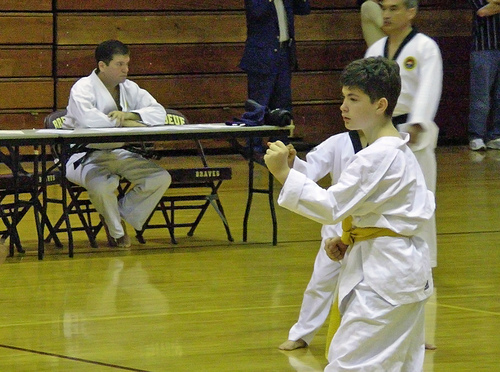}}
            (5)
        \end{minipage}
    }
    \hfill
    \begin{minipage}[t]{0.8\linewidth}
\begin{tabularx}{\linewidth}[t]{ZZ}
            \rowcolor{headergray} \textbf{HMCL (ours)} & \textbf{Vanilla} \\
            \midrule
            \rowcolor{darkblue} \emph{2 boys in the foreground in a karate competition and coaches in background looking on with another coach sitting at table .} & \emph{A young boy demonstrates karate in a gymnasium .} \\
            \midrule
            \rowcolor{middleblue} \emph{A young boy demonstrates karate in a gymnasium .} & \emph{A little girl dressed in yellow splashes in a shallow pool .} \\
            \midrule
            \rowcolor{lightblue} \emph{Man standing by a poster of religious beliefs .} & \emph{A child plays at a playground .} \\
            \midrule
            \texttt{\textbf{[ROOT]}} & \texttt{\textbf{[ROOT]}} \\
        \end{tabularx}
\vspace{3pt}
\begin{tabularx}{\linewidth}[t]{ZZ}
            \rowcolor{headergray} \textbf{HMCL (ours)} & \textbf{Vanilla} \\
            \midrule
            \rowcolor{darkgreen} \emph{2 boys in the foreground in a karate competition and coaches in background looking on with another coach sitting at table .} & \emph{Two people are sitting outdoors on a blanket near a tree .} \\
            \midrule
            \rowcolor{middlegreen} \emph{A young boy demonstrates karate in a gymnasium .} & \emph{A surfer jumps a wave .} \\
            \midrule
            \rowcolor{lightgreen} \emph{Man standing by a poster of religious beliefs .} & \emph{A couple enjoying a glass of white wine .} \\
            \midrule
            \texttt{\textbf{[ROOT]}} & \texttt{\textbf{[ROOT]}} \\
        \end{tabularx}
    \end{minipage}
\end{minipage}

\vspace{3pt}   

\begin{minipage}[t]{\linewidth}
    \adjustbox{valign=t}{
        \begin{minipage}[t]{0.18\linewidth}
            \centering
            \fcolorbox{rulegray}{white}{\includegraphics[width=\dimexpr\linewidth-0.7pt\relax]{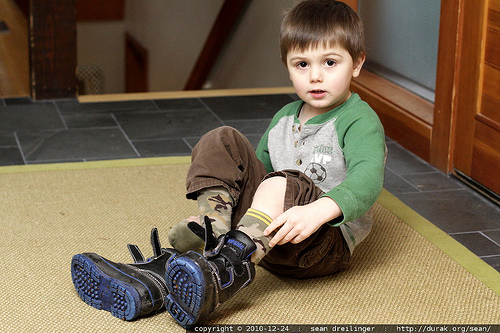}}
            (6)
        \end{minipage}
    }
    \hfill
    \begin{minipage}[t]{0.8\linewidth}
\begin{tabularx}{\linewidth}[t]{ZZ}
            \rowcolor{headergray} \textbf{HMCL (ours)} & \textbf{Vanilla} \\
            \midrule
            \rowcolor{darkblue} \emph{A young boy is sitting on the floor trying to take off his boots .} & \emph{A woman wearing glasses and an orange dress is standing with another woman wearing glasses and a green dress .} \\
            \midrule
            \rowcolor{middleblue} \emph{A young boy puts on his boots at the top of a flight of stairs .} & \emph{Three young men wearing hooded sweatshirts , standing at a bench .} \\
            \midrule
            \rowcolor{lightblue} \emph{A young boy demonstrates karate in a gymnasium .} & \emph{A couple enjoying a glass of white wine .} \\
            \midrule
            \texttt{\textbf{[ROOT]}} & \texttt{\textbf{[ROOT]}} \\
        \end{tabularx}
\vspace{3pt}
\begin{tabularx}{\linewidth}[t]{ZZ}
            \rowcolor{headergray} \textbf{HMCL (ours)} & \textbf{Vanilla} \\
            \midrule
            \rowcolor{darkgreen} \emph{A young boy is sitting on the floor trying to take off his boots .} & \emph{An Indian girl playing a guitar .} \\
            \midrule
            \rowcolor{middlegreen} \emph{A young boy puts on his boots at the top of a flight of stairs .} & \emph{A couple enjoying a glass of white wine .} \\
            \midrule
            \rowcolor{lightgreen} \emph{ couple enjoying a glass of white wine .} & \emph{Two wet dogs run into the surf at sunset .} \\
            \midrule
            \texttt{\textbf{[ROOT]}} & \texttt{\textbf{[ROOT]}} \\
        \end{tabularx}
    \end{minipage}
\end{minipage}

\end{figure}

\begin{journalnew}
\appendixsection{Extended Experimental Protocol and Results}
\label{apdx:extended_experiments}

\appendixsubsection{ImageNet--WordNet Protocol and Per-Task Results}
\label{apdx:imagenet_wordnet_protocol}

We use the unified 16-task stream and metric convention defined in the main paper; Appendix~\ref{apdx:datasets} gives the complete task order and dataset statistics. ImageNet-1K~\citep{deng2009imagenet} uses its official training split and 50,000-image validation split, and each class is associated with a WordNet synset~\citep{miller1995wordnet}. The hierarchy evaluation additionally reports TIE, LCA, ancestor Jaccard, hierarchical precision, and hierarchical recall.

The per-task and pullback analyses below use HyCoCLIP-B. Its cumulative protected basis follows the 80\% explained-variance rule, with retained rank after each completed task
\[
2,12,29,29,29,29,32,32,32,32,32,33,33,33,29.
\]

For both HMCL variants, we select one fixed configuration on seed 42 using the Overall--BWT trade-off and reuse it for the remaining reported seeds. The pullback analysis below varies only its coefficient after fixing the correction family and all other training choices.

\begin{table}[H]
\color{journalblue}
\centering
\scriptsize
\setlength{\tabcolsep}{3.3pt}
\renewcommand{\arraystretch}{1.10}
\caption{Final performance for every task in the extended stream (five-seed mean $\pm$ sample standard deviation). COCO and Flickr30K use symmetric R@5; all other tasks use top-1 accuracy. The two HMCL rows use the complete post-AdamW task-pullback configurations from Table~\ref{tab:main_results16}; shaded cells denote HMCL, and boldface marks the best result for each task.}
\label{tab:supp_extended_per_task}
\begin{adjustbox}{max width=\textwidth}
\begin{tabular}{l c c c c c c c c }
\toprule
\textbf{Method} & \textbf{Aircraft} & \textbf{Caltech} & \textbf{ImageNet} & \textbf{C10} & \textbf{C100} & \textbf{CLEVR} & \textbf{COCO} & \textbf{Country211} \\
\midrule
Vanilla & $4.94\pm0.10$ & $74.26\pm0.15$ & $39.40\pm0.06$ & $87.80\pm0.12$ & $48.52\pm0.33$ & $14.33\pm0.22$ & $48.84\pm0.20$ & $4.48\pm0.10$ \\
EWC & $5.02\pm0.12$ & $74.63\pm0.08$ & $40.04\pm0.06$ & $88.17\pm0.12$ & $49.47\pm0.29$ & $14.40\pm0.20$ & $50.17\pm0.18$ & $4.56\pm0.09$ \\
GEM & $4.99\pm0.19$ & $75.09\pm0.26$ & $39.25\pm0.12$ & $88.71\pm0.31$ & $50.60\pm0.46$ & $15.00\pm0.98$ & $49.36\pm0.21$ & $4.62\pm0.09$ \\
C-FLAT & $4.83\pm0.12$ & $74.03\pm0.22$ & $38.93\pm0.01$ & $87.70\pm0.13$ & $48.87\pm0.19$ & $16.80\pm0.38$ & $49.08\pm0.14$ & $4.60\pm0.06$ \\
\cellcolor{myblue!20}HMCL$_{\mathrm{MR}}$ & \cellcolor{myblue!20}$5.77\pm0.11$ & \cellcolor{myblue!20}$76.63\pm0.14$ & \cellcolor{myblue!20}$\mathbf{45.70\pm0.03}$ & \cellcolor{myblue!20}$89.52\pm0.07$ & \cellcolor{myblue!20}$56.31\pm0.11$ & \cellcolor{myblue!20}$24.49\pm0.89$ & \cellcolor{myblue!20}$\mathbf{60.12\pm0.09}$ & \cellcolor{myblue!20}$5.92\pm0.01$ \\
\cellcolor{myblue!20}HMCL$_{\mathrm{CA}}$ & \cellcolor{myblue!20}$\mathbf{5.80\pm0.08}$ & \cellcolor{myblue!20}$\mathbf{76.91\pm0.11}$ & \cellcolor{myblue!20}$45.59\pm0.05$ & \cellcolor{myblue!20}$\mathbf{89.57\pm0.08}$ & \cellcolor{myblue!20}$\mathbf{56.77\pm0.20}$ & \cellcolor{myblue!20}$\mathbf{25.86\pm0.50}$ & \cellcolor{myblue!20}$60.10\pm0.09$ & \cellcolor{myblue!20}$\mathbf{6.09\pm0.04}$ \\
\bottomrule
\end{tabular}
\end{adjustbox}
\vspace{4pt}

\begin{adjustbox}{max width=\textwidth}
\begin{tabular}{l c c c c c c c c }
\toprule
\textbf{Method} & \textbf{DTD} & \textbf{EuroSAT} & \textbf{Flickr30K} & \textbf{Food101} & \textbf{MNIST} & \textbf{Flowers} & \textbf{PCAM} & \textbf{SST-2} \\
\midrule
Vanilla & $24.49\pm0.17$ & $41.00\pm0.65$ & $75.23\pm0.19$ & $56.78\pm0.20$ & $28.47\pm1.23$ & $31.68\pm0.33$ & $62.97\pm0.17$ & $54.70\pm0.35$ \\
EWC & $24.57\pm0.30$ & $40.64\pm0.87$ & $76.57\pm0.15$ & $56.99\pm0.14$ & $28.45\pm0.93$ & $31.92\pm0.37$ & $61.92\pm0.16$ & $54.90\pm0.26$ \\
GEM & $25.96\pm0.44$ & $47.99\pm1.56$ & $75.81\pm0.19$ & $56.85\pm0.60$ & $\mathbf{30.04\pm1.46}$ & $31.87\pm0.65$ & $63.58\pm0.91$ & $55.00\pm0.23$ \\
C-FLAT & $24.35\pm0.39$ & $40.57\pm0.78$ & $74.94\pm0.17$ & $56.34\pm0.16$ & $27.47\pm1.14$ & $31.13\pm0.23$ & $\mathbf{63.79\pm0.22}$ & $53.89\pm0.33$ \\
\cellcolor{myblue!20}HMCL$_{\mathrm{MR}}$ & \cellcolor{myblue!20}$27.71\pm0.28$ & \cellcolor{myblue!20}$55.87\pm0.50$ & \cellcolor{myblue!20}$\mathbf{85.48\pm0.11}$ & \cellcolor{myblue!20}$61.10\pm0.11$ & \cellcolor{myblue!20}$29.52\pm0.47$ & \cellcolor{myblue!20}$\mathbf{32.74\pm0.15}$ & \cellcolor{myblue!20}$60.80\pm0.08$ & \cellcolor{myblue!20}$\mathbf{55.52\pm0.08}$ \\
\cellcolor{myblue!20}HMCL$_{\mathrm{CA}}$ & \cellcolor{myblue!20}$\mathbf{27.89\pm0.19}$ & \cellcolor{myblue!20}$\mathbf{56.44\pm0.33}$ & \cellcolor{myblue!20}$85.45\pm0.13$ & \cellcolor{myblue!20}$\mathbf{61.46\pm0.06}$ & \cellcolor{myblue!20}$\mathbf{30.49\pm0.54}$ & \cellcolor{myblue!20}$32.48\pm0.15$ & \cellcolor{myblue!20}$60.79\pm0.09$ & \cellcolor{myblue!20}$55.35\pm0.13$ \\
\bottomrule
\end{tabular}
\end{adjustbox}
\end{table}

The two HMCL variants attain the highest final score on 15 of the 16 tasks: $\mathrm{HMCL}_{\mathrm{CA}}$ leads on ten classification tasks, while $\mathrm{HMCL}_{\mathrm{MR}}$ leads on ImageNet-1K, both retrieval datasets, Flowers102, and SST-2. C-FLAT remains strongest on PCAM. This per-task breakdown agrees with the aggregate comparison: after matching both the optimizer-step budget and the global replay capacity, HMCL's gain is distributed across classification and retrieval rather than being driven by a small subset of tasks. The complete per-run matrices, aggregation outputs, and source provenance accompany the extended experiment artifact.

\appendixsubsection{Modality-Specific Radial Contraction}
\label{apdx:norm_contraction}

As a modality-specific complement to Figure~\ref{fig:imagenet_wordnet_radial_geometry} in the main paper, we examine MERU-L at seed 1024 under the fixed 16-task protocol. We include every dataset with a post-task forgetting interval: all 15 old tasks in the sequence, excluding only the final SST-2 task because no later learning follows it. For each old task, we sample 500 fixed test anchors and compare their spatial norms immediately after learning that task with those at the final checkpoint after the complete sequence. This gives 7,500 image and 7,500 matched text anchors per method. Figure~\ref{fig:norm_distribution} pools these anchors across all 15 old tasks; the dashed lines mark the corresponding pooled task-end medians.

\begin{figure}[!tbp]
    \centering
    \includegraphics[width=0.91\linewidth]{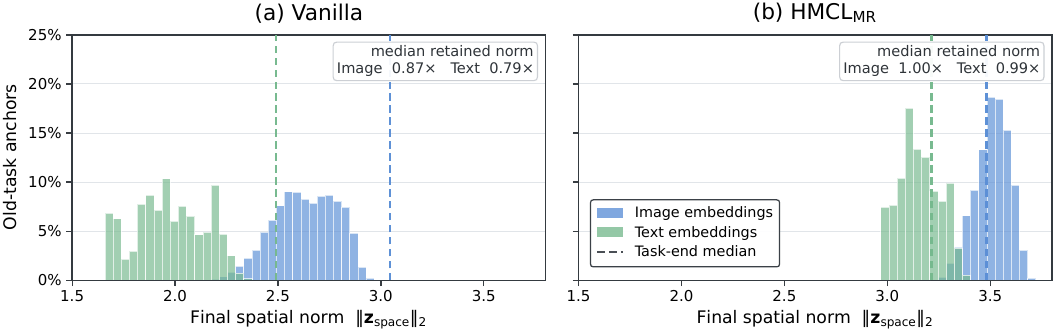}
    \caption{Modality-specific spatial-norm distributions after the complete MERU-L continual stream (seed 1024). The distributions pool 500 fixed test anchors from \revisiontext{each of the 15 old tasks} in the 16-task sequence; SST-2 is excluded only because it is last and has no post-task forgetting interval. Bars show final-checkpoint norms of old-task image and text embeddings, and color-matched dashed lines show their pooled task-end medians. The inset values report the median per-anchor ratio $\lVert\mathbf z_{\mathrm{final,space}}\rVert_2/\lVert\mathbf z_{\mathrm{task\text{-}end,space}}\rVert_2$.}
    \label{fig:norm_distribution}
\end{figure}

Vanilla retains only $0.87\times$ of the image norm and $0.79\times$ of the text norm at the median. The contraction appears in the task-level medians for 14 of 15 image tasks and all 15 text tasks. In contrast, $\mathrm{HMCL}_{\mathrm{MR}}$ retains $1.00\times$ and $0.99\times$, respectively, and no task-level median falls below $0.9\times$ for either modality. Thus, the radial component of forgetting is a systematic contraction, especially for text, rather than merely nondirectional displacement. We avoid calling it a collapse to the origin because the Vanilla distributions remain nondegenerate.

\begin{revisionnew}
\noindent\textbf{Scope of the guarantee.}
The after-AdamW MR correction exactly preserves only the retained subspace represented by $\mathbf V_{t-1}$. Because PCA controls average residual energy rather than the worst individual task, small drift can remain outside that subspace; the correction also leaves AdamW's moment states unchanged. These limits explain why the empirical hierarchy drift is strongly reduced rather than identically zero.
\end{revisionnew}

\begin{revisionnew}
\appendixsubsection{Stage-Wise Stability--Plasticity Diagnostics}
\label{apdx:stagewise_stability_plasticity}

\begin{figure}[H]
\centering
\includegraphics[width=0.94\linewidth]{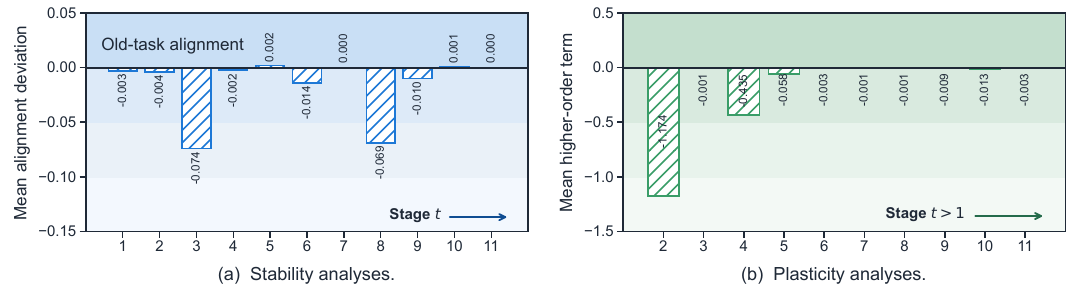}
\caption{Stage-wise stability and plasticity diagnostics replotted from the original CMCL analysis of Liu \emph{et al.}~\cite{liu2025continual}. At stage $t$, panel (a) reports the mean deviation of previously learned modality-pair alignment scores from their preceding reference values; deviations close to zero indicate stable old-pair alignment. For $t>1$, panel (b) reports the mean higher-order term used to assess the source plasticity condition $o(\eta)/\eta\leq0$; nonpositive values satisfy that condition. These theorem-specific quantities are not BWT or task-score gains.}
\label{fig:stability_plasticity_sequence}
\end{figure}

The old-pair alignment deviations remain between $-0.074$ and $0.002$ and approach zero at several stages. The higher-order terms are nonpositive throughout; after the two largest early-stage terms, their magnitudes fall below $0.06$. Because this diagnostic uses the original CMCL protocol, it is included here as a theorem-specific consistency check rather than as a result on the current 16-task HMCL benchmark.
\end{revisionnew}

\appendixsubsection{Fine-Grained Retrieval on the Modality-Extended Stream}
\label{apdx:finegrained_retrieval17}

Figure~\ref{fig:retrieval_finegrained17} in the main paper summarizes final COCO and Flickr30K retrieval in the separate modality-extension protocol, which adds one audio and one thermal retrieval task to the original continual benchmark. Table~\ref{tab:retrieval_finegrained} provides the corresponding text-to-image and image-to-text Recall@$\{1,5,10\}$ means and standard deviations for all three backbones and methods. This protocol is reported separately from the unified 16-task ImageNet experiment.

\begin{table}[t]
\color{journalblue}
\centering
\scriptsize
\setlength{\tabcolsep}{3.2pt}
\renewcommand{\arraystretch}{1.04}
\caption{Fine-grained final retrieval performance in the separate protocol augmented with audio and thermal tasks (mean $\pm$ standard deviation over three paired seeds). T2I and I2T denote text-to-image and image-to-text retrieval, respectively. The methods and configurations match Table~\ref{tab:backbone_robustness17}. Boldface marks the best result for each backbone and metric; shaded cells denote HMCL.}
\label{tab:retrieval_finegrained}

\begin{tabular}{@{}llcccccc@{}}
\toprule
\multicolumn{8}{c}{\textbf{(a) COCO}} \\
\cmidrule(lr){1-8}
\textbf{Backbone} & \textbf{Method} & \multicolumn{3}{c}{\textbf{T2I}} & \multicolumn{3}{c}{\textbf{I2T}} \\
\cmidrule(lr){3-5}\cmidrule(lr){6-8}
 & & \textbf{R@1} & \textbf{R@5} & \textbf{R@10} & \textbf{R@1} & \textbf{R@5} & \textbf{R@10} \\
\midrule
\multirow{3}{*}{MERU-L}
 & Vanilla & $9.303\pm0.072$ & $23.844\pm0.065$ & $32.835\pm0.108$ & $15.113\pm0.114$ & $33.633\pm0.220$ & $43.773\pm0.247$ \\
 & \cellcolor{myblue!20}HMCL$_{\rm MR}$ & \cellcolor{myblue!20}$8.848\pm0.067$ & \cellcolor{myblue!20}$22.787\pm0.183$ & \cellcolor{myblue!20}$31.955\pm0.245$ & \cellcolor{myblue!20}$15.847\pm0.121$ & \cellcolor{myblue!20}$34.573\pm0.549$ & \cellcolor{myblue!20}$45.353\pm0.186$ \\
 & \cellcolor{myblue!20}HMCL$_{\rm CA}$ & \cellcolor{myblue!20}$\mathbf{10.725\pm0.120}$ & \cellcolor{myblue!20}$\mathbf{26.557\pm0.143}$ & \cellcolor{myblue!20}$\mathbf{36.344\pm0.032}$ & \cellcolor{myblue!20}$\mathbf{18.040\pm0.250}$ & \cellcolor{myblue!20}$\mathbf{37.293\pm0.239}$ & \cellcolor{myblue!20}$\mathbf{47.913\pm0.348}$ \\
\midrule
\multirow{3}{*}{MERU-B}
 & Vanilla & $9.491\pm0.038$ & $23.635\pm0.113$ & $32.550\pm0.195$ & $13.067\pm0.522$ & $29.293\pm0.440$ & $38.740\pm0.596$ \\
 & \cellcolor{myblue!20}HMCL$_{\rm MR}$ & \cellcolor{myblue!20}$10.782\pm0.045$ & \cellcolor{myblue!20}$26.439\pm0.048$ & \cellcolor{myblue!20}$36.166\pm0.137$ & \cellcolor{myblue!20}$15.413\pm0.140$ & \cellcolor{myblue!20}$34.113\pm0.090$ & \cellcolor{myblue!20}$44.540\pm0.159$ \\
 & \cellcolor{myblue!20}HMCL$_{\rm CA}$ & \cellcolor{myblue!20}$\mathbf{12.071\pm0.067}$ & \cellcolor{myblue!20}$\mathbf{28.605\pm0.068}$ & \cellcolor{myblue!20}$\mathbf{38.408\pm0.067}$ & \cellcolor{myblue!20}$\mathbf{17.360\pm0.216}$ & \cellcolor{myblue!20}$\mathbf{36.973\pm0.046}$ & \cellcolor{myblue!20}$\mathbf{47.780\pm0.203}$ \\
\midrule
\multirow{3}{*}{HyCoCLIP-B}
 & Vanilla & $21.074\pm0.212$ & $42.901\pm0.334$ & $54.318\pm0.102$ & $34.900\pm0.180$ & $60.640\pm0.322$ & $71.847\pm0.153$ \\
 & \cellcolor{myblue!20}HMCL$_{\rm MR}$ & \cellcolor{myblue!20}$20.219\pm0.063$ & \cellcolor{myblue!20}$42.571\pm0.264$ & \cellcolor{myblue!20}$54.019\pm0.277$ & \cellcolor{myblue!20}$36.927\pm0.390$ & \cellcolor{myblue!20}$62.660\pm0.278$ & \cellcolor{myblue!20}$73.660\pm0.223$ \\
 & \cellcolor{myblue!20}HMCL$_{\rm CA}$ & \cellcolor{myblue!20}$\mathbf{23.843\pm0.138}$ & \cellcolor{myblue!20}$\mathbf{47.056\pm0.164}$ & \cellcolor{myblue!20}$\mathbf{58.643\pm0.077}$ & \cellcolor{myblue!20}$\mathbf{40.767\pm0.250}$ & \cellcolor{myblue!20}$\mathbf{66.833\pm0.163}$ & \cellcolor{myblue!20}$\mathbf{76.827\pm0.129}$ \\
\bottomrule
\end{tabular}

\vspace{4pt}

\begin{tabular}{@{}llcccccc@{}}
\toprule
\multicolumn{8}{c}{\textbf{(b) Flickr30K}} \\
\cmidrule(lr){1-8}
\textbf{Backbone} & \textbf{Method} & \multicolumn{3}{c}{\textbf{T2I}} & \multicolumn{3}{c}{\textbf{I2T}} \\
\cmidrule(lr){3-5}\cmidrule(lr){6-8}
 & & \textbf{R@1} & \textbf{R@5} & \textbf{R@10} & \textbf{R@1} & \textbf{R@5} & \textbf{R@10} \\
\midrule
\multirow{3}{*}{MERU-L}
 & Vanilla & $12.713\pm0.046$ & $29.227\pm0.266$ & $38.187\pm0.280$ & $19.500\pm0.265$ & $39.100\pm0.173$ & $49.933\pm0.404$ \\
 & \cellcolor{myblue!20}HMCL$_{\rm MR}$ & \cellcolor{myblue!20}$11.147\pm0.130$ & \cellcolor{myblue!20}$27.200\pm0.333$ & \cellcolor{myblue!20}$36.073\pm0.489$ & \cellcolor{myblue!20}$21.667\pm0.306$ & \cellcolor{myblue!20}$41.067\pm0.058$ & \cellcolor{myblue!20}$50.933\pm0.252$ \\
 & \cellcolor{myblue!20}HMCL$_{\rm CA}$ & \cellcolor{myblue!20}$\mathbf{14.600\pm0.160}$ & \cellcolor{myblue!20}$\mathbf{33.040\pm0.365}$ & \cellcolor{myblue!20}$\mathbf{43.147\pm0.081}$ & \cellcolor{myblue!20}$\mathbf{22.633\pm0.404}$ & \cellcolor{myblue!20}$\mathbf{44.500\pm0.100}$ & \cellcolor{myblue!20}$\mathbf{54.833\pm0.153}$ \\
\midrule
\multirow{3}{*}{MERU-B}
 & Vanilla & $11.813\pm0.061$ & $27.773\pm0.192$ & $36.613\pm0.428$ & $15.400\pm0.265$ & $34.633\pm0.751$ & $43.267\pm0.321$ \\
 & \cellcolor{myblue!20}HMCL$_{\rm MR}$ & \cellcolor{myblue!20}$13.967\pm0.142$ & \cellcolor{myblue!20}$31.780\pm0.140$ & \cellcolor{myblue!20}$41.300\pm0.300$ & \cellcolor{myblue!20}$18.733\pm0.493$ & \cellcolor{myblue!20}$40.367\pm0.503$ & \cellcolor{myblue!20}$50.767\pm0.603$ \\
 & \cellcolor{myblue!20}HMCL$_{\rm CA}$ & \cellcolor{myblue!20}$\mathbf{16.673\pm0.227}$ & \cellcolor{myblue!20}$\mathbf{35.000\pm0.087}$ & \cellcolor{myblue!20}$\mathbf{44.833\pm0.151}$ & \cellcolor{myblue!20}$\mathbf{20.367\pm0.058}$ & \cellcolor{myblue!20}$\mathbf{43.367\pm0.569}$ & \cellcolor{myblue!20}$\mathbf{54.567\pm0.252}$ \\
\midrule
\multirow{3}{*}{HyCoCLIP-B}
 & Vanilla & $45.073\pm0.391$ & $70.940\pm0.308$ & $79.693\pm0.061$ & $64.433\pm0.802$ & $86.100\pm0.173$ & $92.200\pm0.100$ \\
 & \cellcolor{myblue!20}HMCL$_{\rm MR}$ & \cellcolor{myblue!20}$45.860\pm0.333$ & \cellcolor{myblue!20}$72.207\pm0.469$ & \cellcolor{myblue!20}$81.147\pm0.225$ & \cellcolor{myblue!20}$66.433\pm0.814$ & \cellcolor{myblue!20}$87.100\pm0.265$ & \cellcolor{myblue!20}$92.900\pm0.265$ \\
 & \cellcolor{myblue!20}HMCL$_{\rm CA}$ & \cellcolor{myblue!20}$\mathbf{49.613\pm0.147}$ & \cellcolor{myblue!20}$\mathbf{76.160\pm0.223}$ & \cellcolor{myblue!20}$\mathbf{84.240\pm0.140}$ & \cellcolor{myblue!20}$\mathbf{69.800\pm0.346}$ & \cellcolor{myblue!20}$\mathbf{89.167\pm0.351}$ & \cellcolor{myblue!20}$\mathbf{94.000\pm0.200}$ \\
\bottomrule
\end{tabular}
\end{table}

\appendixsubsection{Reproducibility Notes}
\label{apdx:reproducibility_notes}

Before continual training, we validated the complete cached feature store against the expected sample counts, feature and label shapes, and label ranges.

{\color{black}
The canonical task order is Aircraft, Caltech-101, ImageNet-1K, CIFAR-10, CIFAR-100, CLEVR, COCO, Country211, DTD, EuroSAT, Flickr30K, Food-101, MNIST, Flowers-102, PCAM, and SST-2. The paired seeds are 42, 123, 456, 789, and 1024. All main results use cached 512-dimensional backbone features. ImageNet stores 1,281,167 image features and one seven-prompt text prototype per class; class prototypes are paired to image features by label inside each minibatch.

For classification tasks, AdamW uses learning rate $5\times10^{-4}$, zero weight decay, 10 epochs, batch size 1024, and entailment weight 0.2. Retrieval tasks use learning rate $5\times10^{-5}$, zero weight decay, 15 epochs, batch size 1024, and the same entailment weight. ``One optimizer step per epoch'' means that gradients of all minibatch losses in that epoch are accumulated before a single AdamW update; gradients are cleared only at the epoch boundary. EWC accumulates its regularized loss in the same way. GEM likewise accumulates the current and memory gradients, applies its projection once when the epoch gradient violates a constraint, and then performs one AdamW step. Thus every method receives the same epoch-level parameter-update budget.

Let $\rho$ denote retained PCA variance. The fixed HMCL settings are
\[
\begin{array}{c|ccc|ccc}
&\multicolumn{3}{c|}{\mathrm{MR}}&\multicolumn{3}{c}{\mathrm{CA}}\\
\text{backbone}&\rho&\beta&k_{\min}&\rho&\beta&k_{\min}\\\hline
\text{MERU-L}&.60&.14&8&.65&.13&8\\
\text{MERU-B}&.70&.075&8&.70&.075/.20&11\\
\text{HyCoCLIP-B}&.80&.125&8&.75&.15&1
\end{array}
\]
where the MERU-B CA entry .075/.20 gives classification/retrieval pullback coefficients. These configurations are fixed after selection on seed 42 and then reused without change for the other four seeds.

EWC uses coefficient $10^{-4}$ and memory size 512. GEM uses margin 0.5, the exact \texttt{quadprog} solver, and a task-balanced global buffer capped at 512 feature pairs; after all 16 tasks it contains 32 pairs per task. C-FLAT uses Co2L contrastive distillation with power 1.0 and memory size 512, perturbation radius $\rho=0.2$, and first-order coefficient $\lambda=0.2$; AdamW is its wrapped base optimizer, with base AdamW on task 1 and C-FLAT activated from task 2. DNS uses its dual-sided Euclidean null-space projection under the same task order, cached inputs, schedule, and update budget. Each run writes a task-end checkpoint and partial performance matrix, and is accepted only when all 16 tasks are complete and the matrix is a finite $16\times16$ lower-triangular array. The experiment artifact retains the per-run configurations, matrices, checkpoints, source hashes, exclusion audit, and aggregation outputs.
}

\appendixsubsection{Robustness to AdamW Weight Decay}
\label{apdx:adamw_weight_decay}

The main-table experiments use zero weight decay. To test whether the conclusions depend on this choice, we sweep $\lambda\in\{0,10^{-4},10^{-3},10^{-2}\}$ on HyCoCLIP-B while holding all other training conditions fixed. The $\mathrm{MR}_{\mathrm{pre}}$ control applies the minimal-rotation correction to the raw gradient before AdamW and uses no pullback. We use the distinct notation $\mathrm{MR}_{\mathrm{pre}}$ because this control is not the complete post-AdamW, pullback-equipped $\mathrm{HMCL}_{\mathrm{MR}}$ reported in the main comparison. $\mathrm{HMCL}_{\mathrm{CA}}$ uses the fixed complete configuration from the main comparison.

\begin{table}[H]
\color{journalblue}
\centering
\scriptsize
\setlength{\tabcolsep}{4.5pt}
\renewcommand{\arraystretch}{1.14}
\caption{Robustness to AdamW weight decay on HyCoCLIP-B (mean $\pm$ sample standard deviation over five paired seeds). Overall averages the final scores of 14 classification and two retrieval tasks; signed BWT is final minus immediate performance, so higher is better. All conditions share the task order, learning-rate schedule, epochs, batch size, loss, and update budget; only the weight-decay coefficient $\lambda$ changes. $\mathrm{MR}_{\mathrm{pre}}$ is the raw-gradient, pre-AdamW minimal-rotation control, whereas $\mathrm{HMCL}_{\mathrm{CA}}$ uses the complete post-AdamW correction.}
\label{tab:adamw_weight_decay}
\begin{adjustbox}{max width=\linewidth}
\begin{tabular}{@{}l c c c c c c@{}}
\toprule
\multirow{2}{*}{\textbf{$\lambda$}}
& \multicolumn{3}{c}{\textbf{Overall $\uparrow$}}
& \multicolumn{3}{c}{\textbf{BWT $\uparrow$}} \\
\cmidrule(lr){2-4}\cmidrule(lr){5-7}
& \textbf{Vanilla} & \cellcolor{myblue!20}\textbf{$\mathrm{MR}_{\mathrm{pre}}$} & \cellcolor{myblue!20}\textbf{$\mathrm{HMCL}_{\mathrm{CA}}$}
& \textbf{Vanilla} & \cellcolor{myblue!20}\textbf{$\mathrm{MR}_{\mathrm{pre}}$} & \cellcolor{myblue!20}\textbf{$\mathrm{HMCL}_{\mathrm{CA}}$} \\
\midrule
$0$
& $43.618\pm0.152$ & \cellcolor{myblue!20}$46.400\pm0.182$ & \cellcolor{myblue!20}$\mathbf{48.566\pm0.024}$
& $-4.186\pm0.087$ & \cellcolor{myblue!20}$-2.929\pm0.106$ & \cellcolor{myblue!20}$\mathbf{-0.456\pm0.051}$ \\
$10^{-4}$
& $43.618\pm0.151$ & \cellcolor{myblue!20}$46.398\pm0.181$ & \cellcolor{myblue!20}$\mathbf{48.565\pm0.023}$
& $-4.185\pm0.088$ & \cellcolor{myblue!20}$-2.931\pm0.104$ & \cellcolor{myblue!20}$\mathbf{-0.458\pm0.050}$ \\
$10^{-3}$
& $43.616\pm0.151$ & \cellcolor{myblue!20}$46.399\pm0.183$ & \cellcolor{myblue!20}$\mathbf{48.565\pm0.022}$
& $-4.186\pm0.090$ & \cellcolor{myblue!20}$-2.928\pm0.106$ & \cellcolor{myblue!20}$\mathbf{-0.461\pm0.054}$ \\
$10^{-2}$
& $43.606\pm0.151$ & \cellcolor{myblue!20}$46.384\pm0.180$ & \cellcolor{myblue!20}$\mathbf{48.563\pm0.025}$
& $-4.191\pm0.088$ & \cellcolor{myblue!20}$-2.940\pm0.103$ & \cellcolor{myblue!20}$\mathbf{-0.459\pm0.051}$ \\
\bottomrule
\end{tabular}
\end{adjustbox}
\end{table}

\revisiontext{Table~\ref{tab:adamw_weight_decay} shows negligible sensitivity throughout the tested range. From $\lambda=0$ to $10^{-2}$, the Overall score of $\mathrm{MR}_{\mathrm{pre}}$ remains between 46.384 and 46.400, with BWT between $-2.940$ and $-2.928$; the Overall score of $\mathrm{HMCL}_{\mathrm{CA}}$ remains between 48.563 and 48.566, with BWT between $-0.461$ and $-0.456$, and it is strongest in every condition.} The relatively mild aggregate forgetting of $\mathrm{MR}_{\mathrm{pre}}$ is concentrated in the 14 classification tasks (BWT approximately $-2.69$), whereas its two retrieval tasks have BWT approximately $-4.65$. Hence this control confirms that minimal rotation already protects much of the old classification geometry, while the complete CA correction gives the more consistent cross-task result. Most importantly, the ordering among Vanilla, $\mathrm{MR}_{\mathrm{pre}}$, and $\mathrm{HMCL}_{\mathrm{CA}}$ is unchanged by AdamW weight decay.

\Needspace{0.30\textheight}
\appendixsubsection{Clean Single-Stage Training Time}
\label{apdx:clean_training_time}

\begin{wrapfigure}{r}{0.50\textwidth}
    \centering
    \includegraphics[width=0.98\linewidth]{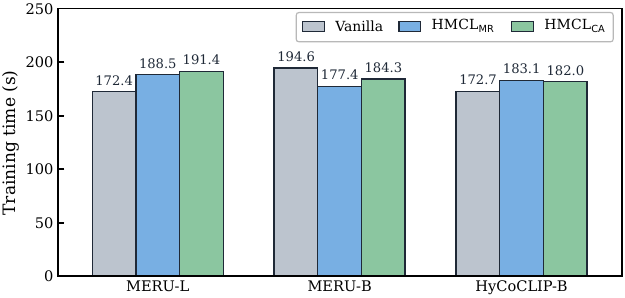}
    \caption{Clean ImageNet-1K training time at the third continual stage (seed 42). Each bar is one complete synchronized \texttt{fit\_task} timing on a common RTX A6000 under the matched schedule; values are seconds. Jobs are serialized, and accepted runs contain no foreign GPU compute PID.}
    \label{fig:clean_training_time}
\end{wrapfigure}

We measure method-side cost on ImageNet-1K, the third task after sequentially learning Aircraft and Caltech-101. For every backbone--method pair, seed 42 uses the same cached features, 10-epoch schedule, batch size, and epoch-level optimizer-step budget as the main comparison. The timer brackets the complete \texttt{fit\_task} call with CUDA synchronization. It therefore includes cached-tensor transfer to the device, optimization and, for HMCL, geometric correction, pullback, and the method's task-end state update, while excluding disk cache loading, evaluation, and checkpoint serialization.

The nine displayed jobs are run serially on the same RTX A6000. The dispatcher first waits for the preceding experiment queue to exit and then requires six consecutive five-second samples with no compute process, at most 2\% utilization, and at most 32 MiB allocated memory. During a run, it records the GPU compute PIDs, utilization, memory, temperature, and power approximately once per second. The presence of any compute PID other than the timed process invalidates that attempt and triggers a fresh idle wait and retry. No displayed attempt was invalidated or retried.

Figure~\ref{fig:clean_training_time} shows that Vanilla and the two HMCL variants remain in the same computational regime: all measurements lie between 172.4 and 194.6 seconds. Averaging the within-backbone ratios to Vanilla gives $1.022\times$ for $\mathrm{HMCL}_{\mathrm{MR}}$ and $1.037\times$ for $\mathrm{HMCL}_{\mathrm{CA}}$. Thus, in this clean single-run diagnostic, the complete HMCL corrections add only a small average cost relative to Vanilla. Because there is one timing run per pair, the \revisiontext{differences below 10\%} should not be interpreted as a statistically resolved ranking.

\appendixsubsection{Pullback Hyperparameter Analysis}
\label{apdx:pullback_sensitivity}

We vary $\beta$ only after fixing the post-optimizer MR correction, task order, protected subspace, and all training hyperparameters. The selected setting uses $\beta=0.125$ from a screening grid spanning 0.05 to 0.25. Relative to the matched $\beta=0.20$ setting, reducing $\beta$ to 0.125 improves \revisiontext{the Overall score} by 0.6\% and mean classification accuracy by 1.2\%, while retrieval R@5 decreases by 2.1\%. It also worsens forgetting, BWT, and final ImageNet retention. The pullback coefficient therefore selects the stability--plasticity operating point and is not treated as a method component in the main ablation.

\end{journalnew}

\end{document}